%% file: sample-manuscript.tex
\documentclass[acmsmall,nonacm]{acmart}

\usepackage{tikz}
\usetikzlibrary{positioning, fit, shapes, arrows.meta}
\usetikzlibrary{matrix, positioning, fit, shapes, arrows.meta,calc}
\usepackage{pifont}
\usepackage{multirow}
\usepackage{array}
\usepackage{rotating}
\usepackage{makecell}
\usepackage{adjustbox}

\newcolumntype{C}[1]{>{\centering\arraybackslash}p{#1}}
\newcolumntype{M}[1]{>{\centering\arraybackslash}m{#1}}

\AtBeginDocument{%
  }

\acmJournal{CSUR}
\acmVolume{37}
\acmNumber{4}
\acmArticle{111}
\acmMonth{8}

\usepackage{tikz}
\usepackage[commandnameprefix=ifneeded]{changes}
\definechangesauthor[name={Changhua Pei}, color=blue]{pch}

\usepackage{makecell}
\usepackage{array}

\begin{document}

\title{A Survey on AgentOps: Categorization, Challenges, and Future Directions}

\author{Zexin Wang}
\authornote{Also with the University of Chinese Academy of Sciences.} 

\author{Jingjing Li}

\affiliation{%
  \institution{Computer Network Information Center, Chinese Academy of Sciences}
  \city{Beijing}
  \country{China}
}

% \author{Changhu Pei}
% \authornote{Also with Hangzhou Institute for Advanced Study, University of Chinese Academy of Sciences.}

% \author{Changhu Pei}
% % \authornote{Both authors contributed equally to this research.}
% \email{chpei@cnic.cn}
% \affiliation{%
%   \institution{Computer Network Information Center, Chinese Academy of Sciences}
%   \city{Beijing}
%   \country{China}
% }

\author{Quan Zhou}
\author{Haotian Si}
\affiliation{%
  \institution{Computer Network Information Center, Chinese Academy of Sciences}
  \city{Beijing}
  \country{China}
}

\author{Yuanhao Liu}
\affiliation{%
  \institution{Hangzhou Institute for Advanced Study, University of Chinese Academy of Sciences.}
  \city{Beijing}
  \country{China}
}

\author{Jianhui Li}
\authornote{Also with Nanjing University.}
%\authornote{Corresponding Authors. Email: lijh@cnic.cn}
% \author{Jingjing Li}
\author{Gaogang Xie}
% \email{zhouquan@cnic.cn}
\affiliation{%
  \institution{Computer Network Information Center, Chinese Academy of Sciences}
  \city{Beijing}
  \country{China}
}

\author{Fei Sun}
% \email{sunfei@ict.ac.cn}
\affiliation{%
  \institution{Institute of Computing Technology, Chinese Academy of Sciences}
  \city{Beijing}
  \country{China}
}

\author{Dan Pei}
% \email{peidan@tsinghua.edu.cn}
\affiliation{%
  \institution{Tsinghua University}
  \city{Beijing}
  \country{China}
}

\author{Changhua Pei}
\authornote{Corresponding Authors. Email: chpei@cnic.cn}
\affiliation{%
  \institution{Computer Network Information Center, Chinese Academy of Sciences}
  \city{Beijing}
  \country{China}
}

% \author{Fengrui Liu}
% \affiliation{%
%   \institution{ByteDance}
%   \city{Beijing}
%   \country{China}
% }
% \author{Tieying Zhang}
% % \affiliation{%
% %   \institution{ByteDance}
% %   \city{San Jose}
% %   \country{United States}
% % }
% \author{Jianjun Chen}
% \affiliation{%
%   \institution{ByteDance}
%   \city{San Jose}
%   \country{United States}
% }

\renewcommand{\shortauthors}{Wang et al.}

\begin{abstract}

As the reasoning capabilities of Large Language Models (LLMs) continue to advance, LLM-based agent systems offer advantages in flexibility and interpretability over traditional systems, garnering increasing attention. However, despite the widespread research interest and industrial application of agent systems, these systems, like their traditional counterparts, frequently encounter anomalies. These anomalies lead to instability and insecurity, hindering their further development. Therefore, a comprehensive and systematic approach to the operation and maintenance of agent systems is urgently needed. Unfortunately, current research on the operations of agent systems is sparse. To address this gap, we have undertaken a survey on agent system operations with the aim of establishing a clear framework for the field, defining the challenges, and facilitating further development. Specifically, this paper begins by systematically defining anomalies within agent systems, categorizing them into intra-agent anomalies and inter-agent anomalies. Next, we introduce a novel and comprehensive operational framework for agent systems, dubbed \textbf{Agent} System \textbf{Op}eration\textbf{s} (\textbf{AgentOps}). We provide detailed definitions and explanations of its four key stages: monitoring, anomaly detection, root cause analysis, and resolution.

\end{abstract}




\maketitle

\section{Introduction}

\input{introduction}

\section{Taxonomy of Agent Systems}
\input{background_categorization}

\section{Anomalies in Agent Systems}
\label{anomalies}
\input{agent_status}

\section{AgentOps: Agent System Operations}
\input{agent_operations}

\section{Challenges and Future Directions of AgentOps}
\input{future_research_directions}

\section{Conclusion}

\input{conclusion}

% \begin{acks}
% \end{acks}

\bibliographystyle{ACM-Reference-Format}
\bibliography{sample-base}

% \appendix


\end{document}

%% file: introduction.tex
With the advent of technologies like DeepSeek-R1 \cite{guo2025deepseekr1} and Claude \cite{claude}, the reasoning capabilities of current Large Language Models (LLMs) are continually being enhanced. Leveraging LLMs as powerful cognitive engines, existing LLM-based agent systems, particularly multi-agent systems, have gained the capacity to accomplish a wide array of complex tasks and social simulations \cite{li2023econagent}, especially when equipped with diverse tools \cite{qin2023toolllm}. Compared to traditional systems like microservice architectures \cite{pei2025flow}, agent systems offer better automation, enhanced interpretability, and greater flexibility. Consequently, research and industrial applications of agent systems are flourishing, with an increasing number of online services \cite{online}, such as customer support and recommendation systems, adopting these agent systems.

% 然而，尽管agent system应用广泛，并不代表其完美无瑕。相比传统的微服务系统，agent system带来的更多的灵活性同时也带来了更多的异常情况。如图1所示，在任务执行过程中，经常由于幻觉导致任务执行失败。在角色扮演中，某个agent 受到攻击也会导致整个模拟过程走向崩溃。因此，为了维持agent system的安全稳定，推动其进一步发展，需要对agent system进行高效的运维。虽然运维技术随着时间在不断发展进步，从早期的manual 运维，到后来的基于规则的运维，再到后来的AIOps。然而不幸的是，agent system在本质上与传统系统有较大区别，LLM驱动的agent的行为特征和hard code 驱动的传统系统行为特征存在本质差别。主要差别在于：（1）agent system中出现的异常种类更加diverse。（2）agent system的可观测性相比传统系统要求更高，还需要额外关注LLM等模块。（3）异常种类的多样化导致agent system中无法使用统一的方法进行异常检测和根因分析。（4）agent system中resolution相对复杂困难，需要从多个视角考虑，需要多轮迭代优化。因此传统的运维技术很难应用到agent system上，亟需一种新的针对agent system的运维技术。

\begin{figure}[t]
    \centering
    \includegraphics[width=0.9\textwidth]{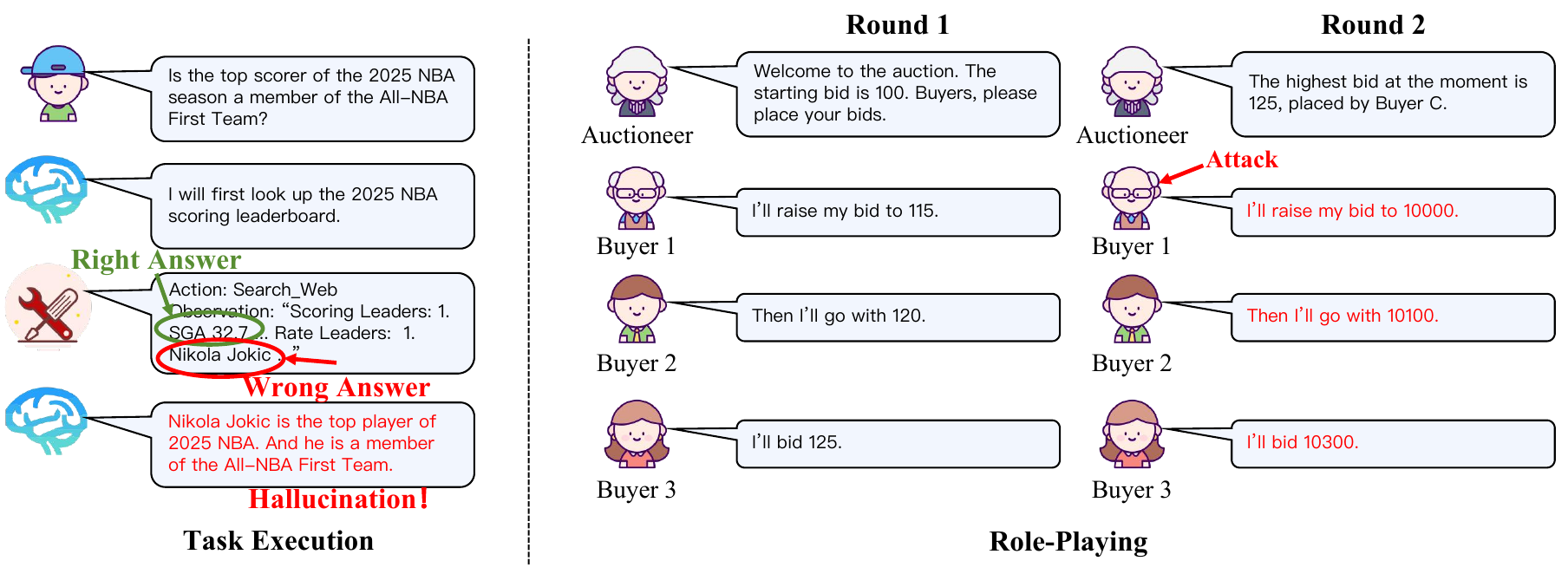}  % 图片文件名（不带扩展名）
    \caption{Anomalies in agent systems. The left side showcases anomalies during task execution, where the agent experiences hallucinations while synthesizing information from web search results, leading to incorrect answers. The right side depicts anomalies in auction role-playing simulations, where an attack on buyer 1 results in abnormally high bids, causing the auction to collapse.}
    \label{fig:intro}  % 图片标签，可用于引用
\end{figure}

However, despite the widespread application of agent systems, they are not without their flaws. Compared to traditional microservice systems, the greater flexibility offered by agent systems also introduces more anomalies. As illustrated in Figure \ref{fig:intro}, task execution often fails due to issues such as hallucinations. In role-playing scenarios, an attack on a single agent can lead to the collapse of the entire simulation. Therefore, to maintain the security and stability of agent systems and to facilitate their further development, efficient operations and maintenance are necessary.

Although operation technologies have been evolving over time, from early manual operations to rule-based methods, and later to Artificial Intelligence for IT Operations (AIOps), agent systems inherently differ significantly from traditional systems. The behavioral characteristics of LLM-driven agents are fundamentally different from those in hard-coded traditional systems. The key differences include: (1) A wider variety of anomalies occurs in agent systems. (2) Agent systems demand higher observability than traditional systems, requiring additional attention on modules like LLMs. (3) The diversity of anomalies makes it impossible to use a unified approach for anomaly detection and root cause analysis in agent systems. (4) Resolution in agent systems is relatively complex and challenging, requiring consideration from multiple perspectives and iterative optimization. As a result, traditional operation techniques are difficult to apply to agent systems, leading to an urgent need for new, tailored operation technologies for these systems.

% 然而，目前并没有研究系统阐述如何有效的针对agent system进行运维，大多数研究还是停留在agent system中的某个小的方面。例如LiFEFEI主要针对agent的范式以及分类进行了详细的阐述。HallucinationSurvey针对foundation model中hallucination进行了详细的阐述，包括hallucination的定义以及检测hallucination的方法。SecuritySurvey主要研究了multi-agent system中的安全问题，这里的安全 threats主要指的是外在的恶意的攻击，其将Threats分为 intra-excution security 和 interaction security。GUIAGENTs 主要针对GUI Agents的安全问题以及相应的评估方法进行了详细介绍。

Currently, there is a lack of comprehensive research on effective operations and maintenance strategies specifically for agent systems. Most studies remain focused on isolated aspects of agent systems rather than addressing their overall operational challenges. For example, \citet{lifeifei} expound on agent paradigms and classifications; \citet{hallucinationsurvey} delve into hallucinations in foundation models, covering its definition and detection methods; \citet{security} explore security issues in multi-agent systems, primarily covering external malicious attacks and categorizing threats into intra-execution security and interaction security. \citet{guiagents} provide detailed insights into security issues and evaluation methods for GUI agents.

% 因此，为了进一步推动agent system的发展，本paper 提出了AgentOps概念，一种新的针对agent system的运维流程。首先我们作出了确切的agent system异常定义，并针对agent system 中的异常进行了系统的分类，主要分为intra-agent anomalies 和 inter-agent anomalies。这两类异常，涵盖了agent system的从pre excution、excution和post excution三个阶段。其次，我们参考传统运维，将agent system中的运维流程同样分为monitoring、anomlay detection，root cause analysis和resolution四个阶段。同时针对每个运维阶段在agent system中出现的新的挑战，做出了详细的定义以及可能的解法。就我们所认知的，我们是第一个系统提出AgentOps概念的工作，并且规范化的定义了AgentOps的各个流程。

To further advance the development of agent systems, this paper introduces the concept of \textbf{Agent} System \textbf{Op}eration\textbf{s} (\textbf{AgentOps}), a novel operations and maintenance framework specifically designed for agent systems. First, we provide a precise definition of anomalies within agent systems and offer a systematic classification, primarily dividing them into intra-agent and inter-agent anomalies. These two categories encompass the stages of pre-execution, execution, and post-execution in the agent system lifecycle. Additionally, drawing inspiration from traditional operation practices, we divide the operations and maintenance process for agent systems into four phases: monitoring, anomaly detection, root cause analysis, and resolution. For each phase, we identify new challenges that arise within agent systems and propose detailed definitions and potential solutions. To the best of our knowledge, this is the first work to systematically propose the concept of AgentOps and standardize the definition of its various processes.

%% file: background_categorization.tex
\subsection{Definition of Agent Systems}

% agent System指的是能够自动化的感知环境、决策、行动并最终完成任务的智能系统，一般由多个Agent构成。其主要包含以下几个核心能力。（1）感知交互环境：Agent本身具备的知识有限，因此需要不断与环境交互来获取知识或者反馈才能最终完成任务目标。 （2）自主思考决策：agent System的一大特点就是智能化，不需要人的参与便可以完成任务。因此其必须像人一样具备自主思考和决策的能力。 （3）知识管理：完成一个任务往往需要大量的背景知识，即便Agent能力足够强，也并没有办法同一时间处理很多知识，因此需要分步骤循序渐进。这就需要有效的知识管理。 （4）多Agent交互： 现实中任务往往会很复杂，一个智能体没有办法解决。因此需要分工协作，多智能体共同完成任务。 这就需要多智能体之间有效的沟通和交互，否则很容易出现各自为战的情况。

% 随着LLM的出现与发展，由于其优秀的多模态数据理解和推理能力、工具调用能力，现有的agent System一般都基于LLM实现。因此，本文主要针对的也是LLM agent System Operations。表1是上文提及的四项能力在LLM-Based 的Agent System中的体现。

Agent systems refer to intelligent systems capable of perceiving their environment, making decisions, taking actions, and ultimately completing tasks autonomously. As shown in Table \ref{tab:llmagentsystem}, these systems are generally composed of multiple agents and embody four core capabilities \cite{baipishu}.

% \begin{itemize}
%     \item \textbf{Perception of the Interactive Environment:} Given the limited inherent knowledge of agents, it is crucial for them to continually interact with their environment to acquire the necessary knowledge or feedback needed to achieve their objectives.
%     \item \textbf{Autonomous Reasoning and Decision-Making:} A defining feature of an agent system is its ability to complete tasks without human intervention. Therefore, it must possess the capability to think and make decisions independently, much like a human.
%     \item \textbf{Knowledge Management:} Accomplishing a task often requires extensive background knowledge. Even the most capable agents cannot handle vast amounts of information simultaneously, necessitating a step-by-step approach. Effective knowledge management is essential to facilitate this process.
%     \item \textbf{Multi-Agent Interaction:} In the real world, tasks are often too complex for a single intelligent agent to handle. Thus, division of labor and cooperation among multiple agents is necessary to complete a task successfully. This requires effective communication and interaction among the agents to avoid isolated efforts.
% \end{itemize}

% \subsection{LLM-based Agent Systems}

\begin{figure}[t]
    \centering
    \includegraphics[width=0.9\textwidth]{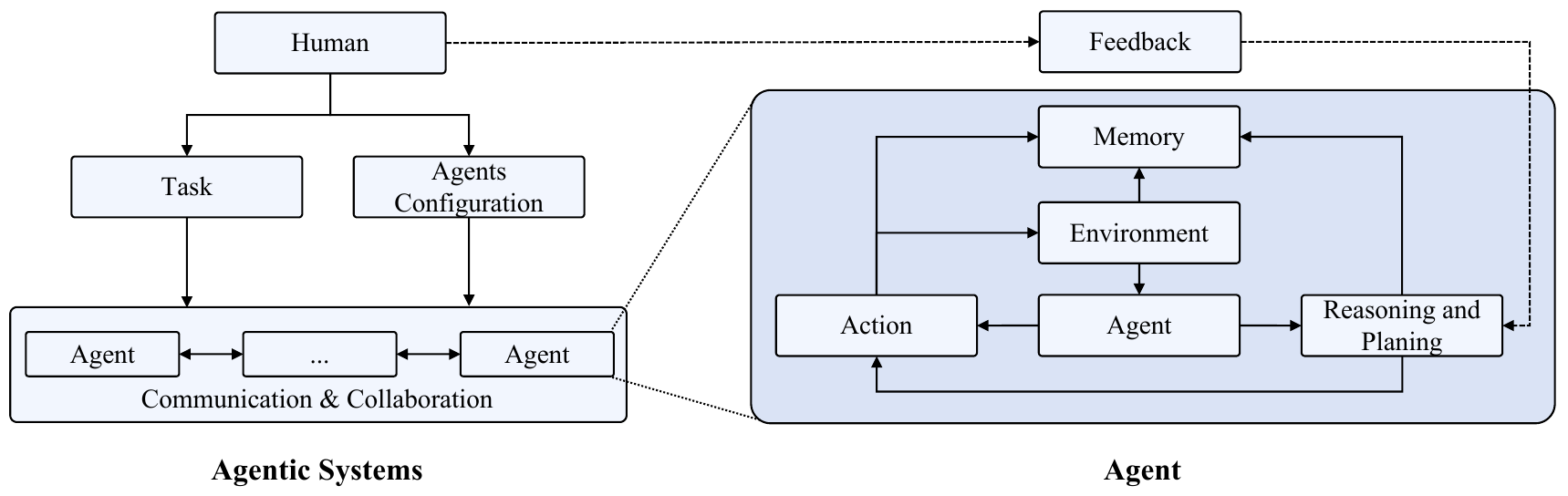}  % 图片文件名（不带扩展名）
    \caption{Components of agent systems.}
    \label{fig:agentsystem}  % 图片标签，可用于引用
\end{figure}

With the advent and development of LLMs, contemporary agent systems are frequently based on these models due to their superior capabilities in understanding and reasoning with multimodal data and their proficiency in tool utilization. Therefore, this paper primarily focuses on LLM-based agent system operations. Table \ref{tab:llmagentsystem} illustrates the specific manifestations of four capabilities within LLM-based agent systems.

\begin{figure}[htbp]
    \centering
    \includegraphics[width=0.9\textwidth]{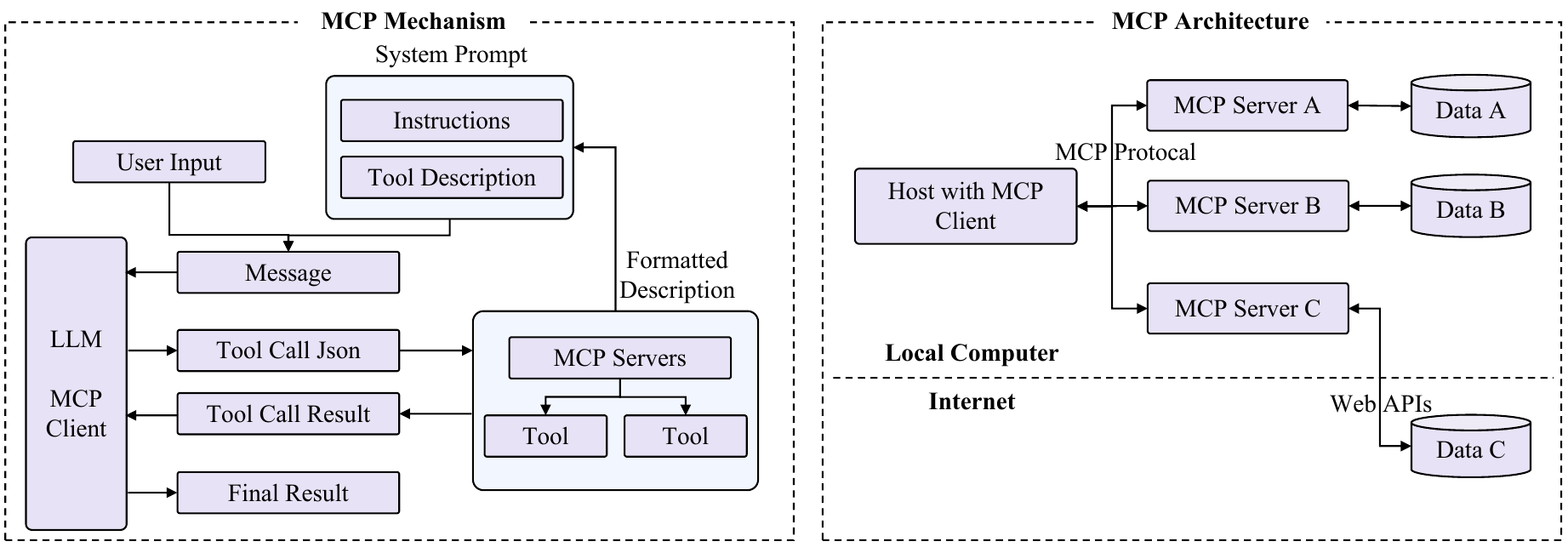}  % 图片文件名（不带扩展名）
    \caption{Mechanism and architecture of MCP. As shown in the diagram on the right, the architecture of MCP features a connection between the host and the MCP server established through the MCP client. The MCP server is responsible for executing specific tools. Specifically, as illustrated in the diagram on the left, the description information of MCP is input into the system prompt. The LLM then decides which MCP server to call and passes the structured calling information to the MCP server to complete the final invocation.}
    \label{fig:mcp}  % 图片标签，可用于引用
\end{figure}

\begin{table}
\centering
\caption{Capabilities in LLM-based agent systems.}
\label{tab:llmagentsystem}
\resizebox{\textwidth}{!}{
\begin{tabular}{lll} 
\toprule
\multicolumn{1}{c}{\textbf{Componets}}    & \multicolumn{1}{c}{\textbf{LLM-based agent Systems}} & \multicolumn{1}{c}{\textbf{Methods}}                                                            \\ 
\midrule
Perception of the Interactive Environment & Tool Calling                                         & Function Call, MCP\cite{mcp}                                                   \\
Autonomous Reasoning and Decision-Making  & Reasoning and Act                                    & ReAct\cite{yao2023react}, Reflexion\cite{shinn2023reflexion}  \\
Knowledge Management                      & Short \& Long Memory                   & Prompt, RAG\cite{edge2024graphrag}                                             \\
Multi-Agent Interaction                   & Agent Communicating                                  & A2A\cite{A2A}, ACP\cite{ACP}, ANP\cite{anp}  \\
\bottomrule
\end{tabular}
}
\end{table}

% LLM-based agent systems通过tool call实现与环境的交互，从而不断获取环境的观测结果作为反馈，为自动化的reasoning和decision-making提供参考。早期的tool call通过function call实现，即针对特定任务编写相应的函数并将函数的描述信息通过特定格式输入给LLM。然而不同LLM之间的差异导致需要不断变更格式等信息，Model Context Protocol的出现从根本上解决了这个问题。MCP统一了LLM与外部数据源和工具之间的通讯协议。MCP的工作流程如如图1所示，Host通过MCP Client与MCP Server建立联系，而MCP Server连接到相应的资源并执行相应的工具最终将结果返回给clinet。MCP的出现利于服务商开发相应的服务API，同时避免了重复服务浪费资源。

\begin{itemize}
    \item \textbf{Tool Call:} LLM-based agent systems utilize tool calls to interact with their environment, continuously acquiring observational data as feedback. This feedback serves as a reference for automating reasoning and decision-making processes. In earlier iterations, tool calls were implemented via function calls, where specific functions were written for particular tasks, and their descriptions were formatted and inputted into the LLM. However, variances between different LLMs necessitated frequent adjustments to formatting and other information. The advent of the Model Context Protocol (MCP) \cite{mcp} fundamentally addresses this issue by standardizing the communication protocol between LLMs, external data sources, and tools. The mechanism and architecture of MCP are illustrated in Figure \ref{fig:mcp}. The host establishes a connection with the MCP server via the MCP client, while the MCP server links to the appropriate resources and executes the necessary tools to finally return the results to the client. The introduction of MCP facilitates the development of corresponding service APIs by providers and simultaneously avoids resource wastage.
    \item \textbf{Reasoning and Act:} With the emergence of reasoning LLMs like DeepSeek-R1 \cite{guo2025deepseekr1}, the reasoning capabilities of LLMs have become increasingly robust, sufficiently supporting LLM-based agent systems in automating decision-making and executing a variety of complex tasks. Numerous methods harness prompt engineering to effectively leverage the reasoning capabilities of LLMs to enhance agent systems. For instance, Chain-of-Thought \cite{wei2022chainofthought} guides LLMs to reason step-by-step using a few-shot approach; ReAct \cite{yao2023react} proposes a method where ``thought'' precedes ``action''; and Reflexion\cite{shinn2023reflexion} suggests reflecting on the entire reasoning path after a certain period.
    \item \textbf{Short \& Long Memory :} Much like humans, LLM-based agent systems have a limited capacity to store knowledge due to token size constraints, necessitating effective knowledge management. A common paradigm for knowledge management involves categorizing information into short-term and long-term memory. Short-term memory consists of a small set of knowledge and observations closely tied to the current task, typically provided to the LLM through prompt engineering. Long-term memory, on the other hand, includes a large amount of knowledge that may not be immediately relevant to the task at hand but is made available to agent systems only when needed at specific steps of task execution. Long-term memory is generally managed using Retrieval-Augmented Generation (RAG) and vector databases. When required, knowledge is recalled based on the similarity between query vectors and knowledge vectors. Innovations in RAG and vector databases, such as GraphRAG \cite{edge2024graphrag}, continue to evolve this landscape.
    \item \textbf{Agent Communicating:} Despite the growing interest in multi-agent systems, numerous studies indicate that in many scenarios, their performance is sometimes no better than that of a single agent. This underscores the critical role of agent communication. Effective division of labor and communication should ideally lead to improved outcomes. Consequently, communication standards and protocols for LLM agents have been proposed, including the well-known Agent-to-Agent (A2A) \cite{google2025a2a} protocol. A2A introduces the concept of an agent card to systematically define each agent's capabilities, while also allowing the client agent to efficiently manage tasks and all agents involved.
\end{itemize}

% 随着Deepseek-R1等reasoning LLM的出现，LLM推理能力越来越强足够支撑LLM-based agent systems自动化的决策并完成各种复杂任务。很多方法利用prompt engineering的方式 有效的利用LLM的推理能力增强agent systems。例如，Chain-of-Thought通过Few-shot的方式引导LLM分步骤推理；ReAct提出先thought再act；Reflexion则提出一段时间后对于整个推理路径进行反思。

% 像人一样，LLM-based agent systems 由于token size的限制，存储知识的能力有限，因此需要进行有效的知识管理。目前常用的知识管理范式是分类处理short和long memory。short memory指的是和当前任务紧密相关的少量知识以及短时间内执行的步骤，一般通过prompt engineering的方式给予LLM。而long memory指的是大量不一定会被当前任务里用到的knowledge，只有在执行任务到特定步骤需要的时候，才会提供给agent systems。long memory一般通过RAG 和向量数据库的方式管理，当需要时，通过query向量与knowledge向量之间的相似度召回知识。RAG和向量数据库也在不断发展，其中包括GraphRAG等。

%尽管多智能体系统火热发展，但是很多研究表明，多智能体在很多场景下的效果甚至不如单个智能体。这表明了agent communication的作用。如果能够有效分工和沟通，理应达到更好的效果。因此针对LLM-Agent的通讯规范和协议被提出，其中包括 著名的Agent-to-Agent （A2A）协议。A2A设计了agent card来规范化的定义每个agent的能力，同时client agent有效的管理task和所有agent。

\subsection{Taxonomy of Agent Systems}

% 在基于大型语言模型（LLMs）的智能体系统（Agent Systems）研究中，根据智能体数量可以进行分类：Single-Agent Systems (SAS） 和 Multi-Agent Systems(MAS)。
% SAS 由单一LLM驱动的智能体构成核心处理单元。目前常见的SAS任务类型包括：
% ~~**Planning**： 智能体需分解目标状态，推断可行操作序列。~~
% **Reasoning**： 利用内置或上下文知识进行逻辑/数学推演、因果推断[The AI Scientist: Towards Fully Automated Open-Ended Scientific Discovery]。
% **Conversation**： 维持对话状态，理解用户意图，生成语义连贯的回复序列。对话不局限于文字语言的交流，也包含图像、音频、视频等多模态的交流[gpt-4o]。
% **Interaction**： 处理与外部相对简单、可预测环境的直接信息交换或动作执行[WebArena]。
% MAS则由多个LLM智能体构成，其在共享环境中交互协作或竞争。MAS的优势在于处理单智能体无法有效应对的分布式、并发、或具有竞争/协同需求的复杂问题。其典型任务类型包括：
% **Role-Playing&Simulation**： 智能体被赋予特定身份、背景与行为准则，在设定框架内进行符合角色特性的互动,模拟复杂系统动态（如社会行为、经济模式、流行病传播），探索宏观现象如何从微观个体互动中涌现[War and Peace (WarAgent): LLM-based Multi-Agent Simulation of World Wars、EconAgent: Large Language Model-Empowered Agents for Simulating Macroeconomic Activities]。
% **Cooperation&Collaboration**： 多个智能体共享或兼容目标，通过分工、协商与信息共享达成共同目的[Communicative Agents for Software Development]。
% **Game-Theoretic Interaction**： 智能体目标存在潜在冲突，决策过程需考虑他者策略，涉及竞争、谈判或激励机制设计（如拍卖、多代理谈判、零和博弈）[A Survey on Large Language Model Based Game Agents]。
% 需要指出，MAS也能完成SAS任务，并达到更优秀的效果，但会导致额外的异常与维护成本，需要根据实际场景进行抉择[Single-agent or Multi-agent Systems? Why Not Both?]。
% 表3中列举了一些代表性的agent任务类型、模型、benchmark和success rate。

\begin{figure}[t]
\centering
\begin{tikzpicture}[
  node distance=0.6cm and 2.2cm,
  box/.style={draw, rectangle, rounded corners=2pt, minimum width=1.5cm, minimum height=4mm, align=center, font=\tiny, fill=blue!5},
  line/.style={-{Stealth}, thick},
  bend line/.style={
    draw, line,
    to path={(\tikztostart.east) -- ++(#1,0) |- (\tikztotarget.west)}
  }
]

% Level 1
\node[box, fill=blue!15] (main) {Agent System};

% Level 2 (reduced to two items, vertically centered)
\node[box, right=1cm of main, yshift=1cm](single) {Single-Agent};
\node[box, right=1cm of main, yshift=-1cm](multi) {Multi-Agent};

% Level 3 under Single-Agent
\node[box, right=1cm of single, yshift=0.6cm] (reason){Reasoning};
%\node[box, right=1cm of single, yshift=0.3cm](plan) {Planning};
\node[box, right=1cm of single, yshift=0cm](conv){Conversation};
\node[box, right=1cm of single, yshift=-0.6cm] (interaction) {Interaction};

% Level 3 under Muti-Agent
\node[box, right=1cm of multi, yshift=-0.6cm](cooper){Cooperation \& Collaborative};
\node[box, right=1cm of multi, yshift=0cm] (game){Game-Theoretic Interaction};
\node[box, right=1cm of multi, yshift=0.6cm](sim){Role-Playing \& Simulation};
%\node[box, right=1cm of multi, yshift=0.9cm](role){Role-Playing};

% Connections from Level 1 to Level 2
\draw[bend line=0.6cm] (main) to (single);
\draw[bend line=0.6cm] (main) to (multi);

% Connections for Single-Agent branch
\draw[bend line=0.6cm] (single) to (reason);
%\draw[bend line=0.6cm] (single) to (plan);
\draw[bend line=0.6cm] (single) to (conv);
\draw[bend line=0.6cm] (single) to (interaction);

% Connections for Multi-Agent branch
\draw[bend line=0.6cm] (multi) to (sim);
%\draw[bend line=0.6cm] (multi) to (role);
\draw[bend line=0.6cm] (multi) to (cooper);
\draw[bend line=0.6cm] (multi) to (game);

\end{tikzpicture}
\caption{Taxonomy of Agent Systems.}
\label{fig:taxonomyofAS}
\end{figure}
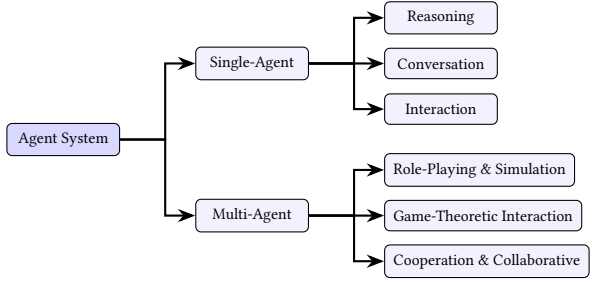

In the research of agent systems powered by LLMs, a fundamental classification can be made based on the number of agents involved. As illustrated in Figure \ref{fig:taxonomyofAS}, agent systems can be divided into Single-Agent Systems (SAS) and Multi-Agent Systems (MAS).

SAS are centered on a single LLM-powered agent, which serves as the core processing unit. These systems are typically employed in the following types of tasks:

\noindent\textbf{Reasoning}: The agent performs logical or mathematical inference, as well as causal reasoning, by leveraging either embedded knowledge or contextual cues. For example, the AI Scientist \cite{lu2024ai} explores automated scientific discovery through open-ended reasoning processes.

\noindent\textbf{Conversation}: The agent maintains dialogue state, interprets user intent, and generates semantically coherent responses. Dialogue tasks extend beyond textual interactions to include multimodal communication involving images, audio, and video, as exemplified by systems such as GPT-4o \cite{islam2024gpt}.

\noindent\textbf{Interaction}: The agent engages directly with relatively simple and predictable external environments, either by exchanging information or executing actions. A representative example is WebArena \cite{zhou2023webarena}, where agents interact with structured web interfaces.

In contrast, MAS consist of multiple LLM-powered agents operating within a shared environment. These agents may cooperate or compete with one another, making MAS particularly well-suited for distributed, concurrent, or strategically complex problems that are challenging for single-agent systems. Typical MAS applications include:

\noindent\textbf{Role-Playing and Simulation}: Agents are assigned specific identities, backgrounds, and behavioral rules and interact in character-consistent ways within a predefined framework. Such simulations enable the study of emergent macro-level phenomena arising from micro-level interactions, e.g., social dynamics, economic systems, and epidemic spread. Examples include WarAgent \cite{hua2024war}, which simulates geopolitical conflicts, and EconAgent \cite{li2023econagent}, which models macroeconomic activities.

\noindent\textbf{Cooperation and Collaboration}: Multiple agents share common or partially aligned goals and achieve them through task decomposition, negotiation, and information exchange. For instance, in the ChatDev \cite{Qian2023ChatDevCA}, agents collaboratively contribute to code generation and debugging.

\noindent\textbf{Game-Theoretic Interaction}: In scenarios with potentially conflicting objectives, agents must account for others’ strategies in their own decision-making processes. This involves elements of competition, negotiation, and incentive mechanism design, such as auctions, multi-agent bargaining, and zero-sum games \cite{hu2024survey}.

\begin{table}
    \centering
     \caption{Representative Tasks \& Methods \cite{swebch_2025, zhang2025webpoilot,wang2025openhands,he2024webvoyager}. }
     \label{tab:taxonomyofAS}
     \small{
    \begin{tabular}{cccc}\toprule
         Taxonomy&  Method &  Benchmark& success rate(\%)\\\midrule
         Web Task&  WebPilot&  WebArena& 37.20%
\\
         Web Task&  WebPilot&  MiniWoB++& 95.60%
\\
         Web Task&  WebVoyager&  WebVoyager& 59.10%
\\
         Software Development&  ChatDev&  SRDD& 88.00%
\\
         Software Development&  SWE-agent&  SWE-bench& 33.83%
\\
         Software Development&  Trae&  SWE-bench-Verified& 75.20%
\\
         Software Development&  ExpeRepair&  SWE-bench-Lite& 60.33%
\\
         Research Assistance&  GPTSwarm(OpenHands)&  GAIA& 32.10%
\\
 Research Assistance& Agent Laboratory& Agent Laboratory&95.70%
\\
         Social Simulation&  Waragent&  WWII& 45.89%
\\ \bottomrule
    \end{tabular}
    
    }
\end{table}

It is worth noting that MAS are capable of accomplishing traditional SAS tasks, often with improved performance. However, this comes at the cost of increased system complexity, potential emergent failures, and higher maintenance overhead. Therefore, the choice between SAS and MAS should be made based on the specific requirements and constraints of the target application domain \cite{gao2025single}. Table \ref{tab:taxonomyofAS} presents selected representative agent task types, methods, benchmarks, and success rates.

%% file: agent_status.tex
\subsection{Definition of Anomalies in Agent Systems}

The previous discussion highlighted that success rates in various types of agent systems are not particularly high, indicating the presence of numerous exceptions hindering the successful completion of tasks. According to Who\&When \cite{icml2025}, they posit that anomalies within these systems predominantly occur at a specific step during task execution. Specifically, if an intervention at a particular anomalous step transforms it into a regular step, thereby ensuring the successful completion of the task, that step can be identified as the anomalous step. Formally, this can be expressed as follows:

Suppose the execution trajectory of a task is represented by $\sigma = (s_0, a_1, s_1, \ldots, a_t, s_t)$, where $s_i$ denotes the state of the agent system and $a_i$ represents a specific action. The function $f(\sigma)$ indicates whether a trajectory $\sigma$ successfully completes a task, yielding 1 for success and 0 for failure. The function $ g(\sigma, i) $ represents the modification of a certain step $a_i$ to a normal step. If $ f(\sigma) = 0$ and $ f(g(\sigma,i)) = 1$, then the $i^{th}$ step is identified as the anomalous step.

% However, this definition is quite limited. In the execution of tasks within an agent system, anomalies do not necessarily manifest at the action level or within specific steps. For instance, task description information, agent configuration details, and environmental factors such as network latency and packet loss can significantly impact task execution. 

% 然而，这个定义有很大局限性。传统的微服务系统一般长期稳定运行，不需要关注pre excution阶段和post excution 阶段。然而agent system执行任务和pre excution阶段的任务提示强相关，并且post excution执行成功也不代表没有异常发生 （reasoning 幻觉等也会给出答案）。因此，如图1所示，我们将agent system中的anomalies定义为涵盖pre excution、excution、post excution 三阶段的任何导致task中断或者无法有效完成的现象。

% 针对以上定义, we propose a new taxonomy method for anomalies in agent systems. As depicted in Figure \ref{fig:taxonomyofanomalies}, we categorize all anomalies into two types: intra-agent anomalies and inter-agent anomalies.

\begin{figure}[htbp]
    \centering
    \includegraphics[width=0.85\textwidth]{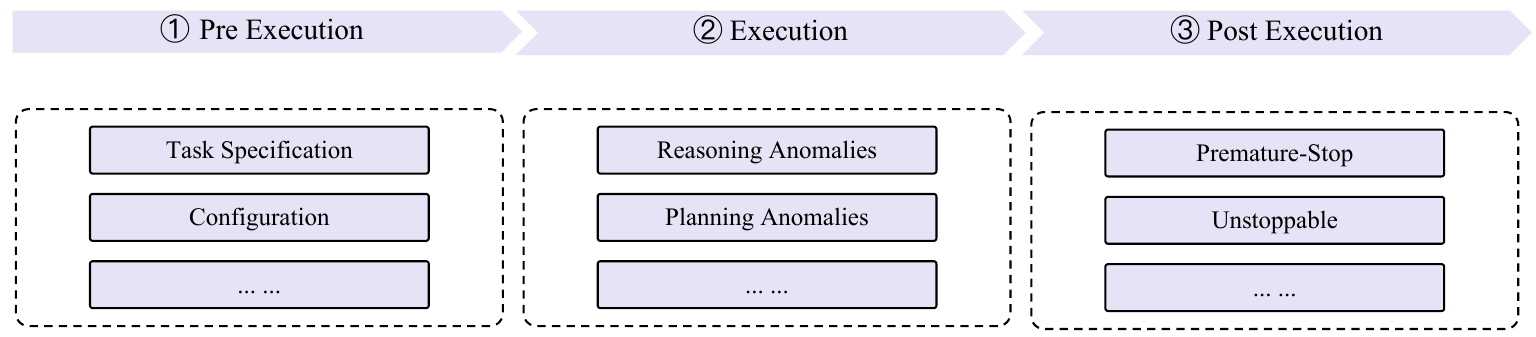}  % 图片文件名（不带扩展名）
    \caption{Definition of anomalies in agent systems.}
    \label{fig:anomalies}  % 图片标签，可用于引用
\end{figure}

However, this definition is quite limited. Traditional microservice systems typically run stably over long periods and do not require consideration of the pre-execution and post-execution phases. In contrast, an agent system's task execution is strongly correlated with its pre-execution prompts, and successful post-execution completion does not necessarily mean no anomalies occurred (as reasoning hallucinations, among others, may still provide incorrect answers). Thus, as illustrated in Figure \ref{fig:anomalies}, we define anomalies in agent systems as any occurrences during the pre-execution, execution, or post-execution phases that lead to task interruption or failure to complete effectively.

% In light of this definition, we propose a new taxonomy method for anomalies in agent systems. As depicted in Figure \ref{fig:taxonomyofanomalies}, we categorize all anomalies into two types: intra-agent anomalies and inter-agent anomalies.

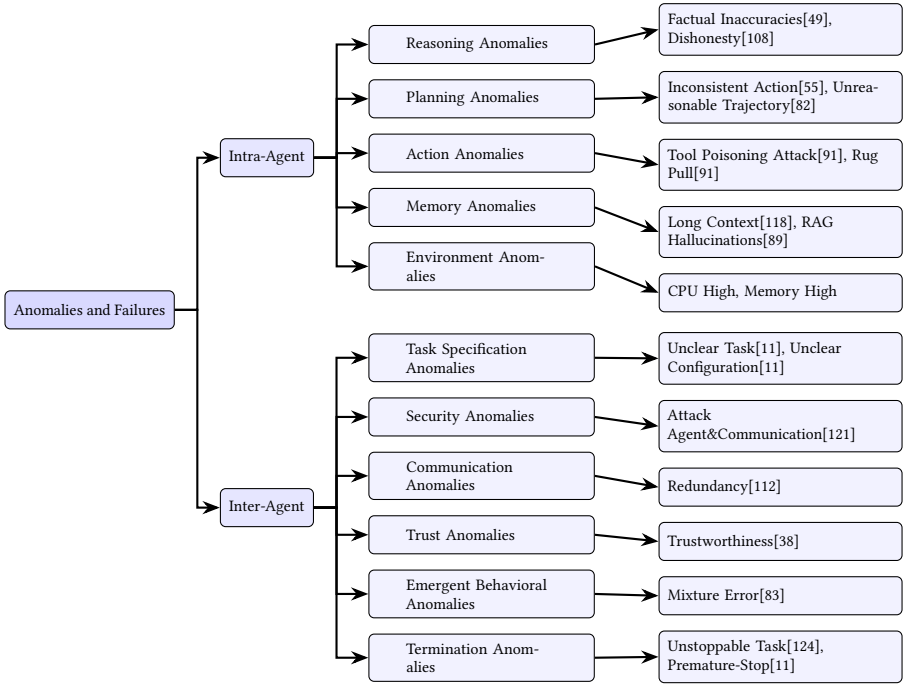
\begin{figure}[t]
\centering
\begin{tikzpicture}[
  node distance=0.6cm and 1.8cm,
  box4/.style={
  draw, rectangle, rounded corners=2pt, 
  minimum width=1cm, minimum height=5mm,
  align=left, font=\tiny, fill=blue!5,
  text width=3cm  % 或你想要的宽度
},
box3/.style={
  draw, rectangle, rounded corners=2pt, 
  minimum width=3cm, minimum height=5mm,
  align=left, font=\tiny, fill=blue!5,
  text width=2cm  % 控制文字自动换行并左对齐
},
  box/.style={
    draw, rectangle, rounded corners=2pt, 
    minimum width=1cm, minimum height=5mm,
    align=left, font=\tiny, fill=blue!5
  },
  box2/.style={
    draw, rectangle, rounded corners=2pt, 
    minimum width=3cm, minimum height=5mm,
    align=left, font=\tiny, fill=blue!5
  },
  line/.style={-{Stealth}, thick},
  bend line/.style={
    draw, line,
    to path={(\tikztostart.east) -- ++(0.3,0) |- (\tikztotarget.west)}
  },
  straight line/.style={
    draw, line,
    to path={(\tikztostart.east) -- (\tikztotarget.west)}
  }
]

% Level 1
\node[box, fill=blue!15, align=center] (main) {Anomalies and Failures};

% Level 2
\node[box, right=0.6cm of main, yshift=2.0cm, fill=blue!10] (agent-level) {Intra-Agent};
\node[box, right=0.6cm of main, yshift=-2.6cm, fill=blue!10] (system-level) {Inter-Agent};

% Level 3 - Intra-Agent
\matrix (intra) [right=0.6cm of agent-level, matrix of nodes, column sep=0.3cm, row sep=0.2cm, nodes={box3, anchor=west}] {
  \node (reason-ano) {Reasoning Anomalies}; \\
  \node (plan-ano) {Planning Anomalies}; \\
  \node (action-ano) {Action Anomalies}; \\
  \node (memory-ano) {Memory Anomalies}; \\
  \node (environment-ano) {Environment Anomalies}; \\
};

% Level 3 - Inter-Agent
\matrix (inter) [right=0.6cm of system-level, matrix of nodes, column sep=0.3cm, row sep=0.2cm, nodes={box3, anchor=west}] {
  \node (taskspec-ano) {Task Specification Anomalies}; \\
  \node (security-ano) {Security Anomalies}; \\
  \node (communicate-ano) {Communication Anomalies}; \\
  \node (trust-ano) {Trust Anomalies}; \\
  \node (emergent-ano) {Emergent Behavioral Anomalies}; \\
  \node (early-ano) {Termination Anomalies}; \\
};

% Level 4 examples - right of level 3
\matrix (intra-ex) [right=0.6cm of intra, matrix of nodes, column sep=0.3cm, row sep=0.2cm, nodes={box4, anchor=west}] {
  \node (reason-ano-example) {Factual Inaccuracies\cite{huang2024opera}, Dishonesty\cite{yang2024alignment}}; \\
  \node (plan-ano-example) {Inconsistent Action\cite{67}, Unreasonable Trajectory\cite{110}}; \\
  \node (action-ano-example) {Tool Poisoning Attack\cite{ai-infra-guard}, Rug Pull\cite{ai-infra-guard}}; \\
  \node (memory-example) {Long Context\cite{CoA}, RAG Hallucinations\cite{sun2024redeep}}; \\
  \node (environment-example) {CPU High, Memory High}; \\
};

\matrix (inter-ex) [right=0.6cm of inter, matrix of nodes, column sep=0.3cm, row sep=0.2cm, nodes={box4, anchor=west}] {
  \node (taskspec-example) {Unclear Task\cite{whydomasfail}, Unclear Configuration\cite{whydomasfail}}; \\
  \node (security-example) {Attack Agent\&Communication\cite{zhou2025guardian}}; \\
  \node (communicate-example) {Redundancy\cite{zhang2024agentprune}}; \\
  \node (trust-example) {Trustworthiness\cite{atrust}}; \\
  \node (emergent-example) {Mixture Error\cite{blogemergent}}; \\
  \node (early-example) {Unstoppable Task\cite{recur}, Premature-Stop\cite{whydomasfail}}; \\
};

% Draw lines
\foreach \from/\to in {
  main/agent-level, main/system-level,
  agent-level/reason-ano, agent-level/plan-ano, agent-level/action-ano,
  agent-level/memory-ano, agent-level/environment-ano,
  system-level/taskspec-ano, system-level/communicate-ano, system-level/security-ano,
  system-level/trust-ano, system-level/emergent-ano,
  system-level/early-ano}
  \draw[bend line] (\from) to (\to);
  
\foreach \from/\to in {
  reason-ano/reason-ano-example, plan-ano/plan-ano-example,
  action-ano/action-ano-example, memory-ano/memory-example,
  environment-ano/environment-example,
  taskspec-ano/taskspec-example, communicate-ano/communicate-example,
  trust-ano/trust-example, emergent-ano/emergent-example,
  security-ano/security-example, early-ano/early-example}
  \draw[straight line] (\from) to (\to);

\end{tikzpicture}
\caption{Taxonomy of Anomalies in Agent Systems.}
\label{fig:taxonomyofanomalies}
\end{figure}

In light of this definition, we propose a new taxonomy method for anomalies in agent systems. As previously discussed, agent systems can be classified into single-agent and multi-agent systems. Consequently, anomalies may arise either within a single agent or during interactions among multiple agents. This is similar to traditional service architectures, where anomalies can occur within the internal processes of a single service or during inter-service communication. Thus, as depicted in Figure \ref{fig:taxonomyofanomalies}, we categorize all anomalies into two types: intra-agent anomalies and inter-agent anomalies.

% \subsection{Taxonomy of Anomalies}

\noindent\textbf{Intra-Agent Anomalies:} Intra-agent anomalies focus primarily on issues within a single agent. Generally, in an agent system, each agent is responsible for a subtask during task execution. However, various errors may occur within an agent as it carries out its subtask. Based on the phase in which these anomalies occur, intra-agent anomalies can be further classified into reasoning anomalies, planning anomalies, action anomalies, memory anomalies, and environment anomalies.

\noindent\textbf{Inter-Agent Anomalies:} Inter-agent anomalies are examined from a global perspective, moving beyond individual agents to focus on the interactions between different agents, as well as the overall system's security, stability, and task execution. These anomalies can be categorized into task specification anomalies, security anomalies, communication anomalies, trust anomalies, emergent behavioral anomalies, and termination anomalies.

The following sections will provide detailed explanations of these types of anomalies.

\subsection{Intra-Agent Anomalies}

% agent system完成复杂的任务，需要多个agent分别完成不同的子任务。agent完成子任务占据了task执行的大部分时间。如图1所示，agent结构复杂，包含多个模块。在执行子任务的时候需要进行reasoning和plan，并且需要通过多种与环境进行多种交互。因此很容易出现异常。gent-level anomalies是agent systems中最常见的异常。根据异常出现的位置的不同，我们将agent-level的异常分为Reasoning Anomalies，Planing Anomalies Action Anomalies Memory Anomalies 和Environment Anomalies。下面将详细介绍每类异常。

To accomplish complex tasks, an agent system requires multiple agents to complete different subtasks. The completion of these subtasks by agents constitutes the majority of the task execution time. As illustrated in Figure \ref{fig:agentsystem}, the structure of an agent is complex and comprises multiple modules. During the execution of subtasks, agents must engage in reasoning and planning, as well as interact with the environment in various ways, which can easily lead to anomalies. Intra-agent anomalies are the most common in agent systems. 
% Based on the location where the anomaly occurs, we categorize intra-agent anomalies into reasoning anomalies, planning anomalies, action anomalies, memory anomalies, and environment anomalies. Each type of anomaly will be discussed in detail below.

\subsubsection{Reasoning Anomalies}

% agent利用认知系统推理，进而指导后续的action，是完成复杂任务的基础。近些年来，很多提升推理能力的方法被提出，其中包括基于finetuning的SFT、RLHF、Search-R1，deepseek-R1，以及基于prompt engineering的CoT、Reflexion、Self- Consistency、CoK、StepBack。然而，即便有了这么多技术的支持，reasoning过程中依旧经常会出现异常，这些异常大部分指的是hallucination。关于幻觉的定义，学术界也在不断完善。【109】认为幻觉是generate一些违背了事实的不可靠的text。【39】认为幻觉是指response中与original prompt不相关的内容。[survay]指出幻觉的四个主要特点- compliance、desirability、relevancy和plausibility。

%总而言之，幻觉就是异常的与正常逻辑和事实相违背的想象，并且在当前LLM中无法避免。因为现有的LLM都是基于大量数据进行训练，对于训练数据很敏感，很容与出现知识遗忘的情况。并且，随着自然界的不断发展，不断有新的知识涌现，然而LLM一旦训练完成便无法改变，因此很容易出现out-of-date的情况。

% 推理异常对于当今agent system造成了严重影响，然而现有agent system 并没有针对此进行检测与处理，造成了agent system在通用任务上效果较差。

Agents use cognitive systems for reasoning, which in turn guides subsequent actions and serves as the foundation for completing complex tasks. In recent years, numerous methods have been proposed to enhance reasoning capabilities, including fine-tuning approaches such as SFT \cite{sft}, RLHF \cite{rlhf}, Search-R1 \cite{jin2025searchr1}, and DeepSeek-R1 \cite{guo2025deepseekr1}, as well as prompt engineering techniques like CoT \cite{wei2022chainofthought}, Reflexion \cite{shinn2023reflexion}, Self-Consistency \cite{wang2022selfconsistency}, CoK \cite{li2023cok}, and StepBack \cite{stepback}. Despite the support of these technologies, anomalies frequently occur during the reasoning process, most notably hallucinations.

The academic community continues to refine the definition of hallucinations. \citet{109} define hallucinations as the generation of unreliable text that contradicts known facts. \citet{39} consider hallucinations to be content in a response that is unrelated to the original prompt. \citet{hallucinationsurvey} identify four main characteristics of hallucinations: compliance, desirability, relevancy, and plausibility. \citet{yang2024alignment} describe hallucinations as a form of dishonesty, where uncertain answers are given with undue confidence. In summary, hallucinations are anomalous imaginings that contradict normal logic and facts, and are currently unavoidable in the use of LLMs. This is because LLMs are trained on large datasets, making them sensitive to training data and prone to knowledge forgetting. Furthermore, as new knowledge constantly emerges in the natural world, LLMs—once trained—cannot be updated, leading to issues with outdated information.

\subsubsection{Planning Anomalies}

% 自主的planning并调用工具完成任务是agent system的一个重要功能。然而，由于现有的LLM都是概率模型，因此会不可避免的出现异常。该类异常大多是由于planning阶段的幻觉导致的。【98】将hallucination描述为predicting an incorrect feasibility of an autonomous system when generating an explanation behind the uncertainty of an action to take.【67】发现LLM经常会给出和前面推理不一致的action。【136】和【110】认为像LLM这样的生成模型有很高的倾向性生成不合理不正确的plan。【48】则认为hallucination为 和不存在的一些事物交互，例如错误的tool或者参数。In summary，Planning Anomalies严重影响了任务的planning阶段，很容易造成任务失败，需要重点关注。

Autonomous planning and tool invocation to accomplish tasks are key functions of agent systems. However, due to the probabilistic nature of current LLMs, anomalies are inevitable. These anomalies often arise from hallucinations occurring during the planning phase. \citet{98} describe hallucination as predicting incorrect feasibility of an autonomous system when generating an explanation for the uncertainty of an action to be taken. \citet{67} observe that LLMs frequently produce actions that are inconsistent with prior reasoning. \citet{136} and \citet{110} argue that generative models like LLMs have a high propensity to generate unreasonable and incorrect plans. \citet{48} describe hallucination as involving interaction with non-existent entities, such as incorrect tools or parameters. In summary, planning anomalies significantly impact the planning phase of tasks, often causing task failures, and therefore require substantial attention.

\subsubsection{Action Anomalies}

%agent system最初的action通过function call来实现。然而由于接口不规范不统一等原因很容易造成action anomalies。Induatial 指出函数调用实践中面临诸如延迟、API 选择错误、系统故障等问题。darkside指出unction‑call 存在的“jailbreak”风险，攻击者可以通过构造特定请求令 LLM 调用敏感函数或绕过限制。MCP的出现规范了LLM和工具之间的交互方式，然而，MCP并不是万能的。在实际应用中，也经常会出现MCP server配置变更等导致action anomalies。

In agent systems, initial actions are realized through function calls. However, due to issues such as non-standard and inconsistent interfaces, action anomalies are prone to occur. \citet{industrialfunction} highlight challenges in function call practices, such as delays, incorrect API selection, and system failures. \citet{wu2024darkside} point out the ``jailbreak'' risks associated with function calls, where attackers can craft specific requests to make an LLM invoke sensitive functions or bypass restrictions. The advent of MCP has standardized the interaction between LLMs and tools. However, MCP is not a panacea; in practical applications, configurations changes in the MCP server frequently lead to action anomalies \cite{ai-infra-guard}.

\subsubsection{Memory Anomalies}

% 如上文所述，agent system的memory 分为短期记忆和长期记忆。短期记忆指的是LLM的context，即便当前LLM的context不断扩大，也无法满足任务的需求。因此很多agent framework采用滑动窗口管理context。然而这会导致丢失开始的一部分重要信息，例如任务结束说明等。即便LLM的context满足任务的需求，【lost in middle】表明LLM在长上下文中经常忽略中间段的信息。PI-LLM也表明LLM的工作记忆存在瓶颈。长期记忆则指的是存储在向量数据库的知识，一般通过RAG的方式召回。然而在召回准确率即生成合理性上都存在很多限制。QE‑RAG指处现有的RAG对于噪声极其敏感。Astute RAG指出存在内外部知识冲突的情况。benchmark指出当前RAG的准确率并不高，即便有很多技术的加持。综上所述，memory anomalies严重影响了复杂任务的有效完成。

As discussed earlier, an agent system's memory is divided into short-term and long-term memory. Short-term memory refers to the context of an LLM. Even as current LLM contexts expand, they often still fail to meet task requirements. Consequently, many agent frameworks use a sliding window to manage context, which can lead to the loss of important initial information, such as task completion instructions. Even when an LLM's context size does meet task needs, \citet{liu2023lostinmiddle} indicate that LLMs often overlook information in the middle of long contexts. PI-LLM \cite{PI-LLM} also demonstrates that LLMs have a bottleneck in working memory.

% Long-term memory refers to knowledge stored in a vector database, typically recalled using RAG. However, there are many limitations in recall accuracy and generation reliability. QE-RAG \cite{zhang2025qerag} points out that existing RAG implementations are extremely sensitive to noise, and Astute RAG \cite{wang2024astuterag} highlights the existence of conflicts between internal and external knowledge, which ultimately result in RAG hallucinations \cite{sun2024redeep}. \citet{chen2024benchmarkrag} indicate that the accuracy of current RAG systems is not high, despite support from numerous technologies.

Long-term memory refers to knowledge stored in a vector database and is typically recalled using RAG. However, there are many limitations in terms of recall accuracy and generation reliability. QE-RAG \cite{zhang2025qerag} points out that existing RAG implementations are highly sensitive to noise. Astute RAG \cite{wang2024astuterag} highlights conflicts between internal and external knowledge, which ultimately result in RAG hallucinations \cite{sun2024redeep}. \citet{chen2024benchmarkrag} indicate that the accuracy of current RAG systems is not high, despite being supported by numerous technologies.

\subsubsection{Environment Anomalies}

% 随着agent system的规模越来越大， 消耗大量的资源，尤其是agent在本地执行相关资源消耗型的操作。这可能会导致资源不足等与环境相关的anomalies产生，例如CPU 使用率过高等。因此环境相关的信息也需要额外关注。

As the scale of agent systems continues to expand, they consume substantial resources, particularly when agents execute resource-intensive operations locally. This can lead to environment-related anomalies, such as insufficient resources or excessive CPU usage.

\subsection{Inter-Agent Anomalies}

\subsubsection{Task Specification Anomalies}

%WHYfail 指出 很多任务级别的失败异常都是由于task定义不清楚例如prompt不够清晰所导致的。Emergence指出当任务定义不清楚时，很容诱发chasing 和 blocking情况，即使每个agent的单独行为是相对合理的。SentinelAgent指出任务描述或者prompt 未充分覆盖潜在协作模式时，agent 可能“走歪路”、发生 collusion 或 prompt‑injection 等未预期行为。因此pre-Excution阶段的task 描述信息完整度判断以及excution阶段的再反思是很必要的operation步骤。

\citet{whydomasfail} indicate that many task-level failures stem from unclear task definitions, such as insufficiently clear prompts. \citet{task1} highlight that when tasks are poorly defined, it easily leads to situations like chasing and blocking, even if each agent's individual actions are relatively reasonable. SentinelAgent \cite{he2025sentinelagent} notes that when task descriptions or prompts do not adequately cover potential collaboration modes, agents may deviate, engage in collusion, or exhibit unforeseen behaviors such as prompt injection. Therefore, assessing the completeness of task descriptions during the pre-execution phase and reflecting during the execution phase are essential operational steps.

% WHYFAIl 指出 agent role 配置错误是常见的异常之一。CONF指出配置不清和角色混淆是coordination failures 与 adversarial misalignment 的常见诱因。CONFBLOG 指出当 agent 执行了原本不属于自己职责的操作，直接导致冲突、不一致与效率低下。AgentFM指出角色配置模糊很容易造成的数据库Agent的失败。

Configuration is also an important part of task specification. \citet{whydomasfail} highlight that incorrect agent role configuration is one of the most common anomalies. \citet{conf1} identify unclear configurations and role confusion as frequent causes of coordination failures and adversarial misalignments. According to \citet{confblog}, when an agent performs actions that fall outside of its designated responsibilities, it directly leads to conflict, inconsistency, and inefficiency. AgentFM \cite{zhang2025agentfm} points out that vague role configuration is a common cause of failure for database agents.

\subsubsection{Security Anomalies}

\begin{figure}[ht]
    \centering
    \includegraphics[width=0.9\textwidth]{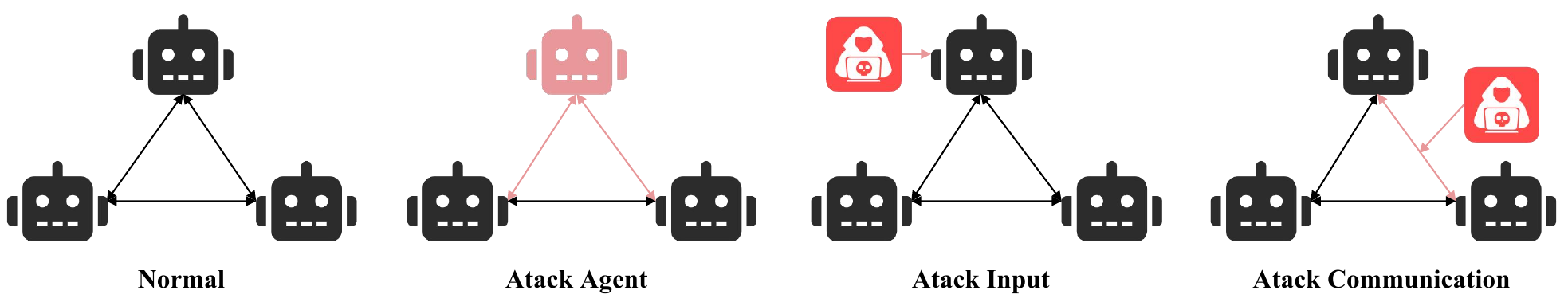}  % 图片文件名（不带扩展名）
    \caption{Different types of attacks.}
    \label{fig:attack}  % 图片标签，可用于引用
\end{figure}

Although protocols like A2A \cite{A2A} and ACP \cite{ACP} have been developed to standardize communication between agents, they only ensure that different agents can communicate using the same protocol and do not address the overall security of the protocol. \citet{communicate} note that in real-world agent systems, some agents might be maliciously attacked, causing them to frequently send requests or messages, akin to a DDoS attack. \citet{he2025redteaming} propose specific attack methods targeting agent systems, as shown in Figure \ref{fig:attack}, which can attack both the agents themselves and the communication between agents.

\subsubsection{Communication Anomalies}

% Blog指出，Communication Anomalies经常发生在agents之间change messages期间，具体表现为过量message导致的message storm从而传输失败或者传输延迟，或者network issue导致无法传输。【2】指出，真实的agent系统中可能某些agent被恶意攻击，从而频繁发出请求发送消息，类似于DDos攻击。同时还可能出现一些连锁反应导致的循环communicate，最终导致资源泛滥。尽管A2A，ACP等协议出现，规范了agent之间的通讯,然而，其仅仅保证了不同agent之间能用同样的协议进行沟通，却没有对于整个协议的安全进行考虑。

As highlighted by \citet{blog}, communication anomalies frequently occur during message exchanges between agents. These are specifically manifested as message storms, resulting from excessive messaging, which can lead to resource exhaustion and increased latency, ultimately causing task failure. AgentPrune \cite{zhang2024agentprune} also identifies the issue of message redundancy, noting that an excessive number of messages does not enhance the efficiency of agent systems. Instead, the redundancy causes agents to become lost.

\subsubsection{Trust Anomalies}

%A-Trust 指出LLM agents将所有到来的消息视为平等的。Trust指出llm agent会接受其他agent消息并加入上下文作为一部分without 一致性验证 and 不考虑是否值得信任。然而，不同agent的backbone可能存在差异，由于memory的等的差别，针对特定领域的能力也存在很大不同。例如一个 code agent的代码编程能力明显强于普通的agent。因此，不同agent的消息不能一视同仁，轻易的相信所有agent的话，会导致信息冲突或错误。然而如果不相信任何agent的话，则可能导致整个系统没有协作能力。因此agents之间的信任问题是一个亟待解决的严重问题，严重影响了合作效率。

ATrust \cite{atrust} points out that LLM agents treat all incoming messages equally. \citet{trust} highlight that LLM agents accept messages from other agents and incorporate them into their context without consistency verification, and without considering whether those messages are trustworthy. However, the foundational models of different agents may vary, and due to differences in memory and other factors, their capabilities in specific domains can also differ significantly. For example, a code agent's programming abilities are markedly stronger than those of a general-purpose agent. Therefore, messages from different agents should not be regarded uniformly. Blindly trusting all agents' messages can lead to information conflicts or errors. Conversely, if no agents are trusted, the system may lack collaborative capabilities. Thus, the issue of trust between agents is a critical problem that urgently needs addressing, as it significantly affects cooperation efficiency.

% \subsubsection{Information Anomalies}

% %INfomration 指出不同agent之间传递的消息可能存在很多冗余噪声，可能干扰沟通的效率。TMC指出当今的agent system之间的沟通机制经常需要在一个可信任信道下同一时间传输大量的消息，这严重影响了真实环境中的应用。因此，过多的信息或者过少的信息都可能造成agent system之间的无法有效沟通，从而导致最终任务的异常

% \citet{information} highlight that messages exchanged between different agents can contain considerable redundant noise, potentially disrupting communication efficiency. TMC \cite{TMC} notes that current communication mechanisms within agent systems often require transmitting a large volume of messages simultaneously through a trusted channel, which significantly hampers their application in real-world environments. Consequently, either an excess or a deficiency of information can render agent systems unable to communicate effectively, leading to anomalies in task execution.

\subsubsection{Emergent Behavioral Anomalies}

%BLOG认为Emergent Behavioraloccurs when the interactions of multiple agents give rise to macroscopic patterns or behaviors that are not predictable or easily explained by analyzing the agents in isolation.BLOG2认为Emergent behavioral anomalies arise from complex interactions between multiple agents, creating system-level behaviors that cannot be attributed to any single agent。因此Emergent Behavioral Anomalies是一类相对较意识到的异常，但是却会造成严重的后果。

\citet{blogemergent} suggests that emergent behavior occurs when interactions among multiple agents give rise to macroscopic patterns or behaviors that are not predictable or easily explained when the agents are analyzed in isolation. \citet{blog} posits that emergent behavioral anomalies arise from complex interactions between multiple agents, resulting in system-level behaviors that cannot be attributed to any single agent. As such, emergent behavioral anomalies are a relatively less understood class of anomalies, yet they can lead to severe consequences.

\subsubsection{Termination Anomalies}

% WhyFail 和microsoft 都把Premature-Stop Anomalies作为agent system异常的重要一类。Smurfs指出DFSDT 在单 agent 模式下容易出现过早结束问题，即系统过快调用 termination tool，而未完成多步推理。这一问题会严重影响复杂任务的完成度与逻辑连贯性。When&WHo 指出premature termination 是 multi-agent system 中常见且细粒度可定位的失败来源。

Both \citet{whydomasfail} and \citet{Microsoft} consider premature-stop anomalies as a significant category of anomalies in agent systems. Smurfs \cite{chen2025smurfs} points out that in single-agent modes, Deep First Search Decision Tree (DFSDT) often faces issues with premature termination, where the system too quickly invokes the termination tool without completing multi-step reasoning. This problem can severely impact the completeness and logical coherence of complex tasks. \citet{icml2025} identify premature termination as a common and precisely locatable source of failure in multi-agent systems.

% Rucur1指出agent system 会出现undercommitment 异常，即智能体不断交给子智能体，形成像链条一样的无限递归，最终命中递归层数限制或超时。RECUR2提出了架空神经反馈”（neural howlround）的概念——智能体陷入无限递归或无限自我优化循环，永无终点。论文分析了这种现象会导致“perseverative thinking”——递归循环永不停歇，使系统停顿或“认知停滞”。
\citet{recur} identify undercommitment anomalies in agent systems, where an agent continuously delegates tasks to sub-agents, creating an infinite recursion chain that ultimately hits recursion depth limits or timeouts. \citet{recur2} introduces the concept of ``neural howlround,'' where agents become trapped in infinite recursion or self-optimization loops with no endpoint. The paper analyzes how this phenomenon can lead to ``perseverative thinking''—an unending recursive loop that causes the system to stall or experience ``cognitive stagnation.''

%% file: agent_operations.tex
% 本章中我们将介绍AgentOps的起源，定义以及具体涵盖的内容。首先，表1中介绍了一些下面将会用到的概念。
% In this section, we will explore the origins, definition, and specific scope of AgentOps. To begin, Table \ref{tab:definition} presents some concepts that will be used in the following discussion.

Section \ref{anomalies} identifies a variety of anomalies within current agent systems, highlighting the need for their operations and maintenance. In this section, we will delve into the origins, definition, and specific scope of AgentOps. We begin by tracing the evolutionary history of operations, which naturally leads to the introduction of the AgentOps concept. Subsequently, we outline the main differences between AgentOps and traditional operations across various stages, explaining why conventional operations fail to address the challenges of agent systems. Finally, we provide a precise definition of AgentOps. To start, Table \ref{tab:definition} presents key concepts that will be referenced throughout the discussion.

\begin{table}
\centering
\caption{Definition of some symbols in the paper.}
\label{tab:definition}
\footnotesize
{
\begin{tabular}{>{\raggedright\arraybackslash}m{1.8cm} p{10cm}}
\toprule
\textbf{Name}     & \multicolumn{1}{c}{\textbf{Definition}}                                                                                                                                                           \\ 
\midrule
\textbf{DevOps}     & DevOps merges software development and IT operations to shorten the development lifecycle and deliver high-quality software.
     \\
\midrule
\textbf{AIOps}    & AIOps uses machine learning to automate IT operations.                                                              \\
\midrule
\textbf{MLOps}    & MLOps refers to the application of various operational techniques for the maintenance and management of machine learning models. \\
\midrule
\textbf{AgenticOps} & Agentic Operations (AgenticOps) refers to the use of agents for the operation and maintenance of traditional systems. \\
\midrule
\textbf{AgentOps} & AgentOps refers to the use of various operational techniques for the management and maintenance of agent systems.  \\
\bottomrule
\end{tabular}
}
\end{table}

\subsection{Evolution of Operations}

\begin{figure}[t]
    \centering
    \includegraphics[width=0.6\textwidth]{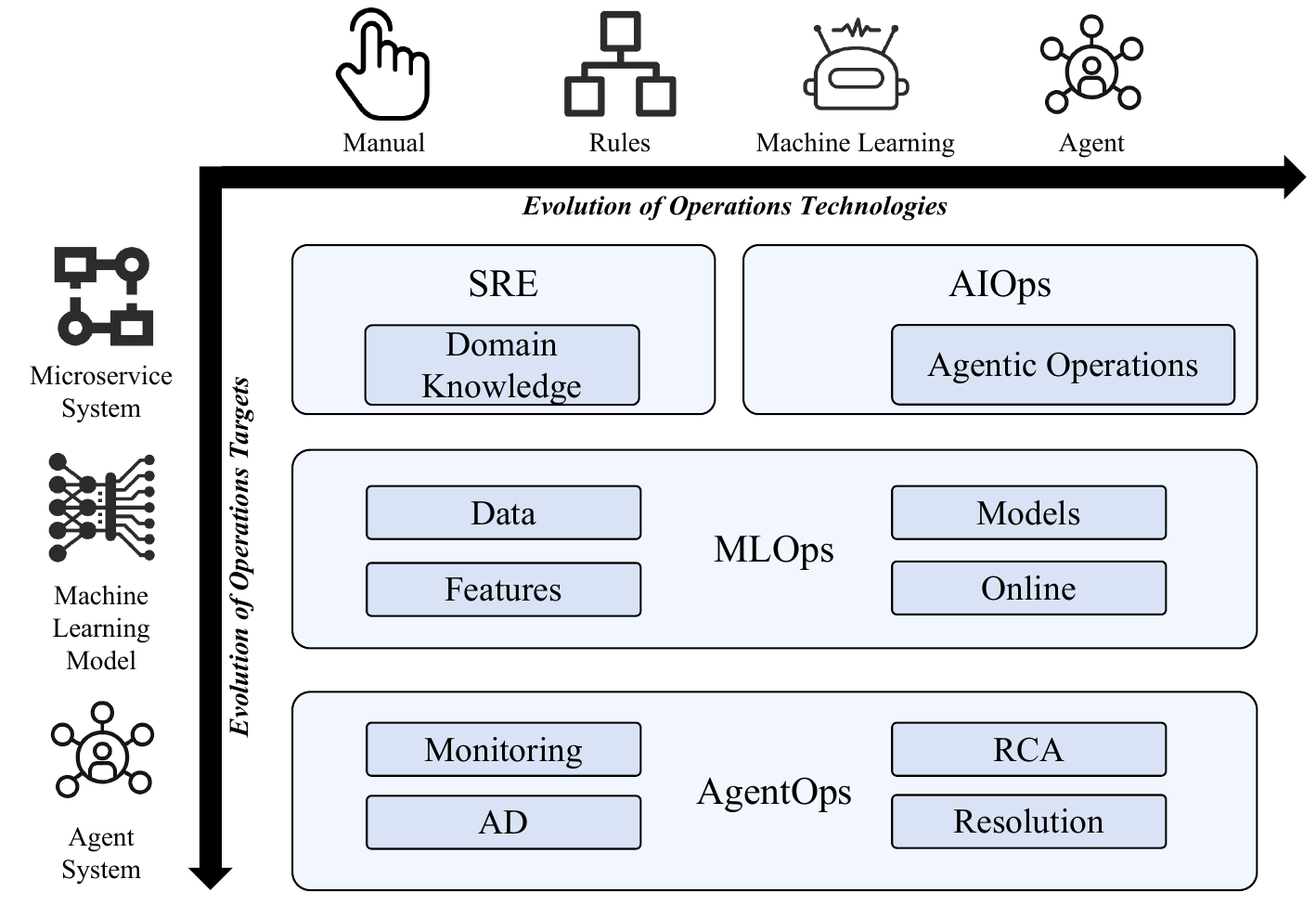}  % 图片文件名（不带扩展名）
    \caption{Evolution of operations. The horizontal axis illustrates the technological progression of operations practices, evolving from manual operations to machine learning-driven and agent-based automation. The vertical axis reflects the shift in operational targets, transitioning from traditional microservice systems to machine learning models, and more recently to complex agent systems.}
    \label{fig:agentopsandopsagent}  % 图片标签，可用于引用
\end{figure}

As shown in Figure \ref{fig:agentopsandopsagent}, operational technologies have continually advanced over time. From the early days of manual operations to the later rule-based automated operations, each phase has marked significant progress. The rapid development of machine learning, and deep learning in particular, has further propelled advancements in operational technologies. In recent years, the powerful reasoning and analytical capabilities of LLMs have led to an increasing number of methods utilizing LLM agents for automated operations.

In addition to the evolution of operational technologies, the objects of operation have also transformed. Initially, operations were focused solely on traditional hardware and software systems, like microservice systems. With the rapid advancement of machine learning, models have grown in size, and faults have become more frequent in processes such as training and inference. This necessitated the application of various operational techniques to manage machine learning models, known as MLOps. In recent years, LLM agents have become capable of autonomously completing tasks, leading to many services adopting agents as replacements. As a result, agent systems have become the mainstream method for service construction in the industry today. However, agent systems differ significantly from traditional systems. For example, while agents exhibit stochasticity in their reasoning and actions due to their probabilistic model foundations, traditional systems feature deterministic behaviors dictated by underlying code. Thus, the operations involved in managing agent systems differ vastly from traditional systems, making it impractical to directly apply traditional techniques to agent systems. This is why we propose \textbf{AgentOps}. Below, we will elaborate on the differences between traditional system operations and AgentOps.

\subsection{Difference between Traditional System Operations and Agent System Operations}

\input{operations}

\section{Monitoring Agent System}
\label{sec:monitor}

\subsection{Illustration of Monitoring Data}

\subsubsection{Traditional Data}

% 传统的monitoring data是在微服务system和agent system中都存在的基于OpenTelemetry采集的metric 、log 、trace。然而正如上面所说，agent system这些data和microservice system中区别很大。

%首先metric 层面，如表1所示，对于传统的microservice system，只需要监控system metric 和 APM metric。然而agent system引入了LLM agent，因此在system metric中会额外加入一些与LLM和agent 相关的metric，例如LM latency，tool call latency。此外，由于成本是agent system中至关重要的一环，因此会引入很多与cost 相关的metric ，例如token消耗数目。此外，在当前垂直领域场景下应用agent system，RAG是必不可少的一环，因此RAG 相关的metric也需要纳入。

% trace层面，如图1所示，microservice的trace指的是不同service之间通过API相互调用。API调用的参数是通过用户的行为或者系统实现定义好的规则确定的，相对来说较为确定，并且参数较为简洁。然而对于agent system，每个agent的输入和输出，以及agent之间的调用，agent调用tool大都由LLM生成实现，存在很大的不确定性。这些不确定是trace的关键内容，因此在agent system 中，trace除了包含agent、tool的调用关系，还包括了每个步骤的输入和输出。因此trace是agent system中最重要的数据。

% log层面，如图1所示，microservice system和agent system较为相似。microservice system的log从宏观层面记录了service的行为，agent system的log则是记录了agent 的行为。

In traditional monitoring data, metrics, logs, and traces collected using OpenTelemetry \cite{opentelemetry2019} exist within both microservice systems and agent systems. However, as mentioned earlier, the data in agent systems differs significantly from that in microservice systems.

Firstly, with regard to metrics, as shown in Table \ref{tab:metric}, traditional microservice systems usually focus on monitoring system metrics and Application Performance Monitoring (APM) metrics. However, in the context of agent systems, which incorporate LLM agents, additional metrics related to LLMs and agents are introduced, such as LLM latency and tool call latency. Moreover, since cost is a critical factor in agent systems, various cost-related metrics are also included, like the number of tokens consumed. In current vertical industry applications of agent systems, RAG is indispensable, so RAG-related metrics are necessary as well.

\begin{figure}[htbp]
    \centering
    \includegraphics[width=\textwidth]{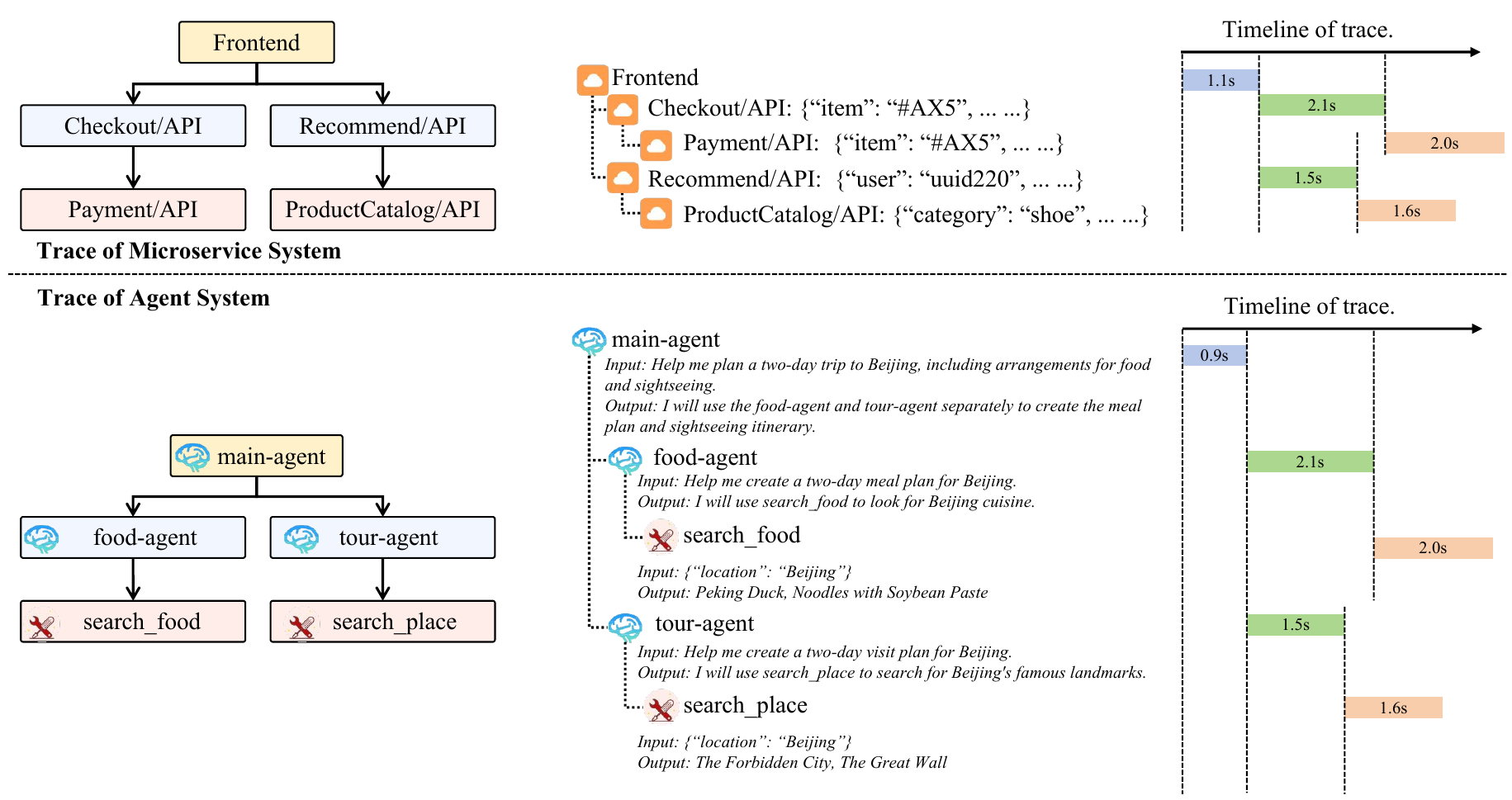}  % 图片文件名（不带扩展名）
    \caption{Comparison of trace data in microservice system and agent system. In the microservice system, trace data captures a user's journey through purchasing on an e-commerce platform. Meanwhile, in the agent system, food and tour agents team up to craft a personalized two-day travel plan for the user.}
    \label{fig:trace}  % 图片标签，可用于引用
\end{figure}

Regarding traces, as illustrated in Figure \ref{fig:trace}, microservice traces typically refer to interactions between services through API calls. The parameters for these API calls are determined by user actions or predefined system rules, making them relatively stable and straightforward. However, in agent systems, the inputs and outputs of each agent and the interactions between agents, including agent-to-tool calls, are often generated by LLMs, introducing a high degree of uncertainty. These uncertainties are key elements of the trace data; hence, in agent systems, a trace encompasses not only the relationships among agents and tools but also the inputs and outputs at each step. Therefore, trace data is the most crucial aspect of an agent system.

\begin{figure}[htbp]
    \centering
    \includegraphics[width=0.7\textwidth]{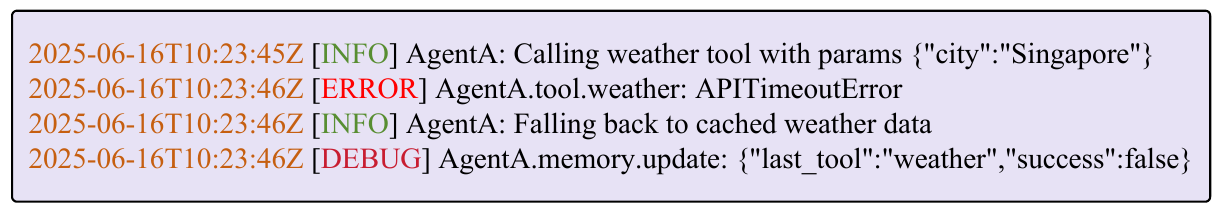}  % 图片文件名（不带扩展名）
    \caption{Example of log data.}
    \label{fig:log}  % 图片标签，可用于引用
\end{figure}

When it comes to logs, as shown in Figure \ref{fig:log}, microservice systems and agent systems are quite similar. The logs in a microservice system record the overall behavior of services, whereas in an agent system, they capture the behavior of agents.

\begin{table}
\centering
\caption{Comparison of metric types in agent system and microservice system. System metrics pertain to indicators like latency and are concerned with system-related performance. Cost metrics address resource consumption indicators, such as the number of tokens used. RAG metrics are specialized for RAG systems and include measures like recall precision. Application Performance Monitoring (APM) focuses on the behavior and performance of applications, tracking metrics like success rate.}
\label{tab:metric}
\resizebox{\textwidth}{!}{
\begin{tabular}{ccl} 
\toprule
\textbf{System Type}                          & \textbf{Metric Type} & \multicolumn{1}{c}{\textbf{Examples}
}                                                       \\ 
\midrule
\multirow{4}{*}{\textbf{Agent System}}        & System Metric        & Latency per Tool Call, Total Task Completion Time, API Call Frequency, LLM Call Latency  
\\
                                              & Cost Metric          & Total Token Count, API Call Cost per Task, Cost per Task Completion                         
\\
                                              & RAG Metric           & Chunk Precision, Document Recall, Mean Average Precision, Mean Reciprocal Rank              
\\
                                              & APM Metric   & Success Rate, Steps per Task, Tool Call Accuracy                                            
\\ 
\midrule
\multirow{2}{*}{\textbf{Microservice System}} & System Metric         & CPU Usage, Memory Usage, Network RTT                                                        
\\
                                              & APM Metric           & Requests Number,~Click‑Through Rate, Conversion Rate                                        \\
\bottomrule
\end{tabular}
}
\end{table}

\subsubsection{Model Data}

% 随着数据安全问题的影响不断扩大，越来越多的线上agent system使用本地部署的开源LLM作为推理引擎。如果单纯将LLM当作黑盒模型，不关注LLM内部的状态，那么我们能拿到的只有LLM的输入和输出，这些信息相当有限，不足以检测LLM内部发生的异常。因此越来越多的方法尝试将LLM 视为白盒模型，收集LLM内部的隐藏层的参数，以及token logits。

As data security concerns continue to grow, an increasing number of online agent systems are opting to use locally deployed open-source LLMs as their inference engines \cite{datasecurity}. When treating an LLM purely as a black-box model without regard for its internal state, we are limited to observing only its inputs and outputs. This information is quite restricted and insufficient for detecting anomalies occurring within the LLM itself. Consequently, more and more approaches are emerging that treat the LLM as a white-box model, aiming to collect internal parameters from its hidden layers and token logits \cite{SAPLMA,huang2024opera}.

\subsubsection{Checkpoint Data}

\begin{figure}[htbp]
    \centering
    \includegraphics[width=0.6\textwidth]{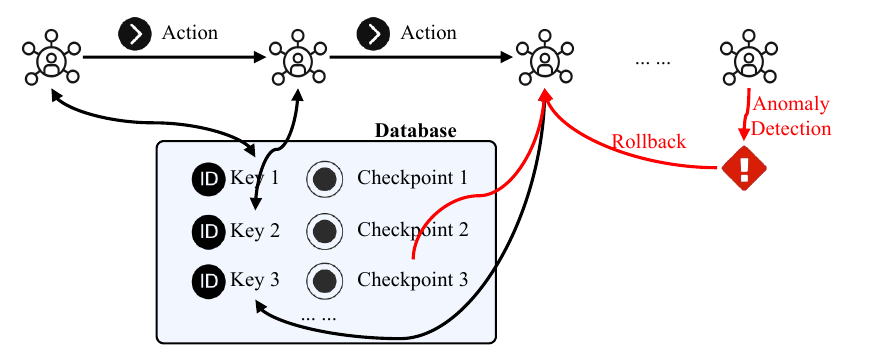}  % 图片文件名（不带扩展名）
    \caption{Collection of checkpoint data and its use in rollback processes.}
    \label{fig:checkpoint}  % 图片标签，可用于引用
\end{figure}

% agent system相比microservice system相对可控，可以通过数据复现某个时刻的状态。这给予agent system operations很大的优势。因此，我们可以记录每个时刻agent system的状态，包括memory environment等checkpint信息。后续一旦发生故障，我们可以通过这些checkpoint数据退回到之前的状态，修复问题，并最终得到正确的结果。

Compared to microservice systems, agent systems offer greater control, allowing the state at any given moment to be reproduced from data. This provides a significant advantage for operations within agent systems. As shown in Figure \ref{fig:checkpoint}, we can record the state of the agent system at various times, including checkpoint information such as the memory environment. In the event of a failure, these checkpoint data enable us to roll back to a previous state, resolve the issue, and ultimately arrive at the correct outcome.

\subsection{Current Monitoring Methods}

\begin{table}
\centering
\caption{Features of different monitoring methods.}
\label{tab:observe}
\resizebox{\textwidth}{!}{
\begin{tabular}{ccccccc} 
\toprule
\multicolumn{1}{l}{}   & \multicolumn{4}{c}{\textbf{Metric}}                                                                                & \multirow{2}{*}{\textbf{Log}} & \multirow{2}{*}{\textbf{Trace}}  \\ 
\cmidrule{2-5}
\textbf{Tools}         & \textbf{System Metric}     & \textbf{Cost Metric}       & \textbf{RAG Metric}        & \textbf{Performance Metric} &                               &                                  \\ 
\midrule
\textbf{LangDB \cite{langdb2025}}        & \ding{51} & \ding{51} & \ding{55} & \ding{55}  & \ding{51}    & \ding{51}       \\
\textbf{LangFuse \cite{langfuse2025}}      & \ding{51} & \ding{51} & \ding{51} & \ding{51}  & \ding{51}    & \ding{51}       \\
\textbf{MLFlow \cite{mlflow2025}}        & \ding{55} & \ding{51} & \ding{55} & \ding{51}  & \ding{51}    & \ding{51}       \\
\textbf{Helicone \cite{helicone2025}}      & \ding{51} & \ding{51} & \ding{55} & \ding{51}  & \ding{51}    & \ding{51}       \\
\textbf{LangWatch \cite{langwatch2025}}     & \ding{51} & \ding{51} & \ding{55} & \ding{51}  & \ding{51}    & \ding{51}       \\
\textbf{LlamaTrace \cite{arize_llamatrace2025}}    & \ding{51} & \ding{51} & \ding{55} & \ding{51}  & \ding{51}    & \ding{51}       \\
\textbf{OpenLLMetry \cite{openllmetry2025}}   & \ding{51} & \ding{55} & \ding{55} & \ding{51}  & \ding{51}    & \ding{51}       \\
\textbf{Arize Phoenix \cite{arize_phoenix2025}} & \ding{51} & \ding{51} & \ding{55} & \ding{51}  & \ding{51}    & \ding{51}       \\
\textbf{Literal AI \cite{literalai2025}}    & \ding{51} & \ding{51} & \ding{55} & \ding{51}  & \ding{51}    & \ding{51}       \\
\textbf{Opik \cite{comet_opik2025}}          & \ding{51} & \ding{51} & \ding{55} & \ding{51}  & \ding{51}    & \ding{51}       \\
\textbf{OpenInference \cite{arize_openinference2025}} & \ding{51} & \ding{55} & \ding{55} & \ding{51}  & \ding{51}    & \ding{51}       \\
\textbf{TruLens \cite{trulens2025}}       & \ding{51} & \ding{51} & \ding{55} & \ding{51}  & \ding{51}    & \ding{51}       \\
\textbf{HoneyHive \cite{honeyhive2025}}     & \ding{51} & \ding{51} & \ding{55} & \ding{51}  & \ding{51}    & \ding{51}       \\
\textbf{PromptLayer \cite{promptlayer2025}}   & \ding{51} & \ding{51} & \ding{55} & \ding{51}  & \ding{51}    & \ding{51}       \\
\textbf{OpenLIT \cite{openlit2025}}       & \ding{51} & \ding{51} & \ding{55} & \ding{51}  & \ding{51}    & \ding{51}       \\
\textbf{AgentOps \cite{agentops}}      & \ding{51} & \ding{51} & \ding{55} & \ding{51}  & \ding{51}    & \ding{51}       \\
\textbf{DeepEval \cite{DeepEval}}      & \ding{51} & \ding{51} & \ding{55} & \ding{51}  & \ding{51}    & \ding{51}       \\
\bottomrule
\end{tabular}
}
\end{table}

Observability tools for LLM-based agent systems are also rapidly advancing. These tools predominantly adhere to the aforementioned principles of collecting metrics, logs, and trace data. Most agent system observability tools integrate functions such as trace, metrics, datasets, experiments, evaluations, prompt optimization, and management. Table \ref{tab:observe} provides a comparison of the functionalities of different tools. LangDB \cite{langdb2025}, the first observability tool entirely developed in Rust, offers higher efficiency and internal integration with router optimization for cost control. Langfuse \cite{langfuse2025} supports OpenTelemetry integration and is the most active observability tool in the open-source community. Helicone \cite{helicone2025}, in addition to its observability tools, integrates cache management to reduce latency and save resources, ensuring the security and scalability of agent systems through mechanisms like gateway fallback. HoneyHive \cite{honeyhive2025} employs distributed tracing and effectively handles multi-modal systems, allowing for custom spans to focus on specific system aspects. PromptLayer \cite{promptlayer2025}, initially designed for prompt optimization, includes observability as a necessary component for task completion, encompassing features like prompt ranking and scoring. TruLens \cite{trulens2025}, as a Python package, seamlessly integrates with various frameworks like LLamaIndex and iteratively optimizes agent systems using human feedback efficiently. OpenLLMetry \cite{openllmetry2025}, by adhering to the OpenTelemetry standard, offers compatibility with diverse frameworks but lacks features such as prompt optimization and evaluation testing. LangWatch \cite{langwatch2025} and Literal AI \cite{literalai2025} serve as standard observability tools, featuring observability evaluation and development functions, with LangWatch already integrated as an MCP server. MLFlow \cite{mlflow2025}, originating from traditional deep learning, supports custom metrics in agent systems. DeepEval \cite{DeepEval} primarily focuses on evaluations and lacks observability features, while AgentOps \cite{agentops} emphasizes operational oversight of the entire system alongside observability functionalities.

% 大多数agent system观测系统都集成了trace、metric、dataset、experiment、评估、prompt 优化管理等功能。表1是不同工具的功能对比。LangDB是第一个完全基于Rust开发的可观测性工具，有更高的效率。同时其内部集成了router优化控制成本。Langfuse支持OpenTelemetry集成，是open-source社区最活跃的可观测性工具。Helicone除了可观测性工具外，还集成了cahe管理来降低延迟和节省消耗，同时通过gateway fallback等方式保证agent system的安全和可扩展。HoneyHive采用分布式traceing，并且能有效处理多模态系统。其还可以通过custom spans来额外关注系统的某些部分。PromptLayer最初是为了优化prompt而设计的，可观测性只是为了完成该任务而必须加入的，其还包含prompt的ranking、scoreing等很多功能。TruLens作为python包，可以轻易集成进大多数框架如LLamaIndex，同时其有效利用human feedback迭代优化agent systems。OpenLLMetry 通过OpenTelemetry 标准接入，兼容各种框架，但是缺乏prompt优化、测试评估等各项功能。LangWatch和Literal AI则是标准的可观测性工具，同时具有observalibity evaluation和development等功能，LangWatch已经集成为MCP server。MLFlow源自于传统的深度学习，在agent systems中支持自定义指标。DeepEval主要用来做评估，在可观测性方面有所欠缺。AgentOps则注重在可观测之余，对于整个system进行operation。

\subsection{Challenges}

%然而，即便现在有很多agent system的可观测性工具，然而，现在依旧存在很多挑战。（1）海量的数据：随着agent system的不断扩展，agent的数量越来越多，每个agent都会产生大量的日志、调用记录和通讯数据。这些大量的数据收集、存储、分析存在很多挑战。（2）监控数据种类缺失：相比传统的微服务系统，对于agent system，log、trace、metric远远不够。还需要监控内存、环境的变化，这些都是agent的重要组成部分。这些种类的数据在之后resolution阶段的回滚中有重要作用。（3）安全漏洞：智能体可以自主调用工具，这些工具可能会对于memory进行一定的改写，如果不加以监控并报警，很容易造成数据泄漏等损失。因此，monitoring agent还存在很长的路需要走。

Despite the availability of numerous observability tools for agent systems, challenges still persist. 
\begin{itemize}
    \item Vast Amounts of Data: As agent systems continue to grow, the number of agents increases, and each agent generates a substantial volume of data. This challenge is particularly pronounced with the addition of model data and checkpoint data. The collection, storage, and analysis of this massive dataset present significant challenges.
    \item Lack of Diverse Monitoring Data: As mentioned earlier, compared to traditional microservice systems, agent systems suffer from insufficient logs, traces, and metrics.
    \item Security Vulnerabilities: Agents can autonomously invoke tools that may modify memory. Without proper monitoring and alerts, this can easily result in data leaks and other losses. Therefore, there is still a long way to go in improving agent monitoring.
\end{itemize}

\section{Anomaly Detection \& Mitigation}

A summary of anomaly detection and mitigation in agent systems is illustrated in Table \ref{tab:anomaly-detection-prevention}.

\begin{figure}[t]
    \centering
    \includegraphics[width=0.5\textwidth]{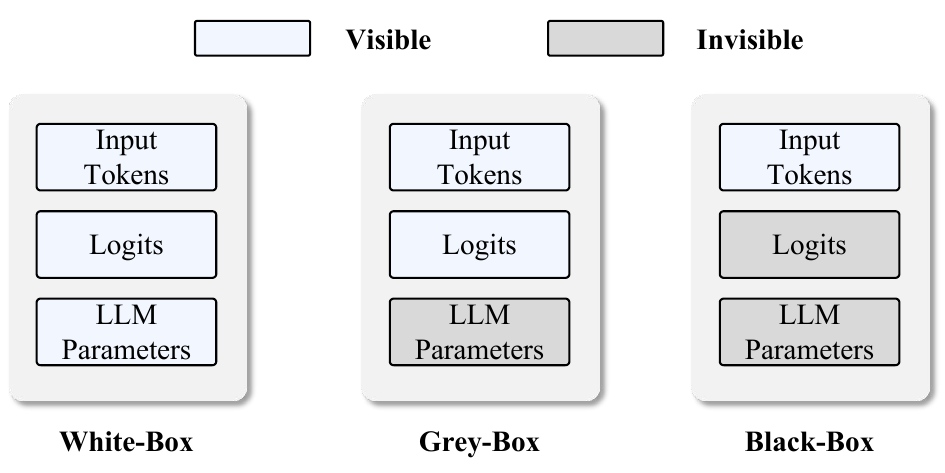}  % 图片文件名（不带扩展名）
    \caption{Illustration of white-box, grey-box and black-box.}
    \label{fig:reasonano}  % 图片标签，可用于引用
\end{figure}

\begin{table}
\centering
\caption{A Summary of Anomaly Detection \& Mitigation Methods.}
\label{tab:anomaly-detection-prevention}
\resizebox{\textwidth}{!}{
\begin{tabular}{cccccccc} 
\toprule
\textbf{Anomaly Level}                   & \textbf{Anomaly Type}                                                        & \textbf{Method Category}           & \textbf{Method}                                   & \textbf{Input}              & \textbf{Output}               & \textbf{Detection}                             & \textbf{Mitigation}                             \\ 
\midrule
\multirow{21}{*}{\textbf{ Intra-Agent }} & \multirow{7}{*}{\textbf{Reasoning Anomalies}}                                & \multirow{3}{*}{White-box}         & SAPLMA \cite{SAPLMA}             & LLM Parameters              & Anomaly Probability           & \ding{51}                     & \ding{55}                      \\
                                         &                                                                              &                                    & OPERA \cite{huang2024opera}      & Attention Map               & Penalty  Regenerated Response & \ding{51}                     & \ding{51}                      \\
                                         &                                                                              &                                    & Honesty \cite{yang2024alignment} & LLM Parameters (Finetuning) & Revised Response              & \ding{51}                     & \ding{51}                      \\ 
\cmidrule{3-8}
                                         &                                                                              & \multirow{2}{*}{Grey-box}          & LURE \cite{zhou202LURE}          & Token Logits, Revisor Model & Revised Response              & \ding{51}                     & \ding{51}                      \\
                                         &                                                                              &                                    & Conformal                                         & Token Logits                & High Quality Response         & \ding{51}                     & \ding{51}                      \\ 
\cmidrule{3-8}
                                         &                                                                              & \multirow{2}{*}{Black-box}         & Debate                                            & Multi-Agents                & Response after Debate         & \ding{51}                     & \ding{51}                      \\
                                         &                                                                              &                                    & CoK                                               & Multiple Data Sources       & High Quality Response         & \ding{51}                     & \ding{51}                      \\ 
\cmidrule{2-8}
                                         & \multirow{6}{*}{\textbf{Planning Anomalies}}                                 & \multirow{3}{*}{Single-step}       & Introspective                                     & Token Logits                & High Quality Plan             & \ding{51} & \ding{51} \\
                                         &                                                                              &                                    & API-bank                                          & LLM Parameters (Finetuning) & High Quality Plan             & \ding{55}                     & \ding{51}                      \\
                                         &                                                                              &                                    & ReAct                                             & Prompt                      & High Quality Plan             & \ding{51}                     & \ding{51}                      \\ 
\cmidrule{3-8}
                                         &                                                                              & \multirow{3}{*}{Multiple-step}     & Toolllm                                           & LLM Parameters (Finetuning) & High Quality Plan             & \ding{55}                     & \ding{51}                      \\
                                         &                                                                              &                                    & Reflexion                                         & External Feedback           & High Quality Plan             & \ding{51}                     & \ding{51}                      \\
                                         &                                                                              &                                    & CodeAct                                           & Tool Description            & Code                          & \ding{55}                     & \ding{51}                      \\ 
\cmidrule{2-8}
                                         & \multirow{3}{*}{\textbf{Action Anomalies}}                                   & Function Call                      & -                                                 & -                           & -                             & -                                              & -                                               \\ 
\cmidrule{3-8}
                                         &                                                                              & \multirow{2}{*}{MCP}               & AI-Infra-Guard                                    & AI Component                & MCP Risks                     & \ding{51}                     & \ding{55}                      \\
                                         &                                                                              &                                    & MCP Guardian                                      & AI Component                & MCP Risks                     & \ding{51}                     & \ding{55}                      \\ 
\cmidrule{2-8}
                                         & \multirow{4}{*}{\textbf{Memory Anomalies}}                                   & \multirow{2}{*}{Short Memory}      & PI                                                & Longer Context              & Response                      & \ding{55}                     & \ding{51}                      \\
                                         &                                                                              &                                    & CoA                                               & Longer Context              & Response                      & \ding{55}                     & \ding{51}                      \\ 
\cmidrule{3-8}
                                         &                                                                              & \multirow{2}{*}{Long Memory (RAG)} & ReDeep                                            & Attention, LLM Parameters   & Hallucination Results         & \ding{51}                     & \ding{51}                      \\
                                         &                                                                              &                                    & LRP4RAG                                           & Token Logits                & Anomaly Probability           & \ding{51}                     & \ding{55}                      \\ 
\cmidrule{2-8}
                                         & \textbf{Environment Anomalies}                                               & -                                  & -                                                 & -                           & -                             & -                                              & -                                               \\ 
\midrule
\multirow{8}{*}{\textbf{Inter-Agent}}   & \textbf{Task Specification Anomalies}                                        & -                                  & -                                                 & -                           & -                             & -                                              & -                                               \\ 
\cmidrule{2-8}
                                         & \multirow{2}{*}{\textbf{\textbf{Security Anomalies}}}                        & \multirow{2}{*}{Graph}             & GUARDIAN                                          & Agent Graph                 & Anomalous Position            & \ding{51}                     & \ding{55}                      \\
                                         &                                                                              &                                    & SentinelAgent                                     & Agent Graph                 & Anomalous Position            & \ding{51}                     & \ding{55}                      \\ 
\cmidrule{2-8}
                                         & \multirow{2}{*}{\textbf{\textbf{\textbf{\textbf{Communication Anomalies}}}}} & \multirow{2}{*}{Redundancy}        & AgentPrune                                        & Agent Graph                 & New Agent Graph               & \ding{51}                     & \ding{51}                      \\
                                         &                                                                              &                                    & G-Designer                                        & Agent Graph                 & New Agent Graph               & \ding{51}                     & \ding{51}                      \\ 
\cmidrule{2-8}
                                         & \textbf{Trust Anomalies}                                                     & -                                  & ATrust                                            & Attention Map               & Trustworthiness               & \ding{51}                     & \ding{51}                      \\ 
\cmidrule{2-8}
                                         & \textbf{Emergent Behavioral Anomalies}                                       & -                                  & -                                                 & -                           & -                             & -                                              & -                                               \\ 
\cmidrule{2-8}
                                         & \textbf{\textbf{Termination Anomalies}}                                      & -                                  & -                                                 & -                           & -                             & -                                              & -                                               \\
\bottomrule
\end{tabular}
}
\end{table}

\subsection{Reasoning Anomalies}

% reasoning anomalies的detection和mitigation方法基于输入信息的种类可以分为white-box，grey-box和black-box。white-box指使用token序列，每个token的相关概率和模型参数。grey-box则不使用模型参数。black-box只使用token序列。

The methods for detecting and mitigating reasoning anomalies can be categorized into white-box, grey-box, and black-box based on the type of input information used. White-box approaches utilize the token sequence, the associated probabilities of each token, and the model parameters. In contrast, grey-box methods do not use model parameters. Lastly, black-box techniques rely solely on the token sequence.

\subsubsection{White-box}

% \subsubsection{SAPLMA}

% SPALMA认为LLM具有独立自主判断之前的state是否正确的能力，然而即便具有这个能力，也不代表LLM不会出现幻觉。原因是一方面LLM是一个token预测器，其在每一个token预测时尽可能的保证预测概率相对大的token，但并不能保证整个token序列的概率最大。另一方面，LLM基于概率采样，而并不是直接输出最大概率。因此SPALMA认为LLM内部的参数信息能一定程序反应幻觉，提出基于LLM参数输入的分类器。其使用大量的来自不同topic的数据训练分类器，使得其具有zero-shot幻觉检测的能力。

SPALMA \cite{SAPLMA} posits that LLMs possess the ability to independently assess whether a previous state is accurate. However, even with this capability, LLMs are not immune to generating hallucinations. This is because LLMs function as token predictors, striving to predict tokens with relatively high probability at each step, but not necessarily ensuring that the entire token sequence has the highest probability. Additionally, LLMs employ probabilistic sampling rather than directly outputting the token with the highest probability. Consequently, SPALMA suggests that the internal parameters of LLMs can, to some extent, reflect hallucinations. To address this, they propose a classifier based on LLM parameters. This classifier is trained on a large dataset comprising various topics, enabling it to detect hallucinations in a zero-shot manner.

% \subsubsection{OPERA}

% OPERA认为现有的方法如SPALMA需要额外的数据模型进行训练，因此从inference阶段来解决幻觉，不引入额外的知识。其motivation 源自于发现了self-attention map的“ partial over-trust ”现象。具体来说，在一些summary token的附近的attention score 会呈现出柱状结构，称之为啊aggregation pattern。并且对于越长的context，就会出现越多的summary token，就越容易出现幻觉。因此OPERA提出了一个column-wise metric来判别这些 aggregation patterns，并且基于此设计penalty来改进token生成的概率，降低幻觉现象。同时提出一个回滚策略，如果根据分数判断发生了幻觉，则回滚到summary token处重新生成token 序列。

OPERA \cite{huang2024opera} suggests addressing hallucinations during the inference phase, without introducing additional data or models for training, as methods like SPALMA do. The motivation stems from the observation of a ``partial over-trust'' phenomenon in self-attention maps. Specifically, attention scores surrounding certain summary tokens exhibit a column-like structure, termed as an aggregation pattern. This pattern tends to appear more frequently with longer contexts, increasing the likelihood of hallucinations. To tackle this, OPERA introduces a column-wise metric to identify these aggregation patterns and designs a penalty to adjust token generation probabilities, thereby reducing hallucinations. Additionally, they propose a rollback strategy where, if a hallucination is detected based on the score, the process reverts to the summary token to regenerate the token sequence.

% \subsubsection{Honesty} 
% Honesty \cite{yang2024alignment}认为幻觉是回答超过LLM知识边界的问题。其专注于LLM honesty，认为一个honest model should candidly answer questions it knows and humbly admit to those it does no。其首先定义了evolutionary metrics去判断模型是否回答了超过其能力的问题。其次其定义了IDK（I don't know） response，要求模型回答IDK如果超过了它的知识边界。最后，尝试用包括prompt engineering和SFT的方式去解决该异常现象。

Honesty \cite{yang2024alignment} posits that hallucinations occur when questions surpass the knowledge boundaries of a LLM. The focus is on the honesty of LLMs, suggesting that an honest model should openly answer questions it is knowledgeable about and humbly acknowledge when it lacks the necessary knowledge. Initially, evolutionary metrics are defined to assess whether the model has responded to questions beyond its capabilities. Additionally, an IDK (I don't know) response is introduced, requiring the model to reply with IDK when a question exceeds its knowledge boundaries. Finally, techniques such as prompt engineering and supervised fine-tuning are explored to address this anomaly.

\subsubsection{Grey-box}

% \subsubsection{LURE}

% LURE认为通过大规模数据高质量数据训练减少幻觉不可行，因此提出了一种轻量化的方法。其通过实验和理论分析得出幻觉来源于三个因素co-occurrence, uncertainty, and object position。因此其利用这三个因素用LLM构造一些幻觉样本和正常样本，基于这些数据训练一个revisor。Revisor负责检测幻觉并在inference阶段直接修改response。
LURE \cite{zhou202LURE} argues that reducing hallucinations through extensive training on large-scale, high-quality data is not feasible. Therefore, it proposes a lightweight approach. Through experimental and theoretical analysis, it identifies three factors contributing to hallucinations: co-occurrence, uncertainty, and object position. Leveraging these factors, LURE constructs both hallucinated and normal samples using LLMs. Based on this data, a revisor is trained, which is responsible for detecting hallucinations and directly modifying responses during the inference phase.

% \subsubsection{Conformal}

% Conformal 参考了Conformal prediction ，其认为虽然LM的采样空间有限，但是依旧可以通过采样剪枝的方式获取较好的回复。具体来说，其通过根据token sequence logtis 计算采样回复的质量，如果满足了stopping rule，则舍弃。

Conformal \cite{quach2023conformal} draws inspiration from conformal prediction and suggests that although the sampling space of language models is limited, it is still possible to obtain better responses through sampling and pruning methods. Specifically, it evaluates the quality of sampled responses by calculating the logits of the token sequence. If the responses meet certain stopping rules, they are retained; otherwise, they are discarded.

\subsubsection{Black-box}

% \subsubsection{Debate}

% Debate认为之前的缓解幻觉的方法大都从单个model instance出发，例如self- consistency。然而，Debate从The Society of Mind的视角出发，认为不同的LLM或者agent对于同一个input，会产生不同的见解。因此其通过prompt使得每个instance读取并评判其他所有instances的回答，并最终更新自己的回答。通过如此不断迭代，最终得到正确率更高的答案。

Debate \cite{debate} argues that previous methods for alleviating hallucinations, such as self-consistency, typically focus on a single model instance. However, drawing inspiration from ``The Society of Mind,'' Debate posits that different LLMs or agents can offer diverse perspectives on the same input. To leverage this, each instance is prompted to read and evaluate the responses of all other instances, enabling them to update their own answers. Through iterative refinement, this approach aims to arrive at a more accurate solution.

% \subsubsection{CoK}

% CoK 是一种新的推理流程，它让 LLM 在回答过程中，主动向多个异构知识源（如wiki百科、数据库）发出查询，以获取准确的事实信息。这些信息被逐步整合到生成的推理链里，从而减少“胡编乱造”的风险。在 CoK 框架下，还会对之前生成的推理链进行回顾和纠错，然后再据此优化后续推理，从而逐步提升整体回答的事实准确性和逻辑连贯性。

CoK \cite{cok} is an innovative inference framework that enables LLMs to actively query multiple heterogeneous knowledge sources (such as wikis and databases) during the answering process to obtain accurate factual information. This information is progressively integrated into the generated inference chain, thereby reducing the risk of ``hallucinations.'' Within the CoK framework, previously generated inference chains are reviewed and corrected, which in turn optimizes subsequent inference. This iterative process gradually enhances the factual accuracy and logical coherence of the overall response.

\subsection{Planning Anomalies}

Planning anomalies can be categorized based on their occurrence location into single-step and multiple-step anomalies.

\subsubsection{Single-step}

% \subsubsection{Introspective}

% introspective planning提出LLM很容易产生与常识相悖或者无法执行的plan。因此提出两个技术解决。一是通过RAG召回类似的plan帮助生成合理的下一步的计划，二是采用Conformal Prediction技术，根据token概率生成多个高质量的答案，之后根据human的feedback进一步选择。

Introspective \cite{liang2024introspective} identifies that LLMs often generate plans that are either impractical or contradict common sense. To address this, two techniques are proposed. First, RAG is used to recall similar plans to aid in generating a plausible subsequent step. Second, conformal prediction techniques are employed to generate multiple high-quality answers based on token probabilities, with human feedback used to further refine the selection.

% \subsubsection{API-bank}

% API-bank认为可以通过 finetuning LLM 增强LLM planning的能力从而降低幻觉。因此API-bank广泛的收集了真实API调用的对话并构建了数据集合，使用该数据集合训练模型，使得模型planning 的能力显著增强。同时其在评估阶段其也分析了主要的错误就是planning幻觉并且给出了未来的研究方向。。

API-bank \cite{li2023apibank} posits that fine-tuning LLMs can enhance their planning capabilities, thereby reducing hallucinations. To achieve this, API-bank has extensively collected dialogues involving real API calls and constructed a dataset, which is then used to train the model. This training substantially enhances the model's planning abilities. During the evaluation phase, the primary errors, identified as planning hallucinations, were analyzed, and potential future research directions were outlined.

% \subsubsection{ReAct}

% ReAct认为，直接让模型根据历史信息生成plan相对困难，因此其提出reasoing and act的方案，即让模型首先针对历史信息进行推理，之后再根据推理能力生成plan。reasoning 和act两阶段 给模型提供了缓冲时间，其可以思考并纠正错误最终得到相对更加完美的plan。具体来说，可以通过prompt engineering 的方式实现。

ReAct \cite{yao2023react} suggests that it is relatively challenging for a model to generate a plan based solely on historical information. Therefore, it proposes a ``reasoning and act'' approach, where the model first engages in reasoning based on historical data and subsequently formulates a plan using its reasoning capabilities. The two stages of reasoning and acting provide the model with a buffer period, allowing it to reflect and correct errors, ultimately leading to a more refined plan. This approach can be implemented through prompt engineering.

\subsubsection{Multiple-step}

% \subsubsection{Toolllm}
% Toolllm认为之前的方法虽然也降低了planning的幻觉，但是存在很多limitations。主要是在于之前的方法如API-bank场景受限，只考虑了单个API的调用，然而真实环境需要多轮多API共同完成一个任务。因此Toolllm收集大量的API，基于API生成single-tool和multi-tool的instructions，同时提出了a novel depth-first search-based decision tree来生成合理的planning path。基于这些数据微调LLM提升了LLM的工具调用能力。

Toolllm \cite{qin2023toolllm} acknowledges that while previous methods have reduced planning hallucinations, they have several limitations. Primarily, methods like API-bank are restricted by context, focusing only on single API calls. However, real-world environments often require multiple APIs and multiple interactions to complete a task. To address this, Toolllm collects a large number of APIs and generates both single-tool and multi-tool instructions. Additionally, a novel depth-first search-based decision tree is proposed to create a coherent planning path. Using this data, LLMs are fine-tuned to enhance their tool-calling capabilities.

% \subsubsection{Reflexion}

% Reflexion认为agent在多轮迭代之后可能存在幻觉。因此其提出了一种基于外部反馈的方式使得agent对于之前的执行步骤进行反思，同时优化后续的trajectory。具体的反馈可以分为三种，simple binary environment feedback, pre-defined heuristics for common failure cases, and self-evaluation such as binary classification using LLMs (decision-making) or self-written unit tests (programming).这样的反思方式相比其他方式例如RL training，更加轻量化同时有效。

Reflexion \cite{shinn2023reflexion} posits that agents may experience hallucinations after multiple iterations. To tackle this, it proposes an approach based on external feedback, enabling the agent to reflect on previous execution steps and optimize subsequent trajectories. The feedback can be categorized into three types: simple binary environment feedback, predefined heuristics for common failure cases, and self-evaluation such as binary classification using LLMs (for decision-making) or self-written unit tests (for programming). This reflection method is more lightweight and effective compared to other approaches like reinforcement learning training.

% \subsubsection{CodeAct}

% CodeAct认为普通的通过json方式进行tool调用的方式，很容易出现参数错误等幻觉产生，导致整个任务失败。因此其提出将整个planning转为代码，通过python执行器执行代码，这样参数会严格正确传递保证执行的正确，任务成功率更高。

CodeAct \cite{codeact} observes that the conventional approach of invoking tools through JSON is prone to hallucinations, such as parameter errors, which can lead to task failure. To address this, CodeAct suggests converting the entire planning process into code, which is then executed by a Python executor. This ensures that parameters are passed correctly and precisely, thereby increasing the likelihood of successful task execution.

\subsection{Action Anomalies}

Since actions are generally executed through function calls and the MCP, action anomalies can be divided into function call anomalies and MCP anomalies.

% 早期的action执行是通过function call实现的，一旦执行错误，错误信息会直接返回，因此异常检测和mitigation都相对容易，因此这里不做过多介绍
\subsubsection{Function Call}

In the early stages, action execution was implemented through function calls. When an error occurred, it would immediately return an error message, making anomaly detection and mitigation relatively straightforward. Therefore, this topic will not be extensively discussed here.

\subsubsection{MCP}

% \subsubsection{AI-Infra-Guard}

% AI-Infra-Guard通过agent技术检测MCP中可能存在的risk，其中包括Tool Poisoning Attack，Rug Pull等常见的risk。MCP 异常检测只是AI-Infra-Guard的其中一个功能。除此之外，其还能对于AI component进行安全扫描。

AI-Infra-Guard \cite{ai-infra-guard} utilizes agent technology to detect potential risks within MCP, including common threats such as tool poisoning attacks and rug pulls. Anomaly detection in MCP is just one of the functionalities of AI-Infra-Guard. Additionally, it can perform security scans on AI components.

% \subsubsection{MCP Guardian}

The core mechanism of MCP Guardian is to add a ``security-first'' middleware layer to AI systems based on the MCP. This layer enhances communication between the client (MCP client) and the data source/tool server (MCP server). It comprises four main components: authentication, rate limiting, detailed logging of call information, and deep inspection of packet contents.

\subsection{Memory Anomalies}

% 由于agent的memory分为short-term memory 和long-term memory，因此memory anomalies也分为 short-term memory anomalies 和long-term memory anomalies 

Since an agent's memory is divided into short-term memory and long-term memory, memory anomalies can be categorized into short-term memory anomalies and long-term memory anomalies.

\subsubsection{Short-Term Memory}

% \subsubsection{PI}

% 之前的方法利用位置编码如RoPE，他们train on short context window然而inference on longer ones，但是这样的 方法have weak extrapolation properties。因此PI提出position interpolation，即利用position encoding 可以是非整数的性质，将整个window进行放缩，例如4096缩小到2048，从而满足模型context window length的限制。

Previous methods, such as Rotary Positional Embedding (RoPE), are trained on short context windows but inferred on longer ones, resulting in weak extrapolation properties. To address this, PI \cite{PI} introduces position interpolation. This method leverages the property that position encodings can be non-integer values, allowing the entire window to be scaled—for example, reducing 4096 to 2048—to fit within the model's context window length limitations.

% \subsubsection{CoA}

% CoA认为之前通过微调等方式扩大context window length的方法，会很容易出现lost in the middle问题。因此其提出一种基于多智能体协作的方式，在阶段1中，不同的agent处理不同的chunk并将处理结果传递给后续的agent，阶段2中 manage agent最终给出最终答案。

CoA \cite{CoA} argues that previous methods for expanding context window length, such as fine-tuning, can easily lead to the ``lost in the middle'' problem. To address this, CoA proposes a multi-agent collaboration approach. In Phase 1, different agents handle different chunks of the input and pass their results to subsequent agents. In Phase 2, a managing agent consolidates these inputs to provide the final answer.

\subsubsection{Long-Term Memory}

% \subsubsection{ReDeep}

% ReDeep认为RAG过程中很容易由于外部知识和内部参数知识冲突导致幻觉异常。因此其通过计算和处理计算两个分数，external context score 和parametric knowledge score来共同评估幻觉。mitigates hallucinations by modulating the contributions of Knowledge FFNs and Copying Heads in the residual stream。

ReDeep \cite{sun2024redeep} posits that during the RAG process, conflicts between external knowledge and the model's internal parameters can easily lead to hallucination anomalies. To address this, ReDeep evaluates hallucinations by calculating and processing two scores: the external context score and the parametric knowledge score. It mitigates hallucinations by modulating the contributions of Knowledge Feed-Forward Networks (FFNs) and Copying Heads within the residual stream.

% \subsubsection{LRP4RAG}

% LRP4RAG使用了一个经典算法LRP，其通过perform LRP backward after the generator outputs logits to obtain a relevance matrix。之后将 relevance matrix输入到提前训练好的分类器中判别幻觉。总的来说，RAG的幻觉检测和普通的幻觉检测有很大的类似之处。

LRP4RAG \cite{hu2024lrp4rag} employs a classic algorithm, Layer-wise Relevance Propagation (LRP), by performing LRP backward after the generator outputs logits to obtain a relevance matrix. This relevance matrix is then fed into a pre-trained classifier to detect hallucinations. Overall, hallucination detection in RAG shares significant similarities with standard hallucination detection.

\subsection{Environment Anomalies}

% environment anomalies一般会体现在环境相关的指标异常，或者日志异常。因此传统的基于微服务的观测和异常检测方法便可以有效解决，如time series anomaly detection。因此这里不再赘述。

Environment anomalies are typically reflected in irregularities in environment-related metrics or log anomalies. As a result, traditional microservice-based monitoring and anomaly detection methods, such as time series anomaly detection, can effectively address these issues. Therefore, they will not be further elaborated here.

\subsection{Task Specification Anomalies}

% 如上文所提及，unclear task specification很容易导致任务的失败，而该类异常却经常被忽略。因为传统系统很少有人的参与，并且运维大都只关注excution阶段，并且不涉及pre excution阶段的一些人为配置。而在AgentOp三种，人是agent system 中的重要一环，因此其导致的异常不可忽略。

%然而当前并没有太多针对此的研究，只是单纯的有一些研究认为其是很重要的一类异常，但是并没有很多有效的解法。一个可行的解法是，类似于幻觉检测，我们通过不同prompt后LLM 隐层的状态判断。原因是，相对清晰的prompt后续的随机性较低，然而不明确的提示后续会产生更多可能的path。

As mentioned earlier, unclear task specifications can easily lead to task failures, yet this type of anomaly is often overlooked. Traditional systems seldom involve human participation, and operations typically focus on the execution stage, without addressing human configurations in the pre-execution stage. In AgentOps, humans are a critical component of the agent system, making the anomalies they cause impossible to ignore.

However, there is currently limited research focused on this issue. While some studies acknowledge it as a significant type of anomaly \cite{whydomasfail}, few effective solutions have been proposed. A feasible approach involves using methods similar to hallucination detection, where the latent states of LLMs are evaluated in response to different prompts. The rationale is that clearer prompts result in less randomness in subsequent stages, whereas unclear prompts can lead to a wider range of possible paths.

\subsection{Security Anomalies}
% security anomalies 严重影响了系统的安全，因此很多方法常识检测这些异常攻击，主要是基于图的方法。
Security anomalies can severely compromise system safety, prompting numerous methods to attempt detection of these malicious attacks, primarily employing graph-based approaches.

\subsubsection{Graph}

% \subsubsection{GUARDIAN}

% 多智能体 vote的方式 是解决外部攻击导致agent安全异常的常见方式，然而GUARDIAN认为该方式忽略了agent之间的dependency。因此其将agent system建模为图，之后利用encoder 和decoder对于整个图压缩重建，根据重建分数判断可能异常的位置。

The multi-agent voting approach is a common method for addressing security anomalies in agents caused by external attacks. However, GUARDIAN \cite{zhou2025guardian} argues that this approach overlooks the dependencies between agents. Therefore, it models the agent system as a graph and uses an encoder-decoder framework to compress and reconstruct the entire graph. Potential anomaly locations are identified based on reconstruction scores.

% \subsubsection{SentinelAgent}

% SentinelAgent认为多智能体系统中会出现很多故障，multi-agent coordination risks是很常见的故障。因此其提出一个三层检测方法，首先从全局视角考虑问题之后再深入检查。

SentinelAgent \cite{he2025sentinelagent} recognizes that many security amomalies can occur in multi-agent systems, with multi-agent coordination risks being particularly common. To address this, it proposes a three-tier detection approach, beginning with a global perspective and then delving into more detailed inspections.

\subsection{Communication Anomalies}
% 通讯异常的主要原因是，随着智能体系统的不断扩大，传递的消息越来越多，然而很多都是不必要的，过多的冗余会导致任务失败。

The primary cause of communication anomalies is the continuous expansion of agent systems, leading to an increasing volume of transmitted messages. However, many of these messages are unnecessary, and excessive redundancy can result in task failure.

\subsubsection{Redundancy}

% \subsubsection{AgentPrune}

%  AgentPrune 认为不论是intra-dialogue communication 还是 inter-dialogue communication都会导致communication redundancy现象，即agent之间的communication 会消耗大量的token，并且沟通也很低效。因此AgentPrune提出，将整个agent系统构建为一个图，之后training a low-rank-principle-guided graph mask，使用这个mask提取图的重要部分，并在这些重要部分上沟通。AgentPrune缓解了communication redundancy异常并且提升了沟通效率。

AgentPrune \cite{zhang2024agentprune} argues that both intra-dialogue and inter-dialogue communication can lead to communication redundancy, with agents consuming a large number of tokens and engaging in inefficient exchanges. To address this, AgentPrune proposes constructing the entire agent system as a graph and then training a low-rank-principle-guided graph mask. This mask is used to extract and focus communication on the important parts of the graph. AgentPrune alleviates communication redundancy and enhances communication efficiency.

% \subsubsection{G-Designer}

% G-Designer则是从更宏观的角度出发，认为不同的task的最优communication topology不同。因此其针对具体不同的task生成不同的拓扑。具体来说，其通过一个graph auto-encoder将整个agent system 压缩并重建出一个最优拓扑。

G-Designer \cite{zhang2024gdesigner} takes a more macro perspective, asserting that the optimal communication topology varies depending on the task. Therefore, it generates different topologies tailored to specific tasks. Specifically, it employs a graph auto-encoder to compress the entire agent system and reconstruct an optimal topology.

\subsection{Trust Anomalies}

% trust anomalies总的来说还是判断一个agent的message是否可靠，简单的方法是直接通过reasoning anomalies中提到的方法解决。当然从system level角度考虑也会有新的见解。

Trust anomalies generally involve assessing whether an agent's message is reliable. A straightforward approach is to address this using the methods discussed for reasoning anomalies. However, considering the issue from a system-level perspective can also provide new insights.

% \subsubsection{ATrust}

% ATrust指出在A2A越来越火热的前提下，越来越需要对于不同source的agent 的message的可信任度进行评估，否则会严重影响任务的安全。然而存在很多挑战，基于提示工程的评估方法存在幻觉，基于外部工具的方法受限制与外部工具质量。因此ATrust基于attention map从六个纬度评估可信任度，分别是factual accuracy, logical consistency, relevance, bias, language quality, and clarity。

ATrust \cite{atrust} highlights the growing need to assess the trustworthiness of messages from various source agents as A2A communication becomes increasingly popular. Failure to do so can severely compromise task security. However, many challenges exist. Methods based on prompt engineering are prone to hallucinations, while those relying on external tools are limited by the quality of these tools. ATrust evaluates trustworthiness based on attention maps across six dimensions: factual accuracy, logical consistency, relevance, bias, language quality, and clarity.

\subsection{Emergent Behavioral Anomalies}

% Emergent behavioral anomalies arise from complex interactions between multiple agents, creating system-level behaviors that cannot be attributed to any single agent. These anomalies are particularly insidious because they may not violate any individual agent's constraints, yet still produce undesirable system outcomes. 该类异常目前业界并没有有效的检测方法。

Emergent behavioral anomalies result from complex interactions among multiple agents, leading to system-level behaviors that cannot be attributed to any single agent. These anomalies are particularly insidious because they may not violate the constraints of any individual agent, yet still produce undesirable system outcomes. Currently, there are no effective detection methods for this type of anomaly in the industry.

\subsection{Termination Anomalies}

% 该类异常可能表现为 循环不停和提前停止。循环不停一般通过log和trace可以轻易检测，而提前停止一般通过检查任务的成功率等指标来检测，目前业界并没有其他前沿的有效的研究。

This type of anomaly may manifest as either endless loops or premature termination. Endless loops can often be easily detected through logs and traces, whereas premature termination is typically identified by monitoring task success rates and similar metrics. Currently, there is no other cutting-edge, effective research in the industry on this topic.

\section{Root Cause Analysis} \label{sec:rca}

\input{rca}

\section{Resolution}

\input{resolution}

%% file: operations.tex
\begin{figure}[t]
    \centering
    \includegraphics[width=0.9\textwidth]{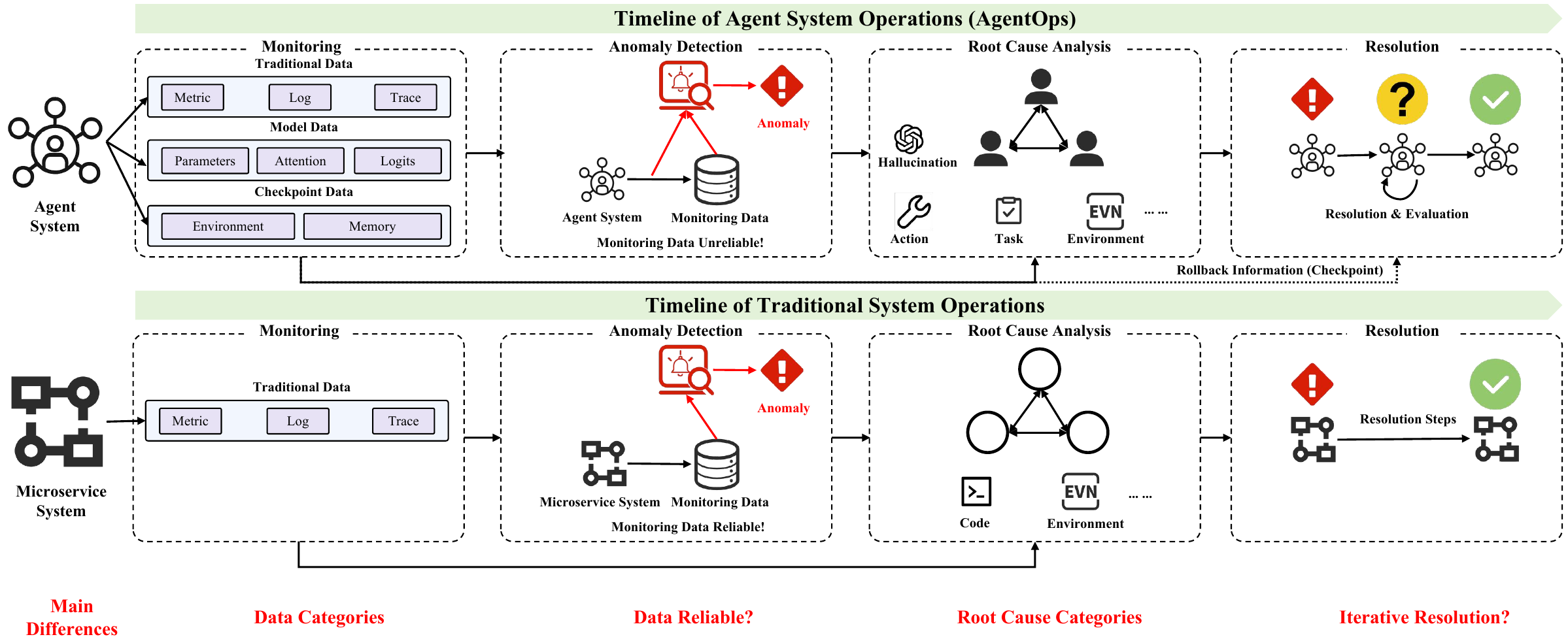}  % 图片文件名（不带扩展名）
    \caption{Comparison of traditional system operations and agent system operations.}
    \label{fig:operations}  % 图片标签，可用于引用
\end{figure}

\subsubsection{Timeline of Operations}

% 如图1所示，一般来说，业界目前把operations分为四个阶段-monitoring、anomaly detection 、root cause analysis和resolution。（1）Monitoring：monitoring阶段，我们部署可观测性工具，从多个角度尽可能全面的收集系统运行时的数据，其中包括metric、log、trace等，这些数据为后续的异常检测等过程提供支持。（2）Anomaly Detection：当system发生故障时，需要利用监控数据及时进行检测，从而在短时间内阻断故障的进一步传播，避免造成更大的影响。（3）Root Cause Analysis：当系统发生故障时候，故障可能通过不同component的调用进行传播，因此多个component都会产生告警，只有通过root cause analysis分析最终的根因才能快速解决故障。（4）Resolution：一旦通过上述过程定位到根因，便可以通知SRE或者采用自动化的修复手段修复故障。

As depicted in Figure \ref{fig:operations}, the industry generally divides operations into four phases: monitoring, anomaly detection, root cause analysis, and resolution.

\begin{itemize}
    \item \textbf{Monitoring}: In the monitoring phase, observability tools are deployed to collect runtime data from the system as comprehensively as possible, from multiple perspectives. This includes metrics, logs, traces, and more, providing crucial support for subsequent processes like anomaly detection.
    \item \textbf{Anomaly Detection}: When a system failure occurs, it is essential to use monitoring data to promptly detect anomalies. This helps prevent further propagation of failures in a short amount of time, thereby minimizing potential impact.
    \item \textbf{Root Cause Analysis}: In the event of a system failure, the issue may spread through the invocation of different components, leading to multiple components generating alerts. Only through root cause analysis can the fundamental cause be identified, allowing for a quick resolution of the issue.
    \item \textbf{Resolution}: Once the root cause has been identified through the aforementioned processes, the issue can be resolved by notifying a Site Reliability Engineer (SRE) or employing automated repair methods.
\end{itemize}

\subsubsection{Difference between Traditional System Operations and Agent System Operations}

% 如图1所示，尽管traditional operations 和AgentOps在time line上一致，然而在每个阶段都存在巨大差别，这也是为什么我们需要新的Operations框架AgentOps的原因，下面将详细介绍差异。

% Monitoring：差别主要体现在监控数据种类上。traditional operations基于OpenTelemetry，只需要监控metric、log、trace，这些数据反映了系统的真实运行状态，基于这些数据也可以对系统状态进行反推。然而对于agent system，其与传统系统的本质区别在于服务由LLM agent提供，而LLM agent具有很大的随机性。因此，我们需要额外监控LLM agent的相关模块。LLM agent主要包括 LLM 和agent 两部分。因此，针对LLM，需要监控LLM的状态例如 model parameters、attention map、token logits等。针对agent，则需要监控每个步骤的memory、enviroment等checkpoint信息。监控agent级别的checkpoint信息，不仅仅可以更加清晰的了解agent system的状态，同时还可以用作rollback，这是agent system相比传统system的巨大优势，可操作性强。

% Anomaly Detection：差别主要体现在阶段上。传统运维的异常检测针对确定性的系统，这些系统产生的数据一般认为是准确无误的，我们可以100%相信这些数据。因此传统运维的异常检测直接使用这些数据作为输入。而agent system则完全不同，数据是LLM agent产生的，正确性很难保证。因此异常检测不仅仅需要检测这些数据的生成是否合理，还需要用这些数据进一步检验系统的状态。

% Root Cause Analysis： 差别主要体现在最终的定位对象和粒度上。传统的root cause analysis只需要定位到service 和pod或者enviroment ，或者进一步深入到code。而对于agent system，由于LLM agent的参与，因此定位可能涉及到agent的某个action 或者LLM某个步骤的hallucinations。

% Resolution： 差别主要体现在修复过程上。传统的operations一旦知道了故障的具体位置和原因，便可以查询相关的领域知识，用确定化的步骤进行快速修复。而对于agent system，由于存在很多随机性，修复是一个长期并且繁杂的过程，其中需要不断尝试，或者回滚（利用monitoring中的相关数据），从而最终达到一个更优的状态。例如agent system异常的根因是prompt不合理，那么需要不断的进行A/B Test进行prompt optimization。

As shown in Figure \ref{fig:operations}, although traditional system operations and agent system operations share a similar timeline, they differ significantly at each stage. This is why a new operations framework, AgentOps, is necessary. Below, we will detail these differences.

\begin{itemize}
    \item \textbf{Monitoring}: The primary differences lie in the types of monitoring data. Traditional system operations, based on OpenTelemetry \cite{opentelemetry2019}, focus on monitoring metrics, logs, and traces, which reflect the actual operational state of the system. These data allows for reverse engineering of the system's state. However, for agent systems, the fundamental distinction is that services are provided by LLM agents, which are inherently stochastic. This necessitates additional monitoring of the relevant modules of LLM agents. An LLM agent consists mainly of the LLM and the agent parts. For the LLM aspect, monitoring includes model parameters, attention maps, token logits, and other relevant states. For the agent part, it is essential to monitor checkpoints like memory and environment at each step. Monitoring these agent-level checkpoints not only provides a clearer understanding of the agent system's state but also facilitates rollback operations, a significant advantage of agent systems over traditional systems due to their strong operability.
    \item \textbf{Anomaly Detection}: The differences are mainly in the application timing of anomaly detection. Traditional system operations applied to deterministic systems rely on data generally considered accurate and reliable, allowing the data to be used directly as input for anomaly detection. In contrast, agent systems generate data through LLM agents, where the correctness is not guaranteed. As a result, anomaly detection must not only verify the reasonableness of the data generation but also use this data to further assess the system's state.
    \item \textbf{Root Cause Analysis}: The differences are mainly in the object and granularity of localization. Traditional root cause analysis focuses on identifying issues at the service, pod, or environment level, or even delving into the code. For agent systems, due to the involvement of LLM agents, localization may involve identifying specific actions of the agent or hallucinations at a particular step of the LLM process.
    \item \textbf{Resolution}: The primary differences lie in the resolution process. In traditional system operations, once the precise location and cause of a fault are known, the issue can be promptly addressed using deterministic procedures based on relevant domain knowledge. In agent systems, however, due to inherent randomness, the resolution process is long-term and complex, requiring continual testing and possible rollback (utilizing relevant data from the monitoring phase) to eventually achieve an optimal state. For instance, if the root cause of an agent system's anomaly is an unreasonable prompt, it may require ongoing A/B testing for prompt optimization.
\end{itemize}

\subsection{Conclusion}

% \begin{table}[h]
% \centering
% \caption{Comparison between OpsAgent and AgentOps.}
% \label{tab:diffopsagentandagentops}
% \footnotesize
% {
% \begin{tabular}{ccc} 
% \toprule
%                     & \textbf{OpsAgent}                    & \textbf{AgentOps}                    \\ 
% \midrule
% \textbf{Object}     & Microservice System, Data Center ...~ & Agentic System                       \\
% \textbf{Technology} & Agent                                & Software, Machine Learning, Agent    \\
% \textbf{Input}      & Monitoring data from general system  & Monitoring data from agentic system  \\
% \bottomrule
% \end{tabular}
% }
% \end{table}

% Inspired by differentce above, we define \textbf{Agent}ic System \textbf{Op}eration\textbf{s} (\textbf{AgentOps}) as a comprehensive operational framework that encompasses the pre-execution, execution, and post-execution stages. 其与traditional system operations一样同样包括四个阶段--monitoring 、anomaly detection、root cause analysis 和resolution。其与traditional system operations（e.g.,AIOps）的根本区别在于运维对象的区别上，同时由于运维对象的差别，导致运维的各个阶段都存在较大区别，都需要新的技术解决。下面将分别介绍每一个阶段。

Inspired by the differences noted above, we define \textbf{Agent}ic System \textbf{Op}eration\textbf{s} (\textbf{AgentOps}) as a comprehensive operational framework that encompasses the pre-execution, execution, and post-execution stages. Like traditional system operations, AgentOps also includes four phases: monitoring, anomaly detection, root cause analysis, and resolution. The fundamental difference between AgentOps and traditional system operations (such as AIOps) lies in the nature of the operational subject. This distinction in the subject of operations leads to significant differences across all phases, each necessitating new technical solutions. Below, we will introduce each phase in detail.

% AgentOps is a specialized operational technique specifically designed for agentic systems. The key difference between AgentOps and AIOps lies in the focus of their operations; AIOps primarily targets traditional operational entities, such as microservice systems. A commonly confused concept is the distinction between AgentOps and OpsAgent. As illustrated in Table \ref{tab:definition} and Table \ref{tab:diffopsagentandagentops}, AgentOps refers to operations for agentic systems, whereas OpsAgent involves using agents for general system operations.

% AgentOps includes monitoring and anomaly detection across all stages, as well as root cause analysis and autonomous resolution in the post-execution stage. The following sections will introduce these components in detail.

%% file: rca.tex
% \section{Root Cause Analysis (RCA): Diagnosing the "Why" of Agent System Failures}

As autonomous agents become integral to critical business operations, conducting rapid and precise Root Cause Analysis (RCA) for their failures has emerged as a core challenge for ensuring system reliability. Unlike traditional software, the root causes of agent failures are dispersed across infrastructure, model behavior, and orchestration logic, often creating complex causal chains where a failure in one dimension triggers a fault in another. This section aims to establish a systematic RCA framework for AgentOps, designed to clearly define the problem space, establish a diagnostic mapping, and propose actionable strategies for execution.

\subsection{A Taxonomy of Root Causes for Agent Failures}

To effectively diagnose failures, we first require a taxonomy that can accurately attribute the source of failure. The classification we propose is not merely an academic exercise; it is deliberately structured to be actionable and to focus the community's attention on the most novel challenges. Its design is guided by two key motivations.

First, it aligns directly with the distinct technical stacks and team responsibilities involved in building and maintaining agent systems. This ensures that the outcome of an RCA process is not an abstract finding, but a clear directive for the appropriate team—be it DevOps/Site Reliability Engineering (SRE) for system issues, ML Engineering for model-centric problems, or the agent developers for orchestration logic. This alignment with practical ownership makes the framework highly actionable.

Second, this taxonomy strategically isolates ``old'' problems from ``new'' ones. It separates the well-understood domain of system reliability from the nascent and most critical challenges of ``soft logic'' orchestration. By doing so, it allows us to leverage decades of existing engineering practice where applicable, while focusing research and tooling efforts on the unique complexities introduced by the Agent paradigm.

We propose that any agent failure's root cause can be classified into one of the following three orthogonal and comprehensive dimensions:

\begin{itemize}

    \item \textbf{System-centric}: This dimension covers the \textit{hard infrastructure and external dependencies} that support the agent's operation. It focuses on traditional system reliability and performance, where problems are typically deterministic.

    \item \textbf{Model-centric}: This dimension focuses on the \textit{inherent uncertainty and capability limitations} of the LLM that serves as the agent's cognitive core, encompassing not only logical reasoning failures but also knowledge gaps, an inability to learn from tool interactions, or performance degradation on general tasks caused by excessive fine-tuning. These issues, originating from the model itself, are often stochastic and unpredictable.

    \item \textbf{Orchestration-centric}: Unique to AgentOps, this dimension concerns the \textit{"soft logic"} that connects the system and the model—that is, the strategies and instructions guiding how the model thinks and acts. This is currently the most complex and least understood area in RCA.

\end{itemize}

This taxonomy provides a standardized language to transform a vague problem like "the agent has failed" into a clearly defined area of responsibility.

\subsection{Mapping from Anomaly Detection to RCA}

In practice, the RCA process is triggered by alerts from an Anomaly Detection (AD) system. AD answers \textit{what} has gone wrong, while RCA must investigate \textit{why}. To ensure a cohesive narrative within this paper, it is crucial to establish a clear mapping between the anomaly types identified in the AD section and the root cause categories of our RCA framework.

The AD framework classifies anomalies at two levels: \textbf{intra-agent} (concerning the internal cognitive processes of a single agent) and \textbf{inter-agent} (concerning the broader interactions between agents and their environment). Our RCA framework can systematically trace these observed anomalies back to their origins within the system-centric, model-centric, or orchestration-centric dimensions. Table \ref{tab:ad_rca_mapping} illustrates this mapping, revealing the multi-faceted nature of agent failures.

\begin{table}[h!]
\centering
\caption{Mapping Anomaly Detection Types to Root Cause Analysis Categories.}
\label{tab:ad_rca_mapping}
\resizebox{\textwidth}{!}{
\begin{tabular}{m{0.1\textwidth}cm{0.25\textwidth}m{0.25\textwidth}m{0.35\textwidth}}
\toprule
\textbf{Level} & \textbf{Anomaly Type} & \multicolumn{1}{c}{\textbf{System-centric Causes}} & \multicolumn{1}{c}{\textbf{Model-centric Causes}} & \multicolumn{1}{c}{\textbf{Orchestration-centric Causes}} \\
\hline
\multirow{13}{*}{\rotcell{\textbf{\mbox{Intra-Agent}}}} & \textbf{Reasoning Anomalies} & \textit{RAG DB providing noisy or outdated context.} & \textit{Core model hallucination; factual incorrectness; logical fallacies.} & \textit{Flawed Chain-of-Thought prompting; ambiguous instructions leading to misinterpretation.} \\
\cline{2-5}
 & \textbf{Planning Anomalies} & \textit{Tool/API documentation is unclear, leading to impossible plans.} & \textit{Model fails to decompose a complex task correctly or fails to learn how to use a new tool after several attempts.} & \textit{Poor task decomposition strategy; ReAct-style prompt errors; failure to select appropriate tools.} \\
\cline{2-5}
 & \textbf{Action Anomalies} & \textit{External API failure (e.g., 500 error); network issues; authentication failure.} & \textit{Model generates incorrectly formatted parameters (e.g., wrong data type).} & \textit{Agent logic fails to parse tool output correctly; incorrect mapping of intent to function call.} \\
\cline{2-5}
 & \textbf{Memory Anomalies} & \textit{Vector DB latency or failure (long-term); infrastructure limits on context size (short-term).} & \textit{"Lost in the middle" phenomenon; inherent model limits on context window.} & \textit{Ineffective context compression strategy; poor RAG retrieval query generation.} \\
\hline
\multirow{14}{*}{\rotcell{\textbf{\mbox{Inter-Agent}}}} & \textbf{Environment \& Termination Anomalies} & \textit{Unstable environment; resource exhaustion; API rate limiting causing loops.} & \textit{Model generates an action or code (e.g., a script requesting excessive memory) that directly causes resource exhaustion or an unrecoverable state.} & \textit{Agent logic enters an infinite loop; premature termination due to flawed exit conditions.} \\
\cline{2-5}
 & \textbf{Task Specification Anomalies} & \textit{Not applicable, as the anomaly originates from user input, not system state.} & \textit{Model misinterprets an ambiguous user request instead of asking for clarification.} & \textit{The initial prompt from the user is vague, incomplete, or contradictory.} \\
\cline{2-5}
 & \textbf{Security \& Trust Anomalies} & \textit{Compromised external tools (Tool Poisoning); insecure communication channels.} & \textit{Model is susceptible to jailbreaking or prompt injection attacks.} & \textit{Flawed agent interaction protocols that can be exploited.} \\
\cline{2-5}
 & \textbf{Communication \& Emergent Anomalies} & \textit{Network latency or partitioning in multi-agent systems.} & \textit{Not applicable, as this is a system-level interaction issue.} & \textit{Inefficient communication protocols; flawed multi-agent coordination logic leading to negative emergent behavior.} \\
\bottomrule
\end{tabular}
}
\end{table}

This mapping reveals a critical insight: \textbf{a single, high-level anomaly type rarely has a single root cause}. For instance, a \textit{reasoning anomaly} detected by the AD system could originate from a \textit{system-centric} issue (bad data from RAG), a \textit{model-centric} issue (a core hallucination), or an \textit{orchestration-centric} issue (a flawed prompt). The role of the RCA process is therefore to use the methodologies outlined in the next section to disambiguate between these potential causes and pinpoint the true origin of the failure. This structured approach prevents teams from prematurely blaming the model for what might actually be a data or prompt engineering problem.

\subsection{Core Methodologies and Strategies for Agent RCA}

An effective RCA strategy for AgentOps does not reinvent the wheel, but rather builds upon decades of operational wisdom while forging new tools for novel challenges. We advocate for a hybrid approach that integrates established AIOps methodologies with emerging, agent-specific diagnostic techniques. This layered strategy allows us to apply the right tool for the right problem, depending on whether the root cause is likely system, model, or orchestration-centric.

\textbf{The Foundation: Adapting Traditional AIOps for Agent Systems}
For diagnosing many \textbf{system-centric} failures, the robust toolkit from AIOps and traditional monitoring remains highly relevant and effective. These methods form the first line of defense in any RCA process.
\begin{itemize}
    \item \textbf{Log and Metric Correlation}: The classical approach of correlating performance metrics (e.g., latency, CPU usage) with structured error logs is fundamental for identifying infrastructure bottlenecks, service failures, or resource exhaustion. This directly addresses many of the deterministic problems in the System-centric dimension.
    \item \textbf{Distributed Tracing}: Concepts from distributed tracing (e.g., OpenTelemetry \cite{opentelemetry2019}) are essential for tracking requests across the various microservices that an agent might interact with, including databases, external APIs, and other internal systems.
\end{itemize}
However, these methods fall short when a failure's origin is semantic or logical. An AIOps tool can report that an API call returned a 404 error, but it cannot explain \textit{why the agent decided to call a non-existent endpoint in the first place}. This is where agent-specific strategies become indispensable.

\textbf{The Vanguard: Novel Strategies for Agent-Specific Failures}
To diagnose the complex, often stochastic failures within the \textbf{model-centric} and \textbf{orchestration-centric} dimensions, we must move beyond system metrics and into the realm of semantic and logical analysis.

\paragraph{Strategy 1: Full-Stack Agent Traceability}
While traditional AIOps observability relies on logs, metrics, and traces to infer system state, this is insufficient for agent systems. The probabilistic nature of LLMs introduces a fundamental challenge: identical inputs may not produce identical outputs, making traditional reproduction difficult. Therefore, Agent traceability must evolve from inferring state to 	 extbf{explicitly recording the agent's internal cognitive state} at each step. This is the prerequisite for achieving true replayability and auditability.

A comprehensive agent trace must therefore capture not just the ``what'' (the action) but the ``why'' (the state that led to the action). We propose that a full-stack agent trace must include:
\begin{itemize}
    \item \textbf{Cognitive State Snapshots}: This goes beyond a simple log of events. It requires capturing the agent's internal state before each decision point. This conceptual framework aligns closely with the classic Belief-Desire-Intention (BDI)~\cite{rao1995bdi} model of agency, where we must record:
        \begin{itemize}
            \item \textbf{Beliefs}: The current understanding of the world state, often derived from initial inputs and prior observations.
            \item \textbf{Memory}: The contents of its short-term (e.g., context window) and long-term (e.g., retrieved RAG documents) memory at that moment.
            \item \textbf{Plan/Intent}: The current high-level plan or sub-goal the agent is attempting to execute (corresponding to Desires and Intentions).
        \end{itemize}

    \item \textbf{Reasoning Process Records}: This is the ``inner monologue'' or Chain-of-Thought. It's the explicit record of the model's reasoning process that transforms its cognitive state into a decision.

    \item \textbf{Action \& Environment Interactions}: This includes the specific tool call generated (the action), the exact parameters used, and the raw, unfiltered output returned from the environment (the observation).
\end{itemize}
By capturing these internal states directly, we are no longer merely logging system behavior. We are creating a high-fidelity, replayable recording of the agent's entire decision-making lifecycle. This allows developers to precisely reconstruct the exact conditions and "cognitive context" under which a failure occurred, even if the LLM's probabilistic outputs make a perfect reproduction of the failure itself challenging. This shift from inferential to direct state observability is arguably the most critical evolution from AIOps to AgentOps, though it necessitates the development of new, deeply integrated monitoring probes within the agent's core framework.

\paragraph{Strategy 2: Hypothesis-Driven Diagnosis with Interactive Counterfactual Simulation}
This strategy represents a paradigm shift from the passive, post-mortem analysis of AIOps to an active, experimental approach unique to AgentOps. In traditional systems, true counterfactual analysis is often impossible; one cannot simply ``rewind'' a complex, stateful system to the moment before a failure and change a single variable. The analysis is limited to inferring causality from historical data.

Agent systems, however, operate in discrete cognitive cycles (e.g., think, act, observe). As established in our traceability strategy, the agent's entire internal state can be checkpointed at each step. This provides a revolutionary capability: \textbf{the ability to interactively ``time-travel'' within the agent's workflow}. This transforms diagnosis from a passive investigation into an active, experimental science, representing a qualitative leap beyond traditional post-mortem debugging.

The methodology is as follows:
\begin{enumerate}
    \item \textbf{Load the Trace}: Begin with the high-fidelity trace of a failed execution, which contains the series of cognitive state checkpoints.
    \item \textbf{Jump to a Checkpoint}: Navigate directly to the specific step in the reasoning process where a failure is suspected to have originated.
    \item \textbf{Perform a Counterfactual Intervention}: Interactively modify a single element of the agent's state at this checkpoint. This is the ``what-if'' experiment.
    \item \textbf{Resume Execution}: Relaunch the agent's execution \textit{from that modified checkpoint forward}.
    \item \textbf{Observe the Outcome}: If the subsequent behavior corrects itself and the task succeeds, the root cause has been isolated with high confidence.
\end{enumerate}
This ability to perform targeted, interactive interventions on a live, replayable workflow is a powerful capability for debugging the non-deterministic and semantic failures common in the orchestration and model-centric dimensions. While this technique is exceptionally powerful for offline, post-mortem analysis, its application in live production environments presents significant performance and cost challenges. The overhead of checkpointing complex states may limit its use to critical, sampled transactions or offline debugging scenarios, rather than as a universal real-time diagnostic tool. It is important to note, however, that the stochastic nature of LLMs (even with low temperature settings) presents a challenge to achieving perfect reproducibility. Therefore, effective counterfactual simulations depend on the ability to checkpoint and restore not just the agent's state, but also the model's internal sampling parameters where feasible.

\paragraph{Strategy 3: Semantic Comparative Analysis}
When a direct hypothesis is elusive, comparing a failed trace with a successful one (for a similar input) provides powerful clues. This approach, inspired by work on contrastive explanations in NLP, is not a comparison of metrics, but a \textit{semantic diff} of their reasoning paths. The analysis focuses on questions like: ``At which step did the reasoning diverge?'', ``Was the RAG context different?'', or ``Did the two agents interpret the same instruction differently?''. This technique is exceptionally effective at uncovering subtle orchestration or model-level issues.

%% file: resolution.tex
\subsection{Resolution Validation}
One of the most critical component of operations for agent systems is the systematic validation of implemented resolutions to ensure they have effectively corrected detected anomalies without introducing adverse side effects. The primary evaluation criterion is the \textit{Task Success Rate}, and the resolution is considered effective if the agent consistently achieves higher task success rate following its application, compared to before. Currently, there are two main methodologies for assessing this criterion:

\begin{itemize}
    \item \textbf{Manual Annotation}: Human experts review agent performance against a detailed rubric. This method is considered the gold standard for accuracy, providing nuanced judgment on task success and the subtlety of emergent behaviors. However, it is resource-intensive and does not scale effectively for continuous, real-time monitoring.
    \item \textbf{LLM-as-a-Judge~\cite{judge}}: As a scalable alternative, this approach utilizes a separate, powerful LLM to evaluate the agent's performance. The ``judge'' LLM assesses agent logs and outputs against predefined success criteria and a scoring rubric, enabling rapid, large-scale, and automated evaluation of the Task Success Rate.
\end{itemize}

\subsection{Resolution Schema: Iterative Remediation and Multi-Turn Validation}

In contrast to traditional IT systems, where anomaly resolution often follows a linear and discrete ``detect-diagnose-remediate'' workflow, addressing anomalies in agentic systems necessitates a paradigm shift towards continuous, iterative management. The assumption of localized impact, where a fix is contained and its effects predictable, is fundamentally invalid in a modern agentic system. This invalidity stems from two core properties. First, when agents are powered by LLMs, their behavior is inherently probabilistic and non-deterministic; a given input does not guarantee an identical output across multiple trials. Second, the system operates as a complex adaptive system where the global behavior is an emergent property of countless local interactions between these non-deterministic agents. Therefore, the remediation of an anomaly in a single agent is not a simple repair but a perturbation of the systemic equilibrium, an intervention that alters the operational environment for all interacting agents.

The core of this challenge lies in the high potential for \textbf{second-order effects} and \textbf{cascading behavioral shifts}. Consider an agent diagnosed with a planning anomaly, causing it to operate with suboptimal efficiency. An operations team might intervene by modifying its planning heuristics or utility function. While the intended first-order effect of the intervention may be achieved—such as improved agent efficiency—the resultant behavioral modifications invariably alter inter-agent dynamics, potentially precipitating unforeseen, systemic failures. For example, the newly efficient agent might now monopolize a critical shared resource, leading to starvation for other agents and triggering a communication \& emergent anomaly manifested as gridlock or escalating resource conflicts. In essence, resolving an intra-agent anomaly has inadvertently created a more complex inter-agent anomaly, a phenomenon that is difficult to predict apriori.

Given these challenges, anomaly resolution must be a cyclical and empirical process. Any fix should be treated as a provisional hypothesis requiring extensive validation. After an intervention, the system is not instantly repaired; it requires a period of multi-turn observation to monitor system-wide performance and interaction patterns. This allows new dynamics to emerge and potential side effects to surface. Therefore, resolution becomes an iterative process of systemic tuning: intervene, observe the system's response over multiple turns, analyze emergent behaviors, and refine the fix until a stable, desired state is achieved.

\subsection{Anomaly Resolution for Agent Systems}

% \usepackage{multirow}
\begin{table}[htb]
\centering
\caption{A summary of the proposed taxonomy for anomaly resolution in agent systems.}
\footnotesize
\label{tab:my-table}
\resizebox{\textwidth}{!}{
\begin{tabular}{cM{0.2\textwidth}m{0.45\textwidth}m{0.3\textwidth}}
\toprule
 &
  \textbf{Resolution} &
  \multicolumn{1}{c}{\textbf{Discription}} &
  \multicolumn{1}{c}{\textbf{Target Anomaly Classes}} \\ \midrule
\multirow{4}{*}{\rotatebox{90}{\textbf{System}}} &
  \textbf{Redundancy \& Voting} &
  \textit{Use multiple agents, responses, or attempts to mitigate errors via consensus mechanisms like voting or ranking.} &
  \textit{Reasoning, Action, Task Specification, Memory} \\ \cline{2-4} 
 &
  \textbf{Guardrails \& Assertions} &
  \textit{Proactively prevents errors by enforcing strict operational boundaries, constraints, and validation rules that the agent must follow.} &
  \textit{Action, Communication, Security, Memory, Execution} \\ \cline{2-4} 
 &
  \textbf{Recovery \& Rollback} &
  \textit{Roll back to a previous safe state or try an alternative path.} &
  \textit{Environment \& Termination, Execution} \\ \cline{2-4} 
 &
  \textbf{Policy \& Strategy Adaptation} &
  \textit{Adjust the agent’s long-term decision-making or policy-learning ability via curriculum learning updates or reinforcement learning.} &
  \textit{Reasoning, Planning, Task Specification, Action} \\ \midrule[1.2pt]
\multirow{2}{*}{\rotatebox[origin=c]{90}{\textbf{Prompt}}} &
  \textbf{Self-Correction \& Introspection} &
  \textit{Encourage agents to autonomously find and fix their own mistakes by reasoning about its internal thought processes and actions.} &
  \textit{Reasoning, Planning, Task Specification, Action} \\ \cline{2-4} 
 &
  \textbf{Re-specification \& Re-prompting} &
  \textit{Resolve failures from unclear instructions by iteratively refining, clarifying, or automatically optimizing the task prompt to better guide the agent.} &
  \textit{Reasoning, Planning, Task Specification, Action} \\ \bottomrule[1.5pt]
\end{tabular}
}
\end{table}

To provide a clear and actionable framework for resolving anomalies in complex agentic systems, we divide the potential resolutions into two primary categories: \textit{System Design Driven Resolutions} and \textit{Prompt Optimization Driven Resolutions}. System design driven resolutions are foundational, architectural patterns that ensure safe and reliable operation by embedding robust frameworks, safety mechanisms, and feedback loops at the system level. These are typically the responsibility of system architects and engineers. In contrast, prompt optimization driven resolutions focus on the agent's direct interaction with its underlying language model, involving the iterative refinement of its cognitive processes and interactive instructions—the domain of AI and prompt engineers. This separation allows teams to more effectively diagnose whether a failure stems from a flaw in the foundational design or in the instructional logic.

\subsubsection{System Design Driven Resolutions}

These solutions are integral to the system's architecture and require formal engineering. They focus on building durable frameworks for resilience (Redundancy \& Voting, Recovery \& Rollback), implementing preventative safety measures (Guardrails \& Assertions), and creating responsive feedback cycles for detection and evolution (Monitoring \& Alerting, Policy \& Strategy Adaptation).
\begin{itemize}
    \item \noindent\textbf{Redundancy \& Voting}. Redundancy-based resolution introduces robustness by not relying on a single agent run or output. Instead, multiple agents or runs are executed in parallel or sequence, and their outputs are aggregated through voting, scoring, or selection mechanisms \cite{vote1, vote2, vote3}. This ensemble-style redundancy mitigates stochastic failure modes, such as planning inconsistencies or semantic drift in generation, by comparing multiple hypotheses and selecting the most consistent or confident one. In practical systems, this may involve generating N candidate responses and reranking them using a reward model, or having multiple agents independently attempt a task and using majority voting to determine correctness. Redundancy is especially valuable for addressing emergent behavior anomalies, epistemic uncertainty, or brittle reasoning paths. It is also well-aligned with distributed agent systems or agent collectives, where parallelism is natural and cost-effective. Though computationally expensive, redundancy enhances reliability in high-stakes or noisy environments.
    \item \noindent\textbf{Guardrails \& Assertions}. Guardrails and assertions serve as proactive safeguards that enforce constraints on agent behavior before or after execution. These include syntactic constraints (e.g., output must follow a JSON schema), semantic validators (e.g., the answer must reference a tool result), and behavioral assertions (e.g., the agent must always output a final decision). Guardrails can be implemented via prompt constraints, post-hoc validators, or runtime execution sandboxes that check for forbidden actions or invalid tool usage. In memory-augmented systems, guardrails might check that the memory structure is not overwritten inconsistently. Assertions offer a high degree of control without fundamentally altering the model, making them ideal for pre-deployment verification or online anomaly filtering. They are particularly effective for addressing action-level anomalies, output formatting issues, or dangerous tool executions. Guardrails also play a foundational role in establishing trust and compliance boundaries in autonomous agents.
    \item \noindent\textbf{Recovery \& Rollback}. When an anomaly causes an agent to enter an invalid or unrecoverable state, the ability to revert to a previous checkpoint becomes essential. As stated before, recovery and rollback strategies involve maintaining persistent snapshots of an agent’s memory, plan stack, or execution environment and restoring them upon failure detection. This may include restoring previous belief states, regenerating an action plan, or even resetting the environment to an earlier point in time. Recovery mechanisms can be manual or automated, often integrated with system-level fault detection. This strategy is especially useful for handling termination anomalies, cascading tool failures, or planning dead-ends. In the context of long-horizon agents, rollback allows for non-linear plan adjustments and experimentation without full task failure. Its effectiveness relies on having a well-structured internal state representation and a reliable method of state capture, both of which are emerging best practices in agent architecture.
    \item \noindent\textbf{Policy \& Strategy Adaptation}. In agent systems that learn or adapt over time, many anomalies can be resolved not by fixing a single failure, but by adjusting the underlying strategy that governs the agent’s behavior. Policy and strategy adaptation involves improving the agent’s planning or learning policies in response to observed performance. This can take the form of curriculum learning updates \cite{cl1, cl2, cl3}, reinforcement learning based on verified reward \cite{rl1, rl2, rl3}, to directly enhance the agents' planning and reasoning ability, or the form of behavior policy switching to provide the agent with more reliable external tools and information. For example, if an agent repeatedly fails on a certain class of tasks, it might shift its policy to rely more on tool assistance or defer to external help. This class of resolutions is most applicable in long-running, learning-based agent systems, such as those using RLHF or online fine-tuning. Unlike guardrails or recovery, which handle local failures, policy adaptation addresses structural deficiencies and improves long-term robustness and generalization.
\end{itemize}

\subsubsection{Prompt Optimization Driven Resolutions}. These strategies focus on the direct interaction with the agent's underlying language model. They are typically implemented within the agent's reasoning loop and involve dynamically generating, refining, and evaluating the prompts used to guide the agent's thought process and actions. 

\begin{itemize}
    \item \noindent\textbf{Self-Correction \& Introspection}. This resolution strategy leverages the model’s own reasoning capacity to detect inconsistencies, hallucinations, or invalid actions and to attempt a repair without external intervention. These techniques are mostly implemented by optimizing the system prompt, enabling behaviors such as re-asking subquestions, regenerating invalid plans, and evaluating previous actions through self-verification. For example, an agent that misuses a tool may be prompted to explain its reasoning and identify mistakes before retrying. This category also encompasses techniques like “Think step-by-step” prompts (\textit{i.e.}, Chain-of-Thought), self-critique loops (\textit{e.g.}, Reflexion), or LLM-generated verification questions. Self-correction is particularly effective for resolving reasoning anomalies, flawed plan execution, or incorrect assumptions made during multi-hop inference. It is a foundational strategy for building agents that are both autonomous and robust. 
    \item \noindent\textbf{Re-specification \& Re-prompting}. The user prompt provided to an agent is most important factor affecting its final performance in most cases. Many agent failures stem from ambiguous, under-specified, or conflicting task instructions. In these cases, the optimal resolution is not to retry the same plan, but to re-specify the task in clearer, more structured terms, which is known as prompt engineering. Re-specification may involve rewriting the original prompt, decomposing the task into subgoals, or introducing auxiliary constraints to guide the agent’s generation. In addition to manually adjusting prompts based on the task, many methods have been proposed to automatically optimize prompts. \textit{Learning-based} prompt optimization methods aim to adapt prompts by leveraging model feedback or task-specific supervision signals. These approaches typically formulate prompt optimization as a differentiable objective, allowing prompts to be fine-tuned through gradient-based methods~\cite{prom11, prom12, prom13} or reinforcement learning~\cite{prom1, prom2, prom3}. \textit{Evolutionary-based} prompt optimization methods are inspired by biological principles such as mutation, crossover, and natural selection. These methods iteratively refine prompts by maintaining a population of candidates, selecting the most effective ones, and generating new variants through stochastic modifications over successive generations~\cite{evo1, evo2}.
\end{itemize}

%% file: future_research_directions.tex
\subsection{Monitoring}
% 当前agent system的monitoring层面主要的挑战一方面是观测能力的限制，传统的monitoring技术只能观测到metric、log 、trace数据。正如Section 1提及，agent system中，model data和checkpoint data都占据了重要作用。另一方面，大量的观测数据，尤其是model data等如果不加以额外的处理，会占据大量的内存。因此future directions是如何monitoring大量多种类的数据并有效存储。

In current agent systems, monitoring faces two main challenges. First, the capabilities of traditional monitoring tools are limited—they primarily capture metrics, logs, and traces. As discussed in Section \ref{sec:monitor}, model data and checkpoint data play a crucial role in agent systems, yet they often fall outside the scope of conventional observability methods. Second, the sheer volume of observed data—especially model-related data—can impose significant memory overhead if not properly managed. Therefore, a key future direction lies in developing monitoring solutions that can effectively observe diverse types of data at scale while ensuring efficient storage and resource utilization.

\subsection{Anomaly Detection}

% 如Section4所述，目前agent system中的异常多种多样，主要的挑战就是缺乏一种有效的检测算法能同时检测多种异常。如果针对每类异常都部署一套异常检测算法，那对于线上服务会造成巨大的负担。因此future directions是如何利用更简洁高效的算法设计，同一时间检测更多种类的异常，换而言之，如何设计一个针对agent system的unified 的 异常检测算法。

As discussed in Section \ref{anomalies}, anomalies in current agent systems are diverse in nature, posing a significant challenge: the lack of a unified detection algorithm capable of identifying multiple types of anomalies simultaneously. Deploying separate anomaly detection models for each anomaly type would introduce substantial overhead to online services. Therefore, a critical future direction is to explore more lightweight and efficient algorithmic designs that can detect a broader spectrum of anomalies in a unified manner—essentially, to develop a comprehensive and scalable anomaly detection framework tailored for agent systems.

\subsection{Root Cause Analysis}

% 正如 Section 7 节所阐述的，尽管一个结构化的RCA框架提供了清晰的诊断路径，但它在实践中面临一个重大挑战：因果归因的模糊性。一个观测到的异常可能源于系统、模型和编排层面多个相互交织的根本原因，从而难以精准定位其真正的失败源头。若为每个潜在原因都部署独立的深度分析，将会引入巨大的诊断延迟和计算开销。因此，一个至关重要的未来方向是发展更自动化、更高效的因果推断技术，从而能以统一的方式厘清复杂的失败场景——其本质，就是为智能体系统创建一个可扩展的、智能化的根本原因分析引擎。

As established in Section \ref{sec:rca}, while a structured RCA framework provides a clear diagnostic path, it confronts a significant challenge in practice: the ambiguity of causal attribution. A single observed anomaly can stem from multiple, intertwined root causes across the system, model, and orchestration layers, making it difficult to isolate the true origin of failure. Deploying separate, in-depth analyses for every potential cause would introduce substantial diagnostic latency and computational overhead. Therefore, a critical future direction is to develop more automated and efficient causal inference techniques that can disentangle complex failure scenarios in a unified manner—essentially, to create a scalable and intelligent root cause analysis engine tailored for agent systems.

\subsection{Resolution}
The primary challenge faced during the resolution stage of AgentOps lies in the inherent complexity and unpredictability of agentic systems powered by LLMs. Compared with traditional IT systems, resolutions in agentic systems cannot rely on deterministic, isolated fixes due to the probabilistic nature of agent behavior and complex adaptive interactions. Addressing a local anomaly in one agent often leads to unforeseen second-order effects, cascading behavioral shifts, and systemic instability, requiring iterative remediation and multi-turn validation. Effective resolution demands continuous monitoring and refinement, leveraging a mix of system-level designs (such as redundancy, guardrails, rollback strategies, and adaptive policies) and model-level interventions (including prompt optimization and self-correction). This iterative approach ensures anomalies are genuinely resolved without introducing unintended side effects, but it significantly complicates the resolution process by necessitating sustained observation, validation, and systemic tuning.

%% file: conclusion.tex
% LLM的推理能力的不断增强推动agent system 快速发展，然而agent system异常频发且缺乏有效的运维手段。因此本paper 首先对agent system的异常进行了全新的定义和全面的分类，分为intra-agent anomalies和inter-agent anomalies。同时，提出了一个全新的针对agent system的运维框架-AgentOps，其中包含monitoring 、anomaly detection、root cause analysis 和resolution。本paper针对每个模块都进行了详细的介绍和分类。总的来说，本paper推动了agent system进一步发展。

The continuous enhancement of LLMs’ reasoning capabilities has significantly accelerated the development of agent systems. However, agent systems frequently encounter anomalies and currently lack effective operations solutions. This paper offers a novel definition and comprehensive taxonomy of agent system anomalies, categorizing them into intra-agent anomalies and inter-agent anomalies. Furthermore, we propose a new operational framework—AgentOps—designed specifically for agent systems. This framework encompasses four key components: monitoring, anomaly detection, root cause analysis, and resolution. Each component is thoroughly analyzed and categorized. Overall, this work advances the understanding and management of agent systems, laying the foundation for their more robust and reliable deployment.

%% file: sample-manuscript.bbl

\begin{thebibliography}{124}

%%% the Web address.

\ifx \showCODEN    \undefined \def \showCODEN     #1{\unskip}     \fi
\ifx \showISBNx    \undefined \def \showISBNx     #1{\unskip}     \fi
\ifx \showISBNxiii \undefined \def \showISBNxiii  #1{\unskip}     \fi
\ifx \showISSN     \undefined \def \showISSN      #1{\unskip}     \fi
\ifx \showLCCN     \undefined \def \showLCCN      #1{\unskip}     \fi
\ifx \shownote     \undefined \def \shownote      #1{#1}          \fi
\ifx \showarticletitle \undefined \def \showarticletitle #1{#1}   \fi
\ifx \showURL      \undefined \def \showURL       {\relax}        \fi
% invisible to TeX
\providecommand\bibfield[2]{#2}
\providecommand\bibinfo[2]{#2}
\providecommand\natexlab[1]{#1}
\providecommand\showeprint[2][]{arXiv:#2}

\bibitem[{AgentOps.ai}(2025)]%
        {agentops}
\bibfield{author}{\bibinfo{person}{{AgentOps.ai}}.} \bibinfo{year}{2025}\natexlab{}.
\newblock \bibinfo{booktitle}{\emph{AgentOps: Developer Platform for AI Agent Observability}}.
\newblock
\urldef\tempurl%
\url{https://www.agentops.ai/}
\showURL{%
\tempurl}


\bibitem[Altmann et~al\mbox{.}(2024)]%
        {task1}
\bibfield{author}{\bibinfo{person}{Philipp Altmann}, \bibinfo{person}{Julian Sch{\"o}nberger}, \bibinfo{person}{Steffen Illium}, \bibinfo{person}{Maximilian Zorn}, \bibinfo{person}{Fabian Ritz}, \bibinfo{person}{Tom Haider}, \bibinfo{person}{Simon Burton}, {and} \bibinfo{person}{Thomas Gabor}.} \bibinfo{year}{2024}\natexlab{}.
\newblock \showarticletitle{Emergence in Multi-agent Systems: A Safety Perspective}. In \bibinfo{booktitle}{\emph{International Symposium on Leveraging Applications of Formal Methods}}. Springer, \bibinfo{pages}{104--120}.
\newblock


\bibitem[{Anthropic}(2024)]%
        {mcp}
\bibfield{author}{\bibinfo{person}{{Anthropic}}.} \bibinfo{year}{2024}\natexlab{}.
\newblock \bibinfo{title}{{Model Context Protocol (MCP)}}.
\newblock \bibinfo{howpublished}{\url{https://modelcontextprotocol.io/specification/2025-03-26}}.
\newblock
\newblock
\shownote{Released Nov 25, 2024; latest spec version 2025‑03‑26}.


\bibitem[Anthropic(2024)]%
        {claude}
\bibfield{author}{\bibinfo{person}{AI Anthropic}.} \bibinfo{year}{2024}\natexlab{}.
\newblock \bibinfo{booktitle}{\emph{The claude 3 model family: Opus, sonnet, haiku. Claude-3 Model Card}}.
\newblock


\bibitem[{Arize AI, Inc.}(2025a)]%
        {arize_phoenix2025}
\bibfield{author}{\bibinfo{person}{{Arize AI, Inc.}}} \bibinfo{year}{2025}\natexlab{a}.
\newblock \bibinfo{booktitle}{\emph{Arize Phoenix: Open‑Source LLM Tracing \& Evaluation Platform}}.
\newblock
\urldef\tempurl%
\url{https://phoenix.arize.com/}
\showURL{%
\tempurl}


\bibitem[{Arize AI, Inc.}(2025b)]%
        {arize_llamatrace2025}
\bibfield{author}{\bibinfo{person}{{Arize AI, Inc.}}} \bibinfo{year}{2025}\natexlab{b}.
\newblock \bibinfo{booktitle}{\emph{LlamaTrace — Hosted Phoenix: LLM Tracing \& Evaluation Platform}}.
\newblock
\urldef\tempurl%
\url{https://phoenix.arize.com/llamatrace/}
\showURL{%
\tempurl}
\newblock
\shownote{Hosted version of Arize Phoenix for LLM observability, tracing, evals (same as Phoenix Cloud) :contentReference[oaicite:1]{index=1}}.


\bibitem[{Arize AI, Inc.}(2025c)]%
        {arize_openinference2025}
\bibfield{author}{\bibinfo{person}{{Arize AI, Inc.}}} \bibinfo{year}{2025}\natexlab{c}.
\newblock \bibinfo{booktitle}{\emph{OpenInference: OpenTelemetry Instrumentation for AI Observability}}.
\newblock
\urldef\tempurl%
\url{https://github.com/Arize-ai/openinference}
\showURL{%
\tempurl}


\bibitem[Authors(2019)]%
        {opentelemetry2019}
\bibfield{author}{\bibinfo{person}{OpenTelemetry Authors}.} \bibinfo{year}{2019}\natexlab{}.
\newblock \bibinfo{title}{OpenTelemetry: A Cloud Native Observability Framework}.
\newblock \bibinfo{howpublished}{\url{https://opentelemetry.io}}.
\newblock


\bibitem[Azaria and Mitchell(2023)]%
        {SAPLMA}
\bibfield{author}{\bibinfo{person}{Amos Azaria} {and} \bibinfo{person}{Tom Mitchell}.} \bibinfo{year}{2023}\natexlab{}.
\newblock \showarticletitle{The internal state of an LLM knows when it's lying}.
\newblock \bibinfo{journal}{\emph{arXiv preprint arXiv:2304.13734}} (\bibinfo{year}{2023}).
\newblock


\bibitem[Bronsdon(2025)]%
        {blog}
\bibfield{author}{\bibinfo{person}{Conor Bronsdon}.} \bibinfo{year}{2025}\natexlab{}.
\newblock \bibinfo{booktitle}{\emph{Real-Time Anomaly Detection for Multi-Agent AI Systems}}.
\newblock
\urldef\tempurl%
\url{https://galileo.ai/blog/challenges-monitoring-multi-agent-systems}
\showURL{%
\tempurl}


\bibitem[Cemri et~al\mbox{.}(2025)]%
        {whydomasfail}
\bibfield{author}{\bibinfo{person}{Mert Cemri}, \bibinfo{person}{Melissa~Z Pan}, \bibinfo{person}{Shuyi Yang}, \bibinfo{person}{Lakshya~A Agrawal}, \bibinfo{person}{Bhavya Chopra}, \bibinfo{person}{Rishabh Tiwari}, \bibinfo{person}{Kurt Keutzer}, \bibinfo{person}{Aditya Parameswaran}, \bibinfo{person}{Dan Klein}, \bibinfo{person}{Kannan Ramchandran}, {et~al\mbox{.}}} \bibinfo{year}{2025}\natexlab{}.
\newblock \showarticletitle{Why do multi-agent llm systems fail?}
\newblock \bibinfo{journal}{\emph{arXiv preprint arXiv:2503.13657}} (\bibinfo{year}{2025}).
\newblock


\bibitem[Chakraborty et~al\mbox{.}(2025)]%
        {hallucinationsurvey}
\bibfield{author}{\bibinfo{person}{Neeloy Chakraborty}, \bibinfo{person}{Melkior Ornik}, {and} \bibinfo{person}{Katherine Driggs-Campbell}.} \bibinfo{year}{2025}\natexlab{}.
\newblock \showarticletitle{Hallucination detection in foundation models for decision-making: A flexible definition and review of the state of the art}.
\newblock \bibinfo{journal}{\emph{Comput. Surveys}} (\bibinfo{year}{2025}).
\newblock


\bibitem[Chang(2024)]%
        {anp}
\bibfield{author}{\bibinfo{person}{Gaowei Chang}.} \bibinfo{year}{2024}\natexlab{}.
\newblock \bibinfo{title}{Agent Network Protocol (ANP): The HTTP of the Agentic Web Era}.
\newblock
\urldef\tempurl%
\url{https://www.agent-network-protocol.com/}
\showURL{%
\tempurl}


\bibitem[Chen et~al\mbox{.}(2025)]%
        {chen2025smurfs}
\bibfield{author}{\bibinfo{person}{Junzhi Chen}, \bibinfo{person}{Juhao Liang}, {and} \bibinfo{person}{Benyou Wang}.} \bibinfo{year}{2025}\natexlab{}.
\newblock \showarticletitle{Smurfs: Multi-Agent System using Context-Efficient DFSDT for Tool Planning}. In \bibinfo{booktitle}{\emph{Proceedings of the 2025 Conference of the Nations of the Americas Chapter of the Association for Computational Linguistics: Human Language Technologies (Volume 1: Long Papers)}}. \bibinfo{pages}{3281--3298}.
\newblock


\bibitem[Chen et~al\mbox{.}(2024a)]%
        {chen2024benchmarkrag}
\bibfield{author}{\bibinfo{person}{Jiawei Chen}, \bibinfo{person}{Hongyu Lin}, \bibinfo{person}{Xianpei Han}, {and} \bibinfo{person}{Le Sun}.} \bibinfo{year}{2024}\natexlab{a}.
\newblock \showarticletitle{Benchmarking large language models in retrieval-augmented generation}. In \bibinfo{booktitle}{\emph{Proceedings of the AAAI Conference on Artificial Intelligence}}, Vol.~\bibinfo{volume}{38}. \bibinfo{pages}{17754--17762}.
\newblock


\bibitem[Chen et~al\mbox{.}(2023)]%
        {PI}
\bibfield{author}{\bibinfo{person}{Shouyuan Chen}, \bibinfo{person}{Sherman Wong}, \bibinfo{person}{Liangjian Chen}, {and} \bibinfo{person}{Yuandong Tian}.} \bibinfo{year}{2023}\natexlab{}.
\newblock \showarticletitle{Extending context window of large language models via positional interpolation}.
\newblock \bibinfo{journal}{\emph{arXiv preprint arXiv:2306.15595}} (\bibinfo{year}{2023}).
\newblock


\bibitem[Chen et~al\mbox{.}(2024b)]%
        {prom2}
\bibfield{author}{\bibinfo{person}{Yuyan Chen}, \bibinfo{person}{Zhihao Wen}, \bibinfo{person}{Ge Fan}, \bibinfo{person}{Zhengyu Chen}, \bibinfo{person}{Wei Wu}, \bibinfo{person}{Dayiheng Liu}, \bibinfo{person}{Zhixu Li}, \bibinfo{person}{Bang Liu}, {and} \bibinfo{person}{Yanghua Xiao}.} \bibinfo{year}{2024}\natexlab{b}.
\newblock \showarticletitle{Mapo: Boosting large language model performance with model-adaptive prompt optimization}.
\newblock \bibinfo{journal}{\emph{arXiv preprint arXiv:2407.04118}} (\bibinfo{year}{2024}).
\newblock


\bibitem[Cloud(2025)]%
        {google2025a2a}
\bibfield{author}{\bibinfo{person}{Google Cloud}.} \bibinfo{year}{2025}\natexlab{}.
\newblock \bibinfo{title}{{Announcing the Agent2Agent Protocol (A2A)}}.
\newblock \bibinfo{howpublished}{\url{https://developers.googleblog.com/.../a2a-a-new-era-of-agent-interoperability/}}.
\newblock
\newblock
\shownote{Accessed: 2025-06-12}.


\bibitem[{Comet ML, Inc.}(2025)]%
        {comet_opik2025}
\bibfield{author}{\bibinfo{person}{{Comet ML, Inc.}}} \bibinfo{year}{2025}\natexlab{}.
\newblock \bibinfo{booktitle}{\emph{Opik — Open‑Source LLM Evaluation Platform}}.
\newblock
\urldef\tempurl%
\url{https://www.comet.com/site/products/opik/}
\showURL{%
\tempurl}


\bibitem[{confident-ai}(2024)]%
        {DeepEval}
\bibfield{author}{\bibinfo{person}{{confident-ai}}.} \bibinfo{year}{2024}\natexlab{}.
\newblock \bibinfo{booktitle}{\emph{DeepEval: The LLM Evaluation Framework}}.
\newblock
\urldef\tempurl%
\url{https://github.com/confident-ai/deepeval}
\showURL{%
\tempurl}


\bibitem[Databricks(2025)]%
        {mlflow2025}
\bibfield{author}{\bibinfo{person}{Databricks}.} \bibinfo{year}{2025}\natexlab{}.
\newblock \bibinfo{booktitle}{\emph{MLflow: Machine Learning Lifecycle Platform}}.
\newblock
\urldef\tempurl%
\url{https://mlflow.org}
\showURL{%
\tempurl}


\bibitem[Deng et~al\mbox{.}(2025b)]%
        {vote1}
\bibfield{author}{\bibinfo{person}{Minghang Deng}, \bibinfo{person}{Ashwin Ramachandran}, \bibinfo{person}{Canwen Xu}, \bibinfo{person}{Lanxiang Hu}, \bibinfo{person}{Zhewei Yao}, \bibinfo{person}{Anupam Datta}, {and} \bibinfo{person}{Hao Zhang}.} \bibinfo{year}{2025}\natexlab{b}.
\newblock \showarticletitle{ReFoRCE: A Text-to-SQL Agent with Self-Refinement, Format Restriction, and Column Exploration}.
\newblock \bibinfo{journal}{\emph{arXiv preprint arXiv:2502.00675}} (\bibinfo{year}{2025}).
\newblock


\bibitem[Deng et~al\mbox{.}(2022)]%
        {prom1}
\bibfield{author}{\bibinfo{person}{Mingkai Deng}, \bibinfo{person}{Jianyu Wang}, \bibinfo{person}{Cheng-Ping Hsieh}, \bibinfo{person}{Yihan Wang}, \bibinfo{person}{Han Guo}, \bibinfo{person}{Tianmin Shu}, \bibinfo{person}{Meng Song}, \bibinfo{person}{Eric~P Xing}, {and} \bibinfo{person}{Zhiting Hu}.} \bibinfo{year}{2022}\natexlab{}.
\newblock \showarticletitle{Rlprompt: Optimizing discrete text prompts with reinforcement learning}.
\newblock \bibinfo{journal}{\emph{arXiv preprint arXiv:2205.12548}} (\bibinfo{year}{2022}).
\newblock


\bibitem[Deng et~al\mbox{.}(2025a)]%
        {security}
\bibfield{author}{\bibinfo{person}{Zehang Deng}, \bibinfo{person}{Yongjian Guo}, \bibinfo{person}{Changzhou Han}, \bibinfo{person}{Wanlun Ma}, \bibinfo{person}{Junwu Xiong}, \bibinfo{person}{Sheng Wen}, {and} \bibinfo{person}{Yang Xiang}.} \bibinfo{year}{2025}\natexlab{a}.
\newblock \showarticletitle{Ai agents under threat: A survey of key security challenges and future pathways}.
\newblock \bibinfo{journal}{\emph{Comput. Surveys}} \bibinfo{volume}{57}, \bibinfo{number}{7} (\bibinfo{year}{2025}), \bibinfo{pages}{1--36}.
\newblock


\bibitem[Drake(2025)]%
        {recur2}
\bibfield{author}{\bibinfo{person}{Seth Drake}.} \bibinfo{year}{2025}\natexlab{}.
\newblock \showarticletitle{'Neural howlround'in large language models: a self-reinforcing bias phenomenon, and a dynamic attenuation solution}.
\newblock \bibinfo{journal}{\emph{arXiv preprint arXiv:2504.07992}} (\bibinfo{year}{2025}).
\newblock


\bibitem[Du et~al\mbox{.}(2023)]%
        {debate}
\bibfield{author}{\bibinfo{person}{Yilun Du}, \bibinfo{person}{Shuang Li}, \bibinfo{person}{Antonio Torralba}, \bibinfo{person}{Joshua~B Tenenbaum}, {and} \bibinfo{person}{Igor Mordatch}.} \bibinfo{year}{2023}\natexlab{}.
\newblock \showarticletitle{Improving factuality and reasoning in language models through multiagent debate}. In \bibinfo{booktitle}{\emph{Forty-first International Conference on Machine Learning}}.
\newblock


\bibitem[Durante et~al\mbox{.}(2024)]%
        {lifeifei}
\bibfield{author}{\bibinfo{person}{Zane Durante}, \bibinfo{person}{Qiuyuan Huang}, \bibinfo{person}{Naoki Wake}, \bibinfo{person}{Ran Gong}, \bibinfo{person}{Jae~Sung Park}, \bibinfo{person}{Bidipta Sarkar}, \bibinfo{person}{Rohan Taori}, \bibinfo{person}{Yusuke Noda}, \bibinfo{person}{Demetri Terzopoulos}, \bibinfo{person}{Yejin Choi}, {et~al\mbox{.}}} \bibinfo{year}{2024}\natexlab{}.
\newblock \showarticletitle{Agent ai: Surveying the horizons of multimodal interaction}.
\newblock \bibinfo{journal}{\emph{arXiv preprint arXiv:2401.03568}} (\bibinfo{year}{2024}).
\newblock


\bibitem[Edge et~al\mbox{.}(2024)]%
        {edge2024graphrag}
\bibfield{author}{\bibinfo{person}{Darren Edge}, \bibinfo{person}{Ha Trinh}, \bibinfo{person}{Newman Cheng}, \bibinfo{person}{Joshua Bradley}, \bibinfo{person}{Alex Chao}, \bibinfo{person}{Apurva Mody}, \bibinfo{person}{Steven Truitt}, \bibinfo{person}{Dasha Metropolitansky}, \bibinfo{person}{Robert~Osazuwa Ness}, {and} \bibinfo{person}{Jonathan Larson}.} \bibinfo{year}{2024}\natexlab{}.
\newblock \showarticletitle{From local to global: A graph rag approach to query-focused summarization}.
\newblock \bibinfo{journal}{\emph{arXiv preprint arXiv:2404.16130}} (\bibinfo{year}{2024}).
\newblock


\bibitem[Feng et~al\mbox{.}(2025)]%
        {rl1}
\bibfield{author}{\bibinfo{person}{Lang Feng}, \bibinfo{person}{Zhenghai Xue}, \bibinfo{person}{Tingcong Liu}, {and} \bibinfo{person}{Bo An}.} \bibinfo{year}{2025}\natexlab{}.
\newblock \showarticletitle{Group-in-Group Policy Optimization for LLM Agent Training}.
\newblock \bibinfo{journal}{\emph{arXiv preprint arXiv:2505.10978}} (\bibinfo{year}{2025}).
\newblock


\bibitem[Fernando et~al\mbox{.}(2023)]%
        {evo1}
\bibfield{author}{\bibinfo{person}{Chrisantha Fernando}, \bibinfo{person}{Dylan Banarse}, \bibinfo{person}{Henryk Michalewski}, \bibinfo{person}{Simon Osindero}, {and} \bibinfo{person}{Tim Rockt{\"a}schel}.} \bibinfo{year}{2023}\natexlab{}.
\newblock \showarticletitle{Promptbreeder: Self-referential self-improvement via prompt evolution}.
\newblock \bibinfo{journal}{\emph{arXiv preprint arXiv:2309.16797}} (\bibinfo{year}{2023}).
\newblock


\bibitem[Frost et~al\mbox{.}(2024)]%
        {communicate}
\bibfield{author}{\bibinfo{person}{Emilie Frost}, \bibinfo{person}{Julia~Catharina Heiken}, \bibinfo{person}{Martin Tr{\"o}schel}, {and} \bibinfo{person}{Astrid Nie{\ss}e}.} \bibinfo{year}{2024}\natexlab{}.
\newblock \showarticletitle{Detecting and Analyzing Agent Communication Anomalies in Distributed Energy System Control.}. In \bibinfo{booktitle}{\emph{ICAART (3)}}. \bibinfo{pages}{625--632}.
\newblock


\bibitem[Gallifant et~al\mbox{.}(2024)]%
        {39}
\bibfield{author}{\bibinfo{person}{Jack Gallifant}, \bibinfo{person}{Amelia Fiske}, \bibinfo{person}{Yulia~A Levites~Strekalova}, \bibinfo{person}{Juan~S Osorio-Valencia}, \bibinfo{person}{Rachael Parke}, \bibinfo{person}{Rogers Mwavu}, \bibinfo{person}{Nicole Martinez}, \bibinfo{person}{Judy~Wawira Gichoya}, \bibinfo{person}{Marzyeh Ghassemi}, \bibinfo{person}{Dina Demner-Fushman}, {et~al\mbox{.}}} \bibinfo{year}{2024}\natexlab{}.
\newblock \showarticletitle{Peer review of GPT-4 technical report and systems card}.
\newblock \bibinfo{journal}{\emph{PLOS digital health}} \bibinfo{volume}{3}, \bibinfo{number}{1} (\bibinfo{year}{2024}), \bibinfo{pages}{e0000417}.
\newblock


\bibitem[Gao et~al\mbox{.}(2025)]%
        {gao2025single}
\bibfield{author}{\bibinfo{person}{Mingyan Gao}, \bibinfo{person}{Yanzi Li}, \bibinfo{person}{Banruo Liu}, \bibinfo{person}{Yifan Yu}, \bibinfo{person}{Phillip Wang}, \bibinfo{person}{Ching-Yu Lin}, {and} \bibinfo{person}{Fan Lai}.} \bibinfo{year}{2025}\natexlab{}.
\newblock \showarticletitle{Single-agent or Multi-agent Systems? Why Not Both?}
\newblock \bibinfo{journal}{\emph{arXiv preprint arXiv:2505.18286}} (\bibinfo{year}{2025}).
\newblock


\bibitem[{Google Cloud}(2025)]%
        {A2A}
\bibfield{author}{\bibinfo{person}{{Google Cloud}}.} \bibinfo{year}{2025}\natexlab{}.
\newblock \bibinfo{title}{Announcing the Agent2Agent Protocol (A2A)}.
\newblock
\urldef\tempurl%
\url{https://developers.googleblog.com/en/a2a-a-new-era-of-agent-interoperability/}
\showURL{%
\tempurl}


\bibitem[Guo et~al\mbox{.}(2025)]%
        {guo2025deepseekr1}
\bibfield{author}{\bibinfo{person}{Daya Guo}, \bibinfo{person}{Dejian Yang}, \bibinfo{person}{Haowei Zhang}, \bibinfo{person}{Junxiao Song}, \bibinfo{person}{Ruoyu Zhang}, \bibinfo{person}{Runxin Xu}, \bibinfo{person}{Qihao Zhu}, \bibinfo{person}{Shirong Ma}, \bibinfo{person}{Peiyi Wang}, \bibinfo{person}{Xiao Bi}, {et~al\mbox{.}}} \bibinfo{year}{2025}\natexlab{}.
\newblock \showarticletitle{Deepseek-r1: Incentivizing reasoning capability in llms via reinforcement learning}.
\newblock \bibinfo{journal}{\emph{arXiv preprint arXiv:2501.12948}} (\bibinfo{year}{2025}).
\newblock


\bibitem[Hanzlik et~al\mbox{.}(2021)]%
        {datasecurity}
\bibfield{author}{\bibinfo{person}{Lucjan Hanzlik}, \bibinfo{person}{Yang Zhang}, \bibinfo{person}{Kathrin Grosse}, \bibinfo{person}{Ahmed Salem}, \bibinfo{person}{Maximilian Augustin}, \bibinfo{person}{Michael Backes}, {and} \bibinfo{person}{Mario Fritz}.} \bibinfo{year}{2021}\natexlab{}.
\newblock \showarticletitle{Mlcapsule: Guarded offline deployment of machine learning as a service}. In \bibinfo{booktitle}{\emph{Proceedings of the IEEE/CVF conference on computer vision and pattern recognition}}. \bibinfo{pages}{3300--3309}.
\newblock


\bibitem[He et~al\mbox{.}(2024)]%
        {he2024webvoyager}
\bibfield{author}{\bibinfo{person}{Hongliang He}, \bibinfo{person}{Wenlin Yao}, \bibinfo{person}{Kaixin Ma}, \bibinfo{person}{Wenhao Yu}, \bibinfo{person}{Yong Dai}, \bibinfo{person}{Hongming Zhang}, \bibinfo{person}{Zhenzhong Lan}, {and} \bibinfo{person}{Dong Yu}.} \bibinfo{year}{2024}\natexlab{}.
\newblock \showarticletitle{WebVoyager: Building an end-to-end web agent with large multimodal models}.
\newblock \bibinfo{journal}{\emph{arXiv preprint arXiv:2401.13919}} (\bibinfo{year}{2024}).
\newblock


\bibitem[He et~al\mbox{.}(2025a)]%
        {atrust}
\bibfield{author}{\bibinfo{person}{Pengfei He}, \bibinfo{person}{Zhenwei Dai}, \bibinfo{person}{Xianfeng Tang}, \bibinfo{person}{Yue Xing}, \bibinfo{person}{Hui Liu}, \bibinfo{person}{Jingying Zeng}, \bibinfo{person}{Qiankun Peng}, \bibinfo{person}{Shrivats Agrawal}, \bibinfo{person}{Samarth Varshney}, \bibinfo{person}{Suhang Wang}, {et~al\mbox{.}}} \bibinfo{year}{2025}\natexlab{a}.
\newblock \showarticletitle{Attention Knows Whom to Trust: Attention-based Trust Management for LLM Multi-Agent Systems}.
\newblock \bibinfo{journal}{\emph{arXiv preprint arXiv:2506.02546}} (\bibinfo{year}{2025}).
\newblock


\bibitem[He et~al\mbox{.}(2025b)]%
        {he2025redteaming}
\bibfield{author}{\bibinfo{person}{Pengfei He}, \bibinfo{person}{Yupin Lin}, \bibinfo{person}{Shen Dong}, \bibinfo{person}{Han Xu}, \bibinfo{person}{Yue Xing}, {and} \bibinfo{person}{Hui Liu}.} \bibinfo{year}{2025}\natexlab{b}.
\newblock \showarticletitle{Red-teaming llm multi-agent systems via communication attacks}.
\newblock \bibinfo{journal}{\emph{arXiv preprint arXiv:2502.14847}} (\bibinfo{year}{2025}).
\newblock


\bibitem[He et~al\mbox{.}(2025d)]%
        {trust}
\bibfield{author}{\bibinfo{person}{Pengfei He}, \bibinfo{person}{Yue Xing}, \bibinfo{person}{Shen Dong}, \bibinfo{person}{Juanhui Li}, \bibinfo{person}{Zhenwei Dai}, \bibinfo{person}{Xianfeng Tang}, \bibinfo{person}{Hui Liu}, \bibinfo{person}{Han Xu}, \bibinfo{person}{Zhen Xiang}, {and} \bibinfo{person}{Charu~C Aggarwal}.} \bibinfo{year}{2025}\natexlab{d}.
\newblock \showarticletitle{Comprehensive Vulnerability Analysis is Necessary for Trustworthy LLM-MAS}.
\newblock \bibinfo{journal}{\emph{arXiv preprint arXiv:2506.01245}} (\bibinfo{year}{2025}).
\newblock


\bibitem[He et~al\mbox{.}(2025c)]%
        {he2025sentinelagent}
\bibfield{author}{\bibinfo{person}{Xu He}, \bibinfo{person}{Di Wu}, \bibinfo{person}{Yan Zhai}, {and} \bibinfo{person}{Kun Sun}.} \bibinfo{year}{2025}\natexlab{c}.
\newblock \showarticletitle{SentinelAgent: Graph-based Anomaly Detection in Multi-Agent Systems}.
\newblock \bibinfo{journal}{\emph{arXiv preprint arXiv:2505.24201}} (\bibinfo{year}{2025}).
\newblock


\bibitem[Helicone(2025)]%
        {helicone2025}
\bibfield{author}{\bibinfo{person}{Helicone}.} \bibinfo{year}{2025}\natexlab{}.
\newblock \bibinfo{booktitle}{\emph{Helicone: LLM Observability Platform}}.
\newblock
\urldef\tempurl%
\url{https://github.com/Helicone/helicone}
\showURL{%
\tempurl}


\bibitem[{HoneyHive AI, Inc.}(2025)]%
        {honeyhive2025}
\bibfield{author}{\bibinfo{person}{{HoneyHive AI, Inc.}}} \bibinfo{year}{2025}\natexlab{}.
\newblock \bibinfo{booktitle}{\emph{HoneyHive: AI Observability and Evaluation Platform}}.
\newblock
\urldef\tempurl%
\url{https://www.honeyhive.ai/}
\showURL{%
\tempurl}


\bibitem[Hu et~al\mbox{.}(2024c)]%
        {hu2024lrp4rag}
\bibfield{author}{\bibinfo{person}{Haichuan Hu}, \bibinfo{person}{Yuhan Sun}, {and} \bibinfo{person}{Quanjun Zhang}.} \bibinfo{year}{2024}\natexlab{c}.
\newblock \showarticletitle{LRP4RAG: Detecting Hallucinations in Retrieval-Augmented Generation via Layer-wise Relevance Propagation}.
\newblock \bibinfo{journal}{\emph{arXiv preprint arXiv:2408.15533}} (\bibinfo{year}{2024}).
\newblock


\bibitem[Hu et~al\mbox{.}(2024a)]%
        {hu2024survey}
\bibfield{author}{\bibinfo{person}{Sihao Hu}, \bibinfo{person}{Tiansheng Huang}, \bibinfo{person}{Gaowen Liu}, \bibinfo{person}{Ramana~Rao Kompella}, \bibinfo{person}{Fatih Ilhan}, \bibinfo{person}{Selim~Furkan Tekin}, \bibinfo{person}{Yichang Xu}, \bibinfo{person}{Zachary Yahn}, {and} \bibinfo{person}{Ling Liu}.} \bibinfo{year}{2024}\natexlab{a}.
\newblock \showarticletitle{A survey on large language model-based game agents}.
\newblock \bibinfo{journal}{\emph{arXiv preprint arXiv:2404.02039}} (\bibinfo{year}{2024}).
\newblock


\bibitem[Hu et~al\mbox{.}(2024b)]%
        {48}
\bibfield{author}{\bibinfo{person}{Zichao Hu}, \bibinfo{person}{Francesca Lucchetti}, \bibinfo{person}{Claire Schlesinger}, \bibinfo{person}{Yash Saxena}, \bibinfo{person}{Anders Freeman}, \bibinfo{person}{Sadanand Modak}, \bibinfo{person}{Arjun Guha}, {and} \bibinfo{person}{Joydeep Biswas}.} \bibinfo{year}{2024}\natexlab{b}.
\newblock \showarticletitle{Deploying and evaluating llms to program service mobile robots}.
\newblock \bibinfo{journal}{\emph{IEEE Robotics and Automation Letters}} \bibinfo{volume}{9}, \bibinfo{number}{3} (\bibinfo{year}{2024}), \bibinfo{pages}{2853--2860}.
\newblock


\bibitem[Hu et~al\mbox{.}(2024d)]%
        {vote2}
\bibfield{author}{\bibinfo{person}{Zhongjian Hu}, \bibinfo{person}{Peng Yang}, \bibinfo{person}{Bing Li}, {and} \bibinfo{person}{Zhenqi Wang}.} \bibinfo{year}{2024}\natexlab{d}.
\newblock \showarticletitle{Multi-Agents Based on Large Language Models for Knowledge-based Visual Question Answering}.
\newblock \bibinfo{journal}{\emph{arXiv preprint arXiv:2412.18351}} (\bibinfo{year}{2024}).
\newblock


\bibitem[Hua et~al\mbox{.}(2024)]%
        {hua2024war}
\bibfield{author}{\bibinfo{person}{Wenyue Hua}, \bibinfo{person}{Lizhou Fan}, \bibinfo{person}{Lingyao Li}, \bibinfo{person}{Kai Mei}, \bibinfo{person}{jianchao ji}, \bibinfo{person}{Yingqiang Ge}, \bibinfo{person}{Libby Hemphill}, {and} \bibinfo{person}{Yongfeng Zhang}.} \bibinfo{year}{2024}\natexlab{}.
\newblock \bibinfo{title}{War and Peace (WarAgent): {LLM}-based Multi-Agent Simulation of World Wars}.
\newblock
\urldef\tempurl%
\url{https://openreview.net/forum?id=RBaDiInDRg}
\showURL{%
\tempurl}


\bibitem[Huang et~al\mbox{.}(2024)]%
        {huang2024opera}
\bibfield{author}{\bibinfo{person}{Qidong Huang}, \bibinfo{person}{Xiaoyi Dong}, \bibinfo{person}{Pan Zhang}, \bibinfo{person}{Bin Wang}, \bibinfo{person}{Conghui He}, \bibinfo{person}{Jiaqi Wang}, \bibinfo{person}{Dahua Lin}, \bibinfo{person}{Weiming Zhang}, {and} \bibinfo{person}{Nenghai Yu}.} \bibinfo{year}{2024}\natexlab{}.
\newblock \showarticletitle{Opera: Alleviating hallucination in multi-modal large language models via over-trust penalty and retrospection-allocation}. In \bibinfo{booktitle}{\emph{Proceedings of the IEEE/CVF Conference on Computer Vision and Pattern Recognition}}. \bibinfo{pages}{13418--13427}.
\newblock


\bibitem[Huang et~al\mbox{.}(2025)]%
        {online}
\bibfield{author}{\bibinfo{person}{Xu Huang}, \bibinfo{person}{Jianxun Lian}, \bibinfo{person}{Yuxuan Lei}, \bibinfo{person}{Jing Yao}, \bibinfo{person}{Defu Lian}, {and} \bibinfo{person}{Xing Xie}.} \bibinfo{year}{2025}\natexlab{}.
\newblock \showarticletitle{Recommender ai agent: Integrating large language models for interactive recommendations}.
\newblock \bibinfo{journal}{\emph{ACM Transactions on Information Systems}} \bibinfo{volume}{43}, \bibinfo{number}{4} (\bibinfo{year}{2025}), \bibinfo{pages}{1--33}.
\newblock


\bibitem[{IBM Research / BeeAI}(2025)]%
        {ACP}
\bibfield{author}{\bibinfo{person}{{IBM Research / BeeAI}}.} \bibinfo{year}{2025}\natexlab{}.
\newblock \bibinfo{title}{Agent Communication Protocol (ACP)}.
\newblock
\urldef\tempurl%
\url{https://agentcommunicationprotocol.dev/introduction/welcome}
\showURL{%
\tempurl}


\bibitem[Islam and Moushi(2024)]%
        {islam2024gpt}
\bibfield{author}{\bibinfo{person}{Raisa Islam} {and} \bibinfo{person}{Owana~Marzia Moushi}.} \bibinfo{year}{2024}\natexlab{}.
\newblock \showarticletitle{Gpt-4o: The cutting-edge advancement in multimodal llm}.
\newblock \bibinfo{journal}{\emph{Authorea Preprints}} (\bibinfo{year}{2024}).
\newblock


\bibitem[Jin et~al\mbox{.}(2025)]%
        {jin2025searchr1}
\bibfield{author}{\bibinfo{person}{Bowen Jin}, \bibinfo{person}{Hansi Zeng}, \bibinfo{person}{Zhenrui Yue}, \bibinfo{person}{Jinsung Yoon}, \bibinfo{person}{Sercan Arik}, \bibinfo{person}{Dong Wang}, \bibinfo{person}{Hamed Zamani}, {and} \bibinfo{person}{Jiawei Han}.} \bibinfo{year}{2025}\natexlab{}.
\newblock \showarticletitle{Search-r1: Training llms to reason and leverage search engines with reinforcement learning}.
\newblock \bibinfo{journal}{\emph{arXiv preprint arXiv:2503.09516}} (\bibinfo{year}{2025}).
\newblock


\bibitem[Kong et~al\mbox{.}(2024)]%
        {prom3}
\bibfield{author}{\bibinfo{person}{Weize Kong}, \bibinfo{person}{Spurthi~Amba Hombaiah}, \bibinfo{person}{Mingyang Zhang}, \bibinfo{person}{Qiaozhu Mei}, {and} \bibinfo{person}{Michael Bendersky}.} \bibinfo{year}{2024}\natexlab{}.
\newblock \showarticletitle{Prewrite: Prompt rewriting with reinforcement learning}.
\newblock \bibinfo{journal}{\emph{arXiv preprint arXiv:2401.08189}} (\bibinfo{year}{2024}).
\newblock


\bibitem[Kwon et~al\mbox{.}(2023)]%
        {67}
\bibfield{author}{\bibinfo{person}{Minae Kwon}, \bibinfo{person}{Sang~Michael Xie}, \bibinfo{person}{Kalesha Bullard}, {and} \bibinfo{person}{Dorsa Sadigh}.} \bibinfo{year}{2023}\natexlab{}.
\newblock \showarticletitle{Reward design with language models}.
\newblock \bibinfo{journal}{\emph{arXiv preprint arXiv:2303.00001}} (\bibinfo{year}{2023}).
\newblock


\bibitem[langdb(2025)]%
        {langdb2025}
\bibfield{author}{\bibinfo{person}{langdb}.} \bibinfo{year}{2025}\natexlab{}.
\newblock \bibinfo{booktitle}{\emph{LangDB: LLM‑enhanced database exploration}}.
\newblock
\urldef\tempurl%
\url{https://github.com/langdb/ai-gateway}
\showURL{%
\tempurl}


\bibitem[Langfuse(2025)]%
        {langfuse2025}
\bibfield{author}{\bibinfo{person}{Langfuse}.} \bibinfo{year}{2025}\natexlab{}.
\newblock \bibinfo{booktitle}{\emph{Langfuse: Open‑Source LLM Tracing and Observability}}.
\newblock
\urldef\tempurl%
\url{https://github.com/langfuse/langfuse}
\showURL{%
\tempurl}


\bibitem[LangWatch(2025)]%
        {langwatch2025}
\bibfield{author}{\bibinfo{person}{LangWatch}.} \bibinfo{year}{2025}\natexlab{}.
\newblock \bibinfo{booktitle}{\emph{LangWatch}}.
\newblock
\urldef\tempurl%
\url{https://github.com/langwatch}
\showURL{%
\tempurl}


\bibitem[Li et~al\mbox{.}(2023d)]%
        {li2023apibank}
\bibfield{author}{\bibinfo{person}{Minghao Li}, \bibinfo{person}{Yingxiu Zhao}, \bibinfo{person}{Bowen Yu}, \bibinfo{person}{Feifan Song}, \bibinfo{person}{Hangyu Li}, \bibinfo{person}{Haiyang Yu}, \bibinfo{person}{Zhoujun Li}, \bibinfo{person}{Fei Huang}, {and} \bibinfo{person}{Yongbin Li}.} \bibinfo{year}{2023}\natexlab{d}.
\newblock \showarticletitle{Api-bank: A comprehensive benchmark for tool-augmented llms}.
\newblock \bibinfo{journal}{\emph{arXiv preprint arXiv:2304.08244}} (\bibinfo{year}{2023}).
\newblock


\bibitem[Li et~al\mbox{.}(2023a)]%
        {li2023econagent}
\bibfield{author}{\bibinfo{person}{Nian Li}, \bibinfo{person}{Chen Gao}, \bibinfo{person}{Mingyu Li}, \bibinfo{person}{Yong Li}, {and} \bibinfo{person}{Qingmin Liao}.} \bibinfo{year}{2023}\natexlab{a}.
\newblock \showarticletitle{Econagent: large language model-empowered agents for simulating macroeconomic activities}.
\newblock \bibinfo{journal}{\emph{arXiv preprint arXiv:2310.10436}} (\bibinfo{year}{2023}).
\newblock


\bibitem[Li et~al\mbox{.}(2023b)]%
        {li2023cok}
\bibfield{author}{\bibinfo{person}{Xingxuan Li}, \bibinfo{person}{Ruochen Zhao}, \bibinfo{person}{Yew~Ken Chia}, \bibinfo{person}{Bosheng Ding}, \bibinfo{person}{Shafiq Joty}, \bibinfo{person}{Soujanya Poria}, {and} \bibinfo{person}{Lidong Bing}.} \bibinfo{year}{2023}\natexlab{b}.
\newblock \showarticletitle{Chain-of-knowledge: Grounding large language models via dynamic knowledge adapting over heterogeneous sources}.
\newblock \bibinfo{journal}{\emph{arXiv preprint arXiv:2305.13269}} (\bibinfo{year}{2023}).
\newblock


\bibitem[Li et~al\mbox{.}(2023c)]%
        {cok}
\bibfield{author}{\bibinfo{person}{Xingxuan Li}, \bibinfo{person}{Ruochen Zhao}, \bibinfo{person}{Yew~Ken Chia}, \bibinfo{person}{Bosheng Ding}, \bibinfo{person}{Shafiq Joty}, \bibinfo{person}{Soujanya Poria}, {and} \bibinfo{person}{Lidong Bing}.} \bibinfo{year}{2023}\natexlab{c}.
\newblock \showarticletitle{Chain-of-knowledge: Grounding large language models via dynamic knowledge adapting over heterogeneous sources}.
\newblock \bibinfo{journal}{\emph{arXiv preprint arXiv:2305.13269}} (\bibinfo{year}{2023}).
\newblock


\bibitem[Liang et~al\mbox{.}(2024)]%
        {liang2024introspective}
\bibfield{author}{\bibinfo{person}{Kaiqu Liang}, \bibinfo{person}{Zixu Zhang}, {and} \bibinfo{person}{Jaime Fern{\'a}ndez~Fisac}.} \bibinfo{year}{2024}\natexlab{}.
\newblock \showarticletitle{Introspective planning: Guiding language-enabled agents to refine their own uncertainty}.
\newblock \bibinfo{journal}{\emph{arXiv e-prints}} (\bibinfo{year}{2024}), \bibinfo{pages}{arXiv--2402}.
\newblock


\bibitem[{Literal AI, Inc.}(2025)]%
        {literalai2025}
\bibfield{author}{\bibinfo{person}{{Literal AI, Inc.}}} \bibinfo{year}{2025}\natexlab{}.
\newblock \bibinfo{booktitle}{\emph{Literal AI: RAG LLM Evaluation \& Observability Platform}}.
\newblock
\urldef\tempurl%
\url{https://www.literalai.com/}
\showURL{%
\tempurl}


\bibitem[Liu et~al\mbox{.}(2023)]%
        {liu2023lostinmiddle}
\bibfield{author}{\bibinfo{person}{Nelson~F Liu}, \bibinfo{person}{Kevin Lin}, \bibinfo{person}{John Hewitt}, \bibinfo{person}{Ashwin Paranjape}, \bibinfo{person}{Michele Bevilacqua}, \bibinfo{person}{Fabio Petroni}, {and} \bibinfo{person}{Percy Liang}.} \bibinfo{year}{2023}\natexlab{}.
\newblock \showarticletitle{Lost in the middle: How language models use long contexts}.
\newblock \bibinfo{journal}{\emph{arXiv preprint arXiv:2307.03172}} (\bibinfo{year}{2023}).
\newblock


\bibitem[Liu et~al\mbox{.}(2021)]%
        {prom12}
\bibfield{author}{\bibinfo{person}{Xiao Liu}, \bibinfo{person}{Kaixuan Ji}, \bibinfo{person}{Yicheng Fu}, \bibinfo{person}{Weng~Lam Tam}, \bibinfo{person}{Zhengxiao Du}, \bibinfo{person}{Zhilin Yang}, {and} \bibinfo{person}{Jie Tang}.} \bibinfo{year}{2021}\natexlab{}.
\newblock \showarticletitle{P-tuning v2: Prompt tuning can be comparable to fine-tuning universally across scales and tasks}.
\newblock \bibinfo{journal}{\emph{arXiv preprint arXiv:2110.07602}} (\bibinfo{year}{2021}).
\newblock


\bibitem[Lu et~al\mbox{.}(2024)]%
        {lu2024ai}
\bibfield{author}{\bibinfo{person}{Chris Lu}, \bibinfo{person}{Cong Lu}, \bibinfo{person}{Robert~Tjarko Lange}, \bibinfo{person}{Jakob Foerster}, \bibinfo{person}{Jeff Clune}, {and} \bibinfo{person}{David Ha}.} \bibinfo{year}{2024}\natexlab{}.
\newblock \showarticletitle{The ai scientist: Towards fully automated open-ended scientific discovery}.
\newblock \bibinfo{journal}{\emph{arXiv preprint arXiv:2408.06292}} (\bibinfo{year}{2024}).
\newblock


\bibitem[{Magniv, Inc.}(2025)]%
        {promptlayer2025}
\bibfield{author}{\bibinfo{person}{{Magniv, Inc.}}} \bibinfo{year}{2025}\natexlab{}.
\newblock \bibinfo{booktitle}{\emph{PromptLayer: Platform for Prompt Engineering, Management, Evaluation, and LLM Observability}}.
\newblock
\urldef\tempurl%
\url{https://www.promptlayer.com/}
\showURL{%
\tempurl}


\bibitem[{Microsoft}(2025)]%
        {baipishu}
\bibfield{author}{\bibinfo{person}{{Microsoft}}.} \bibinfo{year}{2025}\natexlab{}.
\newblock \bibinfo{title}{Taxonomy of Failure Mode in Agentic AI Systems}.
\newblock
\urldef\tempurl%
\url{https://cdn-dynmedia-1.microsoft.com/is/content/microsoftcorp/microsoft/final/en-us/microsoft-brand/documents/Taxonomy-of-Failure-Mode-in-Agentic-AI-Systems-Whitepaper.pdf}
\showURL{%
\tempurl}


\bibitem[Microsoft(2025)]%
        {Microsoft}
\bibfield{author}{\bibinfo{person}{Microsoft}.} \bibinfo{year}{2025}\natexlab{}.
\newblock \showarticletitle{Taxonomy of Failure Mode in Agentic AI Systems}.
\newblock  (\bibinfo{year}{2025}).
\newblock


\bibitem[OG(2025)]%
        {confblog}
\bibfield{author}{\bibinfo{person}{Pondhouse~Data OG}.} \bibinfo{year}{2025}\natexlab{}.
\newblock \bibinfo{booktitle}{\emph{Building High-Quality AI Agent Systems: Best Practices}}.
\newblock
\urldef\tempurl%
\url{https://www.pondhouse-data.com/blog/high-quality-ai-agent-systems?utm_source=chatgpt.com}
\showURL{%
\tempurl}


\bibitem[{OpenLIT}(2025)]%
        {openlit2025}
\bibfield{author}{\bibinfo{person}{{OpenLIT}}.} \bibinfo{year}{2025}\natexlab{}.
\newblock \bibinfo{booktitle}{\emph{OpenLIT: OpenTelemetry-native GenAI and LLM Application Observability}}.
\newblock
\urldef\tempurl%
\url{https://openlit.io/}
\showURL{%
\tempurl}


\bibitem[Ouyang et~al\mbox{.}(2022)]%
        {rlhf}
\bibfield{author}{\bibinfo{person}{Long Ouyang}, \bibinfo{person}{Jeffrey Wu}, \bibinfo{person}{Xu Jiang}, \bibinfo{person}{Diogo Almeida}, \bibinfo{person}{Carroll Wainwright}, \bibinfo{person}{Pamela Mishkin}, \bibinfo{person}{Chong Zhang}, \bibinfo{person}{Sandhini Agarwal}, \bibinfo{person}{Katarina Slama}, \bibinfo{person}{Alex Ray}, {et~al\mbox{.}}} \bibinfo{year}{2022}\natexlab{}.
\newblock \showarticletitle{Training language models to follow instructions with human feedback}.
\newblock \bibinfo{journal}{\emph{Advances in neural information processing systems}}  \bibinfo{volume}{35} (\bibinfo{year}{2022}), \bibinfo{pages}{27730--27744}.
\newblock


\bibitem[Park et~al\mbox{.}(2023)]%
        {98}
\bibfield{author}{\bibinfo{person}{Jeongeun Park}, \bibinfo{person}{Seungwon Lim}, \bibinfo{person}{Joonhyung Lee}, \bibinfo{person}{Sangbeom Park}, \bibinfo{person}{Minsuk Chang}, \bibinfo{person}{Youngjae Yu}, {and} \bibinfo{person}{Sungjoon Choi}.} \bibinfo{year}{2023}\natexlab{}.
\newblock \showarticletitle{Clara: classifying and disambiguating user commands for reliable interactive robotic agents}.
\newblock \bibinfo{journal}{\emph{IEEE Robotics and Automation Letters}} \bibinfo{volume}{9}, \bibinfo{number}{2} (\bibinfo{year}{2023}), \bibinfo{pages}{1059--1066}.
\newblock


\bibitem[Pei et~al\mbox{.}(2025)]%
        {pei2025flow}
\bibfield{author}{\bibinfo{person}{Changhua Pei}, \bibinfo{person}{Zexin Wang}, \bibinfo{person}{Fengrui Liu}, \bibinfo{person}{Zeyan Li}, \bibinfo{person}{Yang Liu}, \bibinfo{person}{Xiao He}, \bibinfo{person}{Rong Kang}, \bibinfo{person}{Tieying Zhang}, \bibinfo{person}{Jianjun Chen}, \bibinfo{person}{Jianhui Li}, {et~al\mbox{.}}} \bibinfo{year}{2025}\natexlab{}.
\newblock \showarticletitle{Flow-of-Action: SOP Enhanced LLM-Based Multi-Agent System for Root Cause Analysis}. In \bibinfo{booktitle}{\emph{Companion Proceedings of the ACM on Web Conference 2025}}. \bibinfo{pages}{422--431}.
\newblock


\bibitem[Platon et~al\mbox{.}(2007)]%
        {conf1}
\bibfield{author}{\bibinfo{person}{Eric Platon} {et~al\mbox{.}}} \bibinfo{year}{2007}\natexlab{}.
\newblock \emph{\bibinfo{title}{Modeling exception management in multi-agent systems.}}
\newblock \bibinfo{thesistype}{Ph.\,D. Dissertation}. \bibinfo{school}{Citeseer}.
\newblock


\bibitem[Qian et~al\mbox{.}(2023)]%
        {Qian2023ChatDevCA}
\bibfield{author}{\bibinfo{person}{Cheng Qian}, \bibinfo{person}{Wei Liu}, \bibinfo{person}{Hongzhang Liu}, \bibinfo{person}{Nuo Chen}, \bibinfo{person}{Yufan Dang}, \bibinfo{person}{Jiahao Li}, \bibinfo{person}{Cheng Yang}, \bibinfo{person}{Weize Chen}, \bibinfo{person}{Yusheng Su}, \bibinfo{person}{Xin Cong}, \bibinfo{person}{Juyuan Xu}, \bibinfo{person}{Dahai Li}, \bibinfo{person}{Zhiyuan Liu}, {and} \bibinfo{person}{Maosong Sun}.} \bibinfo{year}{2023}\natexlab{}.
\newblock \showarticletitle{ChatDev: Communicative Agents for Software Development}. In \bibinfo{booktitle}{\emph{Annual Meeting of the Association for Computational Linguistics}}.
\newblock
\urldef\tempurl%
\url{https://api.semanticscholar.org/CorpusID:270257715}
\showURL{%
\tempurl}


\bibitem[Qin et~al\mbox{.}(2023)]%
        {qin2023toolllm}
\bibfield{author}{\bibinfo{person}{Yujia Qin}, \bibinfo{person}{Shihao Liang}, \bibinfo{person}{Yining Ye}, \bibinfo{person}{Kunlun Zhu}, \bibinfo{person}{Lan Yan}, \bibinfo{person}{Yaxi Lu}, \bibinfo{person}{Yankai Lin}, \bibinfo{person}{Xin Cong}, \bibinfo{person}{Xiangru Tang}, \bibinfo{person}{Bill Qian}, {et~al\mbox{.}}} \bibinfo{year}{2023}\natexlab{}.
\newblock \showarticletitle{Toolllm: Facilitating large language models to master 16000+ real-world apis}.
\newblock \bibinfo{journal}{\emph{arXiv preprint arXiv:2307.16789}} (\bibinfo{year}{2023}).
\newblock


\bibitem[Quach et~al\mbox{.}(2023)]%
        {quach2023conformal}
\bibfield{author}{\bibinfo{person}{Victor Quach}, \bibinfo{person}{Adam Fisch}, \bibinfo{person}{Tal Schuster}, \bibinfo{person}{Adam Yala}, \bibinfo{person}{Jae~Ho Sohn}, \bibinfo{person}{Tommi~S Jaakkola}, {and} \bibinfo{person}{Regina Barzilay}.} \bibinfo{year}{2023}\natexlab{}.
\newblock \showarticletitle{Conformal language modeling}.
\newblock \bibinfo{journal}{\emph{arXiv preprint arXiv:2306.10193}} (\bibinfo{year}{2023}).
\newblock


\bibitem[Rao et~al\mbox{.}(1995)]%
        {rao1995bdi}
\bibfield{author}{\bibinfo{person}{Anand~S Rao}, \bibinfo{person}{Michael~P Georgeff}, {et~al\mbox{.}}} \bibinfo{year}{1995}\natexlab{}.
\newblock \showarticletitle{BDI agents: from theory to practice.}. In \bibinfo{booktitle}{\emph{Icmas}}, Vol.~\bibinfo{volume}{95}. \bibinfo{pages}{312--319}.
\newblock


\bibitem[Rawte et~al\mbox{.}(2023)]%
        {109}
\bibfield{author}{\bibinfo{person}{Vipula Rawte}, \bibinfo{person}{Amit Sheth}, {and} \bibinfo{person}{Amitava Das}.} \bibinfo{year}{2023}\natexlab{}.
\newblock \showarticletitle{A survey of hallucination in large foundation models}.
\newblock \bibinfo{journal}{\emph{arXiv preprint arXiv:2309.05922}} (\bibinfo{year}{2023}).
\newblock


\bibitem[Ren et~al\mbox{.}(2023)]%
        {110}
\bibfield{author}{\bibinfo{person}{Allen~Z Ren}, \bibinfo{person}{Anushri Dixit}, \bibinfo{person}{Alexandra Bodrova}, \bibinfo{person}{Sumeet Singh}, \bibinfo{person}{Stephen Tu}, \bibinfo{person}{Noah Brown}, \bibinfo{person}{Peng Xu}, \bibinfo{person}{Leila Takayama}, \bibinfo{person}{Fei Xia}, \bibinfo{person}{Jake Varley}, {et~al\mbox{.}}} \bibinfo{year}{2023}\natexlab{}.
\newblock \showarticletitle{Robots that ask for help: Uncertainty alignment for large language model planners}.
\newblock \bibinfo{journal}{\emph{arXiv preprint arXiv:2307.01928}} (\bibinfo{year}{2023}).
\newblock


\bibitem[Sanjeev(2025)]%
        {blogemergent}
\bibfield{author}{\bibinfo{person}{Sanjeev}.} \bibinfo{year}{2025}\natexlab{}.
\newblock \bibinfo{booktitle}{\emph{Emergent Behavior in Multi-Agent Systems: How Complex Behaviors Arise from Simple Agent Interactions}}.
\newblock
\urldef\tempurl%
\url{https://medium.com/%40sanjeevseengh/emergent-behavior-in-multi-agent-systems-how-complex-behaviors-arise-from-simple-agent-0e4503b376ce}
\showURL{%
\tempurl}


\bibitem[Shen et~al\mbox{.}(2025)]%
        {cl3}
\bibfield{author}{\bibinfo{person}{Junhong Shen}, \bibinfo{person}{Hao Bai}, \bibinfo{person}{Lunjun Zhang}, \bibinfo{person}{Yifei Zhou}, \bibinfo{person}{Amrith Setlur}, \bibinfo{person}{Shengbang Tong}, \bibinfo{person}{Diego Caples}, \bibinfo{person}{Nan Jiang}, \bibinfo{person}{Tong Zhang}, \bibinfo{person}{Ameet Talwalkar}, {et~al\mbox{.}}} \bibinfo{year}{2025}\natexlab{}.
\newblock \showarticletitle{Thinking vs. Doing: Agents that Reason by Scaling Test-Time Interaction}.
\newblock \bibinfo{journal}{\emph{arXiv preprint arXiv:2506.07976}} (\bibinfo{year}{2025}).
\newblock


\bibitem[Shi et~al\mbox{.}(2025)]%
        {guiagents}
\bibfield{author}{\bibinfo{person}{Yucheng Shi}, \bibinfo{person}{Wenhao Yu}, \bibinfo{person}{Wenlin Yao}, \bibinfo{person}{Wenhu Chen}, {and} \bibinfo{person}{Ninghao Liu}.} \bibinfo{year}{2025}\natexlab{}.
\newblock \showarticletitle{Towards trustworthy gui agents: A survey}.
\newblock \bibinfo{journal}{\emph{arXiv preprint arXiv:2503.23434}} (\bibinfo{year}{2025}).
\newblock


\bibitem[Shi and Lipani(2023)]%
        {prom13}
\bibfield{author}{\bibinfo{person}{Zhengxiang Shi} {and} \bibinfo{person}{Aldo Lipani}.} \bibinfo{year}{2023}\natexlab{}.
\newblock \showarticletitle{Dept: Decomposed prompt tuning for parameter-efficient fine-tuning}.
\newblock \bibinfo{journal}{\emph{arXiv preprint arXiv:2309.05173}} (\bibinfo{year}{2023}).
\newblock


\bibitem[Shin et~al\mbox{.}(2020)]%
        {prom11}
\bibfield{author}{\bibinfo{person}{Taylor Shin}, \bibinfo{person}{Yasaman Razeghi}, \bibinfo{person}{Robert~L Logan~IV}, \bibinfo{person}{Eric Wallace}, {and} \bibinfo{person}{Sameer Singh}.} \bibinfo{year}{2020}\natexlab{}.
\newblock \showarticletitle{Autoprompt: Eliciting knowledge from language models with automatically generated prompts}.
\newblock \bibinfo{journal}{\emph{arXiv preprint arXiv:2010.15980}} (\bibinfo{year}{2020}).
\newblock


\bibitem[Shinn et~al\mbox{.}(2023)]%
        {shinn2023reflexion}
\bibfield{author}{\bibinfo{person}{Noah Shinn}, \bibinfo{person}{Federico Cassano}, \bibinfo{person}{Ashwin Gopinath}, \bibinfo{person}{Karthik Narasimhan}, {and} \bibinfo{person}{Shunyu Yao}.} \bibinfo{year}{2023}\natexlab{}.
\newblock \showarticletitle{Reflexion: Language agents with verbal reinforcement learning}.
\newblock \bibinfo{journal}{\emph{Advances in Neural Information Processing Systems}}  \bibinfo{volume}{36} (\bibinfo{year}{2023}), \bibinfo{pages}{8634--8652}.
\newblock


\bibitem[Sun et~al\mbox{.}(2024)]%
        {sun2024redeep}
\bibfield{author}{\bibinfo{person}{Zhongxiang Sun}, \bibinfo{person}{Xiaoxue Zang}, \bibinfo{person}{Kai Zheng}, \bibinfo{person}{Yang Song}, \bibinfo{person}{Jun Xu}, \bibinfo{person}{Xiao Zhang}, \bibinfo{person}{Weijie Yu}, {and} \bibinfo{person}{Han Li}.} \bibinfo{year}{2024}\natexlab{}.
\newblock \showarticletitle{Redeep: Detecting hallucination in retrieval-augmented generation via mechanistic interpretability}.
\newblock \bibinfo{journal}{\emph{arXiv preprint arXiv:2410.11414}} (\bibinfo{year}{2024}).
\newblock


\bibitem[{SWE-bench Team}(2025)]%
        {swebch_2025}
\bibfield{author}{\bibinfo{person}{{SWE-bench Team}}.} \bibinfo{year}{2025}\natexlab{}.
\newblock \bibinfo{booktitle}{\emph{{SWE-bench}: A Benchmark for Evaluating Software Engineering Agents}}.
\newblock
\urldef\tempurl%
\url{https://www.swebench.com}
\showURL{%
\tempurl}
\newblock
\shownote{Leaderboard tracking AI agent performance on software engineering tasks}.


\bibitem[Tencent(2025)]%
        {ai-infra-guard}
\bibfield{author}{\bibinfo{person}{Tencent}.} \bibinfo{year}{2025}\natexlab{}.
\newblock \bibinfo{booktitle}{\emph{Ai-Infra-Guard}}.
\newblock
\urldef\tempurl%
\url{https://github.com/Tencent/AI-Infra-Guard}
\showURL{%
\tempurl}


\bibitem[Tong et~al\mbox{.}(2025)]%
        {evo2}
\bibfield{author}{\bibinfo{person}{Zeliang Tong}, \bibinfo{person}{Zhuojun Ding}, {and} \bibinfo{person}{Wei Wei}.} \bibinfo{year}{2025}\natexlab{}.
\newblock \showarticletitle{EvoPrompt: Evolving Prompts for Enhanced Zero-Shot Named Entity Recognition with Large Language Models}. In \bibinfo{booktitle}{\emph{Proceedings of the 31st International Conference on Computational Linguistics}}. \bibinfo{pages}{5136--5153}.
\newblock


\bibitem[{Traceloop}(2025)]%
        {openllmetry2025}
\bibfield{author}{\bibinfo{person}{{Traceloop}}.} \bibinfo{year}{2025}\natexlab{}.
\newblock \bibinfo{booktitle}{\emph{OpenLLMetry: Open‑source observability for your LLM application}}.
\newblock
\urldef\tempurl%
\url{https://github.com/traceloop/openllmetry}
\showURL{%
\tempurl}


\bibitem[{TruEra Inc.}(2025)]%
        {trulens2025}
\bibfield{author}{\bibinfo{person}{{TruEra Inc.}}} \bibinfo{year}{2025}\natexlab{}.
\newblock \bibinfo{booktitle}{\emph{TruLens: Open‑Source LLM Evaluation \& Observability Platform}}.
\newblock
\urldef\tempurl%
\url{https://www.trulens.org/}
\showURL{%
\tempurl}


\bibitem[Wang and Sun(2025)]%
        {PI-LLM}
\bibfield{author}{\bibinfo{person}{Chupei Wang} {and} \bibinfo{person}{Jiaqiu~Vince Sun}.} \bibinfo{year}{2025}\natexlab{}.
\newblock \showarticletitle{Unable to Forget: Proactive lnterference Reveals Working Memory Limits in LLMs Beyond Context Length}. In \bibinfo{booktitle}{\emph{ICML 2025 Workshop on Long-Context Foundation Models}}.
\newblock


\bibitem[Wang et~al\mbox{.}(2024b)]%
        {wang2024astuterag}
\bibfield{author}{\bibinfo{person}{Fei Wang}, \bibinfo{person}{Xingchen Wan}, \bibinfo{person}{Ruoxi Sun}, \bibinfo{person}{Jiefeng Chen}, {and} \bibinfo{person}{Sercan~{\"O} Ar{\i}k}.} \bibinfo{year}{2024}\natexlab{b}.
\newblock \showarticletitle{Astute rag: Overcoming imperfect retrieval augmentation and knowledge conflicts for large language models}.
\newblock \bibinfo{journal}{\emph{arXiv preprint arXiv:2410.07176}} (\bibinfo{year}{2024}).
\newblock


\bibitem[Wang et~al\mbox{.}(2023)]%
        {136}
\bibfield{author}{\bibinfo{person}{Jun Wang}, \bibinfo{person}{Jiaming Tong}, \bibinfo{person}{Kaiyuan Tan}, \bibinfo{person}{Yevgeniy Vorobeychik}, {and} \bibinfo{person}{Yiannis Kantaros}.} \bibinfo{year}{2023}\natexlab{}.
\newblock \showarticletitle{Conformal temporal logic planning using large language models}.
\newblock \bibinfo{journal}{\emph{arXiv preprint arXiv:2309.10092}} (\bibinfo{year}{2023}).
\newblock


\bibitem[WANG et~al\mbox{.}(2025)]%
        {industrialfunction}
\bibfield{author}{\bibinfo{person}{MAOLIN WANG}, \bibinfo{person}{YINGYI ZHANG}, \bibinfo{person}{CUNYIN PENG}, \bibinfo{person}{YICHENG CHEN}, \bibinfo{person}{WEI ZHOU}, \bibinfo{person}{JINJIE GU}, \bibinfo{person}{CHENYI ZHUANG}, \bibinfo{person}{RUOCHENG GUO}, \bibinfo{person}{BOWEN YU}, \bibinfo{person}{WANYU WANG}, {et~al\mbox{.}}} \bibinfo{year}{2025}\natexlab{}.
\newblock \showarticletitle{Function Calling in Large Language Models: Industrial Practices, Challenges, and Future Directions}.
\newblock  (\bibinfo{year}{2025}).
\newblock


\bibitem[Wang et~al\mbox{.}(2024a)]%
        {codeact}
\bibfield{author}{\bibinfo{person}{Xingyao Wang}, \bibinfo{person}{Yangyi Chen}, \bibinfo{person}{Lifan Yuan}, \bibinfo{person}{Yizhe Zhang}, \bibinfo{person}{Yunzhu Li}, \bibinfo{person}{Hao Peng}, {and} \bibinfo{person}{Heng Ji}.} \bibinfo{year}{2024}\natexlab{a}.
\newblock \showarticletitle{Executable code actions elicit better llm agents}. In \bibinfo{booktitle}{\emph{Forty-first International Conference on Machine Learning}}.
\newblock


\bibitem[Wang et~al\mbox{.}(2021)]%
        {cl1}
\bibfield{author}{\bibinfo{person}{Xin Wang}, \bibinfo{person}{Yudong Chen}, {and} \bibinfo{person}{Wenwu Zhu}.} \bibinfo{year}{2021}\natexlab{}.
\newblock \showarticletitle{A survey on curriculum learning}.
\newblock \bibinfo{journal}{\emph{IEEE transactions on pattern analysis and machine intelligence}} \bibinfo{volume}{44}, \bibinfo{number}{9} (\bibinfo{year}{2021}), \bibinfo{pages}{4555--4576}.
\newblock


\bibitem[Wang et~al\mbox{.}(2025a)]%
        {wang2025openhands}
\bibfield{author}{\bibinfo{person}{Xingyao Wang}, \bibinfo{person}{Boxuan Li}, \bibinfo{person}{Yufan Song}, \bibinfo{person}{Frank~F. Xu}, \bibinfo{person}{Xiangru Tang}, \bibinfo{person}{Mingchen Zhuge}, \bibinfo{person}{Jiayi Pan}, \bibinfo{person}{Yueqi Song}, \bibinfo{person}{Bowen Li}, \bibinfo{person}{Jaskirat Singh}, \bibinfo{person}{Hoang~H. Tran}, \bibinfo{person}{Fuqiang Li}, \bibinfo{person}{Ren Ma}, \bibinfo{person}{Mingzhang Zheng}, \bibinfo{person}{Bill Qian}, \bibinfo{person}{Yanjun Shao}, \bibinfo{person}{Niklas Muennighoff}, \bibinfo{person}{Yizhe Zhang}, \bibinfo{person}{Binyuan Hui}, \bibinfo{person}{Junyang Lin}, \bibinfo{person}{Robert Brennan}, \bibinfo{person}{Hao Peng}, \bibinfo{person}{Heng Ji}, {and} \bibinfo{person}{Graham Neubig}.} \bibinfo{year}{2025}\natexlab{a}.
\newblock \showarticletitle{OpenHands: An Open Platform for {AI} Software Developers as Generalist Agents}. In \bibinfo{booktitle}{\emph{The Thirteenth International Conference on Learning Representations}}.
\newblock
\urldef\tempurl%
\url{https://openreview.net/forum?id=OJd3ayDDoF}
\showURL{%
\tempurl}


\bibitem[Wang et~al\mbox{.}(2022)]%
        {wang2022selfconsistency}
\bibfield{author}{\bibinfo{person}{Xuezhi Wang}, \bibinfo{person}{Jason Wei}, \bibinfo{person}{Dale Schuurmans}, \bibinfo{person}{Quoc Le}, \bibinfo{person}{Ed Chi}, \bibinfo{person}{Sharan Narang}, \bibinfo{person}{Aakanksha Chowdhery}, {and} \bibinfo{person}{Denny Zhou}.} \bibinfo{year}{2022}\natexlab{}.
\newblock \showarticletitle{Self-consistency improves chain of thought reasoning in language models}.
\newblock \bibinfo{journal}{\emph{arXiv preprint arXiv:2203.11171}} (\bibinfo{year}{2022}).
\newblock


\bibitem[Wang et~al\mbox{.}(2025b)]%
        {rl3}
\bibfield{author}{\bibinfo{person}{Zihan Wang}, \bibinfo{person}{Kangrui Wang}, \bibinfo{person}{Qineng Wang}, \bibinfo{person}{Pingyue Zhang}, \bibinfo{person}{Linjie Li}, \bibinfo{person}{Zhengyuan Yang}, \bibinfo{person}{Xing Jin}, \bibinfo{person}{Kefan Yu}, \bibinfo{person}{Minh~Nhat Nguyen}, \bibinfo{person}{Licheng Liu}, {et~al\mbox{.}}} \bibinfo{year}{2025}\natexlab{b}.
\newblock \showarticletitle{Ragen: Understanding self-evolution in llm agents via multi-turn reinforcement learning}.
\newblock \bibinfo{journal}{\emph{arXiv preprint arXiv:2504.20073}} (\bibinfo{year}{2025}).
\newblock


\bibitem[Wei et~al\mbox{.}(2021)]%
        {sft}
\bibfield{author}{\bibinfo{person}{Jason Wei}, \bibinfo{person}{Maarten Bosma}, \bibinfo{person}{Vincent~Y Zhao}, \bibinfo{person}{Kelvin Guu}, \bibinfo{person}{Adams~Wei Yu}, \bibinfo{person}{Brian Lester}, \bibinfo{person}{Nan Du}, \bibinfo{person}{Andrew~M Dai}, {and} \bibinfo{person}{Quoc~V Le}.} \bibinfo{year}{2021}\natexlab{}.
\newblock \showarticletitle{Finetuned language models are zero-shot learners}.
\newblock \bibinfo{journal}{\emph{arXiv preprint arXiv:2109.01652}} (\bibinfo{year}{2021}).
\newblock


\bibitem[Wei et~al\mbox{.}(2022)]%
        {wei2022chainofthought}
\bibfield{author}{\bibinfo{person}{Jason Wei}, \bibinfo{person}{Xuezhi Wang}, \bibinfo{person}{Dale Schuurmans}, \bibinfo{person}{Maarten Bosma}, \bibinfo{person}{Fei Xia}, \bibinfo{person}{Ed Chi}, \bibinfo{person}{Quoc~V Le}, \bibinfo{person}{Denny Zhou}, {et~al\mbox{.}}} \bibinfo{year}{2022}\natexlab{}.
\newblock \showarticletitle{Chain-of-thought prompting elicits reasoning in large language models}.
\newblock \bibinfo{journal}{\emph{Advances in neural information processing systems}}  \bibinfo{volume}{35} (\bibinfo{year}{2022}), \bibinfo{pages}{24824--24837}.
\newblock


\bibitem[Wu et~al\mbox{.}(2024)]%
        {wu2024darkside}
\bibfield{author}{\bibinfo{person}{Zihui Wu}, \bibinfo{person}{Haichang Gao}, \bibinfo{person}{Jianping He}, {and} \bibinfo{person}{Ping Wang}.} \bibinfo{year}{2024}\natexlab{}.
\newblock \showarticletitle{The dark side of function calling: Pathways to jailbreaking large language models}.
\newblock \bibinfo{journal}{\emph{arXiv preprint arXiv:2407.17915}} (\bibinfo{year}{2024}).
\newblock


\bibitem[Xia et~al\mbox{.}(2025)]%
        {cl2}
\bibfield{author}{\bibinfo{person}{Peng Xia}, \bibinfo{person}{Jinglu Wang}, \bibinfo{person}{Yibo Peng}, \bibinfo{person}{Kaide Zeng}, \bibinfo{person}{Xian Wu}, \bibinfo{person}{Xiangru Tang}, \bibinfo{person}{Hongtu Zhu}, \bibinfo{person}{Yun Li}, \bibinfo{person}{Shujie Liu}, \bibinfo{person}{Yan Lu}, {et~al\mbox{.}}} \bibinfo{year}{2025}\natexlab{}.
\newblock \showarticletitle{MMedAgent-RL: Optimizing Multi-Agent Collaboration for Multimodal Medical Reasoning}.
\newblock \bibinfo{journal}{\emph{arXiv preprint arXiv:2506.00555}} (\bibinfo{year}{2025}).
\newblock


\bibitem[Yang et~al\mbox{.}(2024)]%
        {yang2024alignment}
\bibfield{author}{\bibinfo{person}{Yuqing Yang}, \bibinfo{person}{Ethan Chern}, \bibinfo{person}{Xipeng Qiu}, \bibinfo{person}{Graham Neubig}, {and} \bibinfo{person}{Pengfei Liu}.} \bibinfo{year}{2024}\natexlab{}.
\newblock \showarticletitle{Alignment for honesty}.
\newblock \bibinfo{journal}{\emph{Advances in Neural Information Processing Systems}}  \bibinfo{volume}{37} (\bibinfo{year}{2024}), \bibinfo{pages}{63565--63598}.
\newblock


\bibitem[Yao et~al\mbox{.}(2023)]%
        {yao2023react}
\bibfield{author}{\bibinfo{person}{Shunyu Yao}, \bibinfo{person}{Jeffrey Zhao}, \bibinfo{person}{Dian Yu}, \bibinfo{person}{Nan Du}, \bibinfo{person}{Izhak Shafran}, \bibinfo{person}{Karthik Narasimhan}, {and} \bibinfo{person}{Yuan Cao}.} \bibinfo{year}{2023}\natexlab{}.
\newblock \showarticletitle{React: Synergizing reasoning and acting in language models}. In \bibinfo{booktitle}{\emph{International Conference on Learning Representations (ICLR)}}.
\newblock


\bibitem[Yuan and Xie(2025)]%
        {vote3}
\bibfield{author}{\bibinfo{person}{Yurun Yuan} {and} \bibinfo{person}{Tengyang Xie}.} \bibinfo{year}{2025}\natexlab{}.
\newblock \showarticletitle{Reinforce LLM Reasoning through Multi-Agent Reflection}.
\newblock \bibinfo{journal}{\emph{arXiv preprint arXiv:2506.08379}} (\bibinfo{year}{2025}).
\newblock


\bibitem[Zeng et~al\mbox{.}(2025)]%
        {rl2}
\bibfield{author}{\bibinfo{person}{Siliang Zeng}, \bibinfo{person}{Quan Wei}, \bibinfo{person}{William Brown}, \bibinfo{person}{Oana Frunza}, \bibinfo{person}{Yuriy Nevmyvaka}, {and} \bibinfo{person}{Mingyi Hong}.} \bibinfo{year}{2025}\natexlab{}.
\newblock \showarticletitle{Reinforcing Multi-Turn Reasoning in LLM Agents via Turn-Level Credit Assignment}.
\newblock \bibinfo{journal}{\emph{arXiv preprint arXiv:2505.11821}} (\bibinfo{year}{2025}).
\newblock


\bibitem[Zhang et~al\mbox{.}(2024b)]%
        {zhang2024agentprune}
\bibfield{author}{\bibinfo{person}{Guibin Zhang}, \bibinfo{person}{Yanwei Yue}, \bibinfo{person}{Zhixun Li}, \bibinfo{person}{Sukwon Yun}, \bibinfo{person}{Guancheng Wan}, \bibinfo{person}{Kun Wang}, \bibinfo{person}{Dawei Cheng}, \bibinfo{person}{Jeffrey~Xu Yu}, {and} \bibinfo{person}{Tianlong Chen}.} \bibinfo{year}{2024}\natexlab{b}.
\newblock \showarticletitle{Cut the crap: An economical communication pipeline for llm-based multi-agent systems}.
\newblock \bibinfo{journal}{\emph{arXiv preprint arXiv:2410.02506}} (\bibinfo{year}{2024}).
\newblock


\bibitem[Zhang et~al\mbox{.}(2024c)]%
        {zhang2024gdesigner}
\bibfield{author}{\bibinfo{person}{Guibin Zhang}, \bibinfo{person}{Yanwei Yue}, \bibinfo{person}{Xiangguo Sun}, \bibinfo{person}{Guancheng Wan}, \bibinfo{person}{Miao Yu}, \bibinfo{person}{Junfeng Fang}, \bibinfo{person}{Kun Wang}, \bibinfo{person}{Tianlong Chen}, {and} \bibinfo{person}{Dawei Cheng}.} \bibinfo{year}{2024}\natexlab{c}.
\newblock \showarticletitle{G-designer: Architecting multi-agent communication topologies via graph neural networks}.
\newblock \bibinfo{journal}{\emph{arXiv preprint arXiv:2410.11782}} (\bibinfo{year}{2024}).
\newblock


\bibitem[Zhang et~al\mbox{.}(2025b)]%
        {zhang2025qerag}
\bibfield{author}{\bibinfo{person}{Kepu Zhang}, \bibinfo{person}{Zhongxiang Sun}, \bibinfo{person}{Weijie Yu}, \bibinfo{person}{Xiaoxue Zang}, \bibinfo{person}{Kai Zheng}, \bibinfo{person}{Yang Song}, \bibinfo{person}{Han Li}, {and} \bibinfo{person}{Jun Xu}.} \bibinfo{year}{2025}\natexlab{b}.
\newblock \showarticletitle{QE-RAG: A Robust Retrieval-Augmented Generation Benchmark for Query Entry Errors}.
\newblock \bibinfo{journal}{\emph{arXiv preprint arXiv:2504.04062}} (\bibinfo{year}{2025}).
\newblock


\bibitem[Zhang et~al\mbox{.}(2025d)]%
        {zhang2025agentfm}
\bibfield{author}{\bibinfo{person}{Lingzhe Zhang}, \bibinfo{person}{Yunpeng Zhai}, \bibinfo{person}{Tong Jia}, \bibinfo{person}{Xiaosong Huang}, \bibinfo{person}{Chiming Duan}, {and} \bibinfo{person}{Ying Li}.} \bibinfo{year}{2025}\natexlab{d}.
\newblock \showarticletitle{AgentFM: Role-Aware Failure Management for Distributed Databases with LLM-Driven Multi-Agents}.
\newblock \bibinfo{journal}{\emph{arXiv preprint arXiv:2504.06614}} (\bibinfo{year}{2025}).
\newblock


\bibitem[Zhang et~al\mbox{.}(2025c)]%
        {icml2025}
\bibfield{author}{\bibinfo{person}{Shaokun Zhang}, \bibinfo{person}{Ming Yin}, \bibinfo{person}{Jieyu Zhang}, \bibinfo{person}{Jiale Liu}, \bibinfo{person}{Zhiguang Han}, \bibinfo{person}{Jingyang Zhang}, \bibinfo{person}{Beibin Li}, \bibinfo{person}{Chi Wang}, \bibinfo{person}{Huazheng Wang}, \bibinfo{person}{Yiran Chen}, {et~al\mbox{.}}} \bibinfo{year}{2025}\natexlab{c}.
\newblock \showarticletitle{Which agent causes task failures and when? on automated failure attribution of llm multi-agent systems}.
\newblock \bibinfo{journal}{\emph{arXiv preprint arXiv:2505.00212}} (\bibinfo{year}{2025}).
\newblock


\bibitem[Zhang et~al\mbox{.}(2025a)]%
        {zhang2025webpoilot}
\bibfield{author}{\bibinfo{person}{Yao Zhang}, \bibinfo{person}{Zijian Ma}, \bibinfo{person}{Yunpu Ma}, \bibinfo{person}{Zhen Han}, \bibinfo{person}{Yu Wu}, {and} \bibinfo{person}{Volker Tresp}.} \bibinfo{year}{2025}\natexlab{a}.
\newblock \showarticletitle{WebPilot: A Versatile and Autonomous Multi-Agent System for Web Task Execution with Strategic Exploration}.
\newblock \bibinfo{journal}{\emph{Proceedings of the AAAI Conference on Artificial Intelligence}} \bibinfo{volume}{39}, \bibinfo{number}{22} (\bibinfo{date}{Apr.} \bibinfo{year}{2025}), \bibinfo{pages}{23378--23386}.
\newblock
\href{https://doi.org/10.1609/aaai.v39i22.34505}{doi:\nolinkurl{10.1609/aaai.v39i22.34505}}


\bibitem[Zhang et~al\mbox{.}(2024a)]%
        {CoA}
\bibfield{author}{\bibinfo{person}{Yusen Zhang}, \bibinfo{person}{Ruoxi Sun}, \bibinfo{person}{Yanfei Chen}, \bibinfo{person}{Tomas Pfister}, \bibinfo{person}{Rui Zhang}, {and} \bibinfo{person}{Sercan Arik}.} \bibinfo{year}{2024}\natexlab{a}.
\newblock \showarticletitle{Chain of agents: Large language models collaborating on long-context tasks}.
\newblock \bibinfo{journal}{\emph{Advances in Neural Information Processing Systems}}  \bibinfo{volume}{37} (\bibinfo{year}{2024}), \bibinfo{pages}{132208--132237}.
\newblock


\bibitem[Zheng et~al\mbox{.}(2023b)]%
        {stepback}
\bibfield{author}{\bibinfo{person}{Huaixiu~Steven Zheng}, \bibinfo{person}{Swaroop Mishra}, \bibinfo{person}{Xinyun Chen}, \bibinfo{person}{Heng-Tze Cheng}, \bibinfo{person}{Ed~H Chi}, \bibinfo{person}{Quoc~V Le}, {and} \bibinfo{person}{Denny Zhou}.} \bibinfo{year}{2023}\natexlab{b}.
\newblock \showarticletitle{Take a step back: Evoking reasoning via abstraction in large language models}.
\newblock \bibinfo{journal}{\emph{arXiv preprint arXiv:2310.06117}} (\bibinfo{year}{2023}).
\newblock


\bibitem[Zheng et~al\mbox{.}(2023a)]%
        {judge}
\bibfield{author}{\bibinfo{person}{Lianmin Zheng}, \bibinfo{person}{Wei-Lin Chiang}, \bibinfo{person}{Ying Sheng}, \bibinfo{person}{Siyuan Zhuang}, \bibinfo{person}{Zhanghao Wu}, \bibinfo{person}{Yonghao Zhuang}, \bibinfo{person}{Zi Lin}, \bibinfo{person}{Zhuohan Li}, \bibinfo{person}{Dacheng Li}, \bibinfo{person}{Eric Xing}, {et~al\mbox{.}}} \bibinfo{year}{2023}\natexlab{a}.
\newblock \showarticletitle{Judging llm-as-a-judge with mt-bench and chatbot arena}.
\newblock \bibinfo{journal}{\emph{Advances in Neural Information Processing Systems}}  \bibinfo{volume}{36} (\bibinfo{year}{2023}), \bibinfo{pages}{46595--46623}.
\newblock


\bibitem[Zhou et~al\mbox{.}(2025)]%
        {zhou2025guardian}
\bibfield{author}{\bibinfo{person}{Jialong Zhou}, \bibinfo{person}{Lichao Wang}, {and} \bibinfo{person}{Xiao Yang}.} \bibinfo{year}{2025}\natexlab{}.
\newblock \showarticletitle{GUARDIAN: Safeguarding LLM Multi-Agent Collaborations with Temporal Graph Modeling}.
\newblock \bibinfo{journal}{\emph{arXiv preprint arXiv:2505.19234}} (\bibinfo{year}{2025}).
\newblock


\bibitem[Zhou et~al\mbox{.}(2023b)]%
        {zhou2023webarena}
\bibfield{author}{\bibinfo{person}{Shuyan Zhou}, \bibinfo{person}{Frank~F Xu}, \bibinfo{person}{Hao Zhu}, \bibinfo{person}{Xuhui Zhou}, \bibinfo{person}{Robert Lo}, \bibinfo{person}{Abishek Sridhar}, \bibinfo{person}{Xianyi Cheng}, \bibinfo{person}{Tianyue Ou}, \bibinfo{person}{Yonatan Bisk}, \bibinfo{person}{Daniel Fried}, {et~al\mbox{.}}} \bibinfo{year}{2023}\natexlab{b}.
\newblock \showarticletitle{Webarena: A realistic web environment for building autonomous agents}.
\newblock \bibinfo{journal}{\emph{arXiv preprint arXiv:2307.13854}} (\bibinfo{year}{2023}).
\newblock


\bibitem[Zhou et~al\mbox{.}(2023a)]%
        {zhou202LURE}
\bibfield{author}{\bibinfo{person}{Yiyang Zhou}, \bibinfo{person}{Chenhang Cui}, \bibinfo{person}{Jaehong Yoon}, \bibinfo{person}{Linjun Zhang}, \bibinfo{person}{Zhun Deng}, \bibinfo{person}{Chelsea Finn}, \bibinfo{person}{Mohit Bansal}, {and} \bibinfo{person}{Huaxiu Yao}.} \bibinfo{year}{2023}\natexlab{a}.
\newblock \showarticletitle{Analyzing and mitigating object hallucination in large vision-language models}.
\newblock \bibinfo{journal}{\emph{arXiv preprint arXiv:2310.00754}} (\bibinfo{year}{2023}).
\newblock


\bibitem[Zhu et~al\mbox{.}(2024)]%
        {recur}
\bibfield{author}{\bibinfo{person}{Andrew Zhu}, \bibinfo{person}{Liam Dugan}, {and} \bibinfo{person}{Chris Callison-Burch}.} \bibinfo{year}{2024}\natexlab{}.
\newblock \showarticletitle{ReDel: A Toolkit for LLM-Powered Recursive Multi-Agent Systems}.
\newblock \bibinfo{journal}{\emph{arXiv preprint arXiv:2408.02248}} (\bibinfo{year}{2024}).
\newblock


\end{thebibliography}
